# Decoding the mechanisms of the Hattrick football manager game using Bayesian network structure learning

Anthony C. Constantinou*, Nicholas Higgins, and Neville K. Kitson.

Bayesian AI research lab, Machine Intelligence and Decision Systems (MInDS) research group, School of Electronic Engineering and Computer Science, Queen Mary University of London, London, United Kingdom, E1 4NS.

E-mails: a.constantinou@qmul.ac.uk, nicholashiggins23@gmail.com, and n.k.kitson@qmul.ac.uk.

* Corresponding author.

**Abstract:** Hattrick is a free web-based probabilistic football manager game with over 200,000 users competing for titles at national and international levels. Launched in Sweden in 1997 as part of an MSc project, the game's slow-paced design has fostered a loyal community, with users remaining active for decades. Hattrick's game-engine mechanics are partially hidden, and users have attempted to decode them with incremental success over the years. Rule-based, statistical and machine learning models have been developed to aid this effort and are widely used by the community, but have not been formally evaluated in the scientific literature. This study is the first to explore Hattrick using structure learning techniques and Bayesian networks, integrating expert knowledge with data to develop models that simulate and explain the game-engine. We assess the effectiveness of structure learning algorithms in relation to knowledge-based structures, and publicly share a fully specified Bayesian network model that matches the performance of top models used by the Hattrick community. We further demonstrate how analysis extends beyond prediction by providing a visual representation of dependencies between features, and using the optimal model for in-game decision-making. To support future research, we make all data, graphical structures, and models publicly available online.

## 1. Introduction

Hattrick is an online, browser-based, multilingual football management simulation game developed in Sweden, first launched in 1997. In this game, players build and coach a club over time, typically over many years, and compete with other human managers. As with any management game, players must not only assign positions to their football players and choose from various tactical and strategic options, but also manage the club. This includes training players, trading players on the transfer market, overseeing team staff, managing the club's finances and economic stability, scouting youth players, and managing the team's stadium, amongst other activities. Hattrick's competitive structure is organised by countries, each with its own league pyramid and national cups. The winners of the top division and the main national cup of each country earn spots in the Hattrick Masters, which is Hattrick's most competitive cup.

While Hattrick is free to play, players may choose to upgrade to paid membership which provides extra features to enhance the game experience, but strictly offer no in-game advantage. The number of users peaked in 2009, reaching nearly one million and making it the largest football management game on the web. As one of the oldest such games, Hattrick has a highly loyal player base with today's community consisting of more than 200,000 users worldwide. Unlike other football management games, Hattrick is slow-paced where matches occur once or twice a week for most users with each match lasting 90 minutes just like in real life, whereas players age by one year every 112 days.

The game design focuses on long-term planning, allowing for a more strategic approach to team management and encouraging users to participate for many years, often decades. New players start in the lower divisions of their country, where they are assigned to manage a new club with limited financial resources and player strength. Additionally, because much of the game mechanics remain hidden, users are active on the forums and engage in data analyses aiming to uncover patterns to aid decision-making. There is a wealth of user-led initiatives, ranging from local and online meet-ups to innovative open-source third-party software that supports decision-making across all aspects of Hattrick.

In this paper, we focus on discovering and modelling the underlying mechanisms of Hattrick's match simulation engine using structure learning and Bayesian Networks (BNs). Preliminary information is provided in Section 2. Although the official in-game manual provides some insight into how the game-engine works, many critical details remain undisclosed. Additionally, data often do not align precisely with the manual's descriptions, suggesting that the manual serves more as a general guide than a definitive explanation of the game-engine. Understanding and quantifying the relationships between in-game match simulation factors are important for optimising decisions for match line-ups and strategies against opponents. As a result, the lack of transparency has driven users to study the game-engine since the game's inception, leading to the development of numerous third-party tools designed to replicate its mechanics. In this paper, we also evaluate some of these tools, in addition to the models presented in this study, as part of the evaluation analysis.

Despite Hattrick nearing three decades of existence and having a relatively large user base, it has been studied academically only twice. Moreover, both studies were conducted within the first decade of the game's creation. The first study by Ajalin et al. [1] examined Hattrick as a case study on whether online simulation games could serve as a business model for online gambling. The second study by Borg [2] used Hattrick to test a system for extracting time information from football reports, using the minute-by-minute in-game reports generated by Hattrick during live play.

It is worth noting that Hattrick models have been developed and shared within the community for over a decade, and several of these community-built models are included as evaluation baselines. The models presented and evaluated in this manuscript follow that tradition, providing a more detailed, interpretable, and verifiable formulation while relying solely on publicly observable game outcomes and community knowledge, and disclosing no proprietary content.

While numerous studies have explored various aspects of gaming, such as entertainment, bullying, and learning, this work specifically focuses on the application of structure learning and BNs to understand and simulate game mechanics. Notably, we found no prior research on the application of structure learning to gaming, highlighting the novelty of this study. Moreover, only a limited number of studies are found to have applied BNs to simulate game environments. One of the earliest examples is the work of Værge and Jarlskov [3], who used BNs to simulate the behaviour of game agents by modelling decision-making processes. They showed that while their approach improved agents by displaying more adaptive and intelligent behaviour in certain scenarios, it limited the learning capabilities of agents when dealing with large conditional probability tables (see Section 2.1). Another relevant study is by Price and Goodwin [4], who used BNs to reconstruct areas that a player leaves and later revisits, demonstrating how BNs can efficiently model the likely effects of in-game factors on the environment. Lastly, Hsiao et al. [5] applied Dynamic BNs to capture the stochastic processes of spatial evolutionary games, providing an alternative to agent-based Monte Carlo simulations and achieving improved accuracy in stochastic simulations.

The contributions of this paper are twofold. Firstly, we employ structure learning (also referred to as causal machine learning) algorithms to automatically discover the structure of BN models and parameterise their conditional distributions from data. Secondly, we use domain expertise and high-level information extracted from the game's manual, as well as a much more detailed unofficial user-led manual, to build fully specified knowledge-based BN models. We then evaluate these data-driven and knowledge-based BN models against each other, as well as against well-established user-built models within Hattrick.

What makes this study particularly compelling is not just the novelty of applying BN structural learning to a new problem in an understudied area, but also the opportunity to analyse a probabilistic game-engine that has been assessed by a large user base of gamers for two decades. This allows us to evaluate existing knowledge about Hattrick's game-engine. Moreover, it enables us to evaluate the effectiveness of structure learning algorithms in a case study that differs from existing relevant research. Specifically, unlike traditional BN applications which often focus on real-world observations in critical domains, this study focuses on largely clean, synthetically generated large dataset from computational processes, with some of the mechanisms generating these data remaining deliberately hidden.

We provide preliminary information about BN modelling and structure learning in Section 2, followed by the methodological framework in Section 3. The paper adopts a comprehensive validation strategy combining qualitative validation and quantitative evaluation. To aid readability, the results are organised as follows:

- Section 4 provides qualitative validation through expert review of the knowledge-based BN, including focused assessments of attack simulation, goal scoring, tactical behaviour, and special match events.
- Section 5 presents a quantitative evaluation of graph structures learnt from data against expert-elicited structures, using multiple graphical metrics and model selection scores.
- Section 6 reports quantitative evaluation of predictive performance, including goal distributions by tactic and match outcome prediction, benchmarks results against established Hattrick models (subsections 6.1-6.3), and illustrates decision support via simulated decision scenarios in subsection 6.4).

Finally, Section 7 discusses findings and limitations, Section 8 provides concluding remarks, and extensive supplementary qualitative and quantitative results are included in the appendices.

## 2. Preliminaries

### *2.1. Bayesian networks*

BNs are graphical models that represent the probabilistic relationships amongst a set of variables. The graphical representation of a BN is usually a Directed Acyclic Graph (DAG), in which each node denotes a variable, and each edge a conditional or – in the case of a causal BN - causal dependency between parent (cause) and child (effect) nodes. For instance, if variable $A$ influences variable $B$, there will be a directed edge from $A$ to $B$. The existence, strength and shape of these relationships is determined from data or, in the absence of data, expert knowledge, or a combination of the two.

Relationships between variables can be specified in different ways. For discrete variables, they are defined through a Conditional Probability Table (CPT). For continuous variables, relationships are specified through conditional distributions or functions. Figure 1 presents two fragments taken from the knowledge-based BN model described later in Section 4. The graphical fragment on the left illustrates the relationship between team midfield ratings, possession, and the allocation of exclusive and shared chances between teams (see Section 3 for details). It represents node outputs as conditional Binomial distributions given the two midfield inputs (observations) of 17 and 14, a relationship type applicable to continuous or integer-interval variables. In contrast, the graphical fragment on the right demonstrates how relationships are specified as a CPT when both child and parent nodes are discrete. In this example, the parent nodes are integer intervals treated as discrete variables, enabling the CPT parameterisation shown just below the network on the right in Figure 1.

Judea Pearl [6] laid the foundation for BNs in the 1980s. Since then, BNs have been successfully applied in diverse areas, including healthcare, economics and legal reasoning, sports and marketing. Importantly, BNs greatly contribute to explainability of AI systems by providing clear and interpretable structural models that help users understand how predictions are generated and how decisions are made. They support decision-making by allowing decision-makers to visualise the impact of different variables and interventions before actual implementation. This is achieved through hypothetical scenario analysis, enabling users to explore the consequences of various actions and optimise their decisions based on those results. This capability is particularly valuable in complex and uncertain environments where traditional deterministic models fall short. Moreover, transparency beyond black-box optimisation [7] is crucial in critical applications, such as healthcare and government policy, where understanding the reasoning behind decisions is essential. For a detailed introduction to BNs, see Pearl [8], Darwiche [9], Koller and Friedman [10], Korb and Nicholson [11], and Fenton and Neil [12].

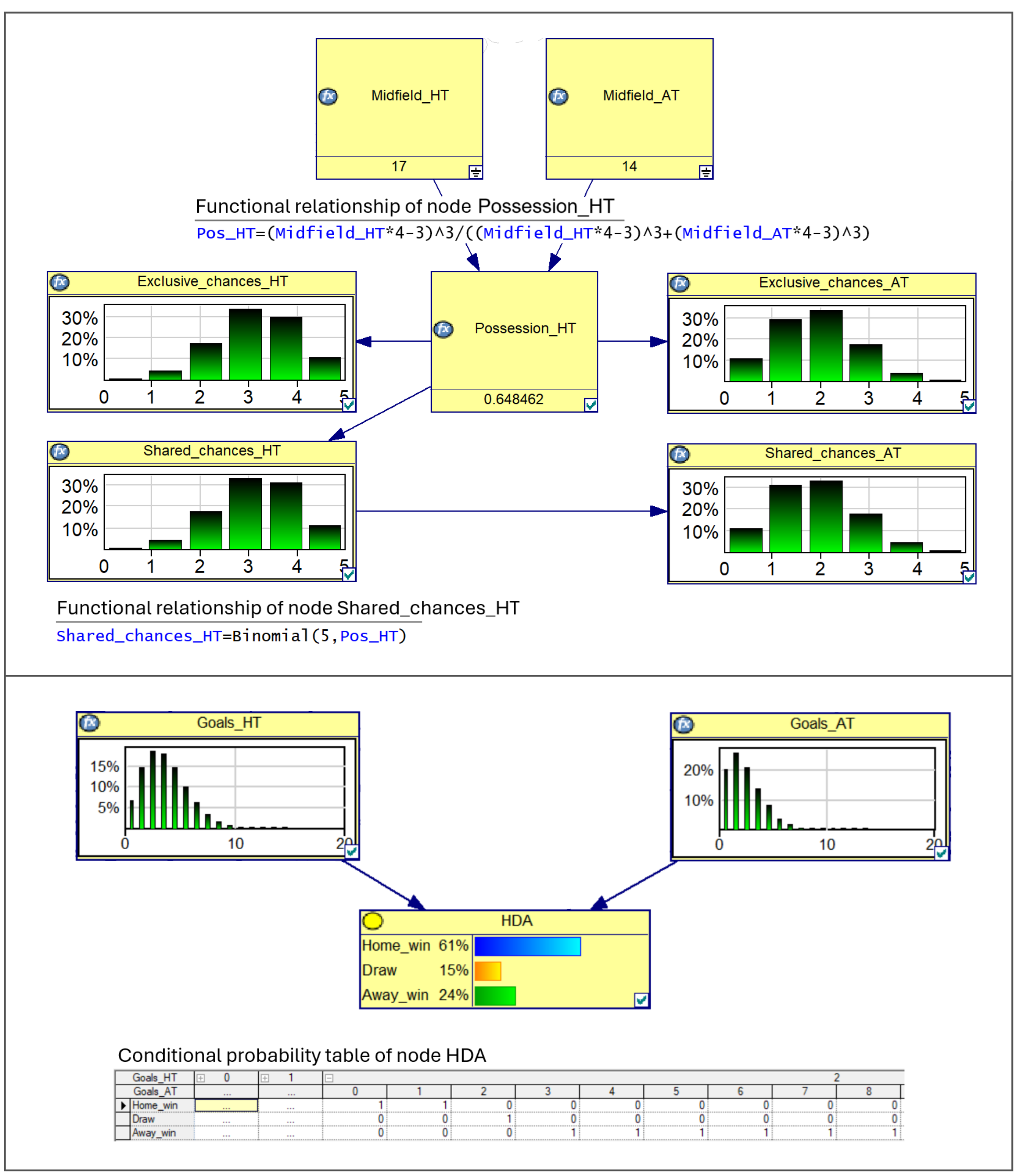

**Figure 1.** Two fragments of the knowledge-based BN described in Section 4 are shown, with hypothetical evidence entered in the Midfield (for HT and AT) nodes for illustration purposes. The top fragment illustrates the output of continuous relationships defined by functions, while the bottom fragment on discrete nodes whose relationships are specified via a CPT. Visualisations are generated using the GeNIe BN software [13]. Detailed variable definitions are provided in Appendix A.

### 2.2. Structure learning

In the previous subsection, we briefly described BNs and the process of parameterising conditional dependencies between variables. Constructing these models involves two critical steps; first determining their graphical structure, and then parameterising the conditional dependencies given the structure. The first step, known as structure learning, is considerably more challenging, both in computational complexity and accuracy, than the subsequent step of parameterisation.

Because BNs are often viewed as causal models (referred to as causal BNs), their underlying causal structure can be elicited from domain experts. However, because experts may be biased, their elicited knowledge can often contradict one another, making it necessary to consider learning such structures from data, or a combining knowledge with machine learning. For example, if we already know that $A$ causes $B$, this information can be provided to an algorithm as a restriction, limiting the search-space of graphs to those containing $A \rightarrow B$. Another example is when knowing that $C$ occurs after observing $D$ and hence, $C$ could not be cause of $D$, restricting the search-space of graphs to those *not* containing $C \rightarrow D$. There are different types of constraints, some of which guide, rather than restrict structure learning. In this paper, we will be exploring all three types of structure learning; from data alone, from knowledge alone, and from a combination of both.

Structure learning involves identifying the DAG structure that best represents the dependencies amongst a set of variables. The challenge in structure learning arises from the combinatorial nature of the problem. Specifically, and as illustrated by Robinson [14] and Kitson et al. [15], the number of possible DAGs grows super-exponentially with the number of variables, making exhaustive search impractical for all but the smallest networks. For example, there are 25 different DAG structures when there are just 3 variables, 29,281 for 5 variables, and over 1 billion for just 7 variables. Therefore, efficient algorithms and heuristic methods are essential to make structure learning feasible in practice.

Various algorithms have been developed to tackle the problem of structure learning, often referred to as causal discovery or causal machine learning, with each algorithm having its own strengths and weaknesses. These algorithms can be broadly categorised into three groups:

i. **Constraint-based** algorithms are designed to perform a set of conditional independence tests to determine the presence of edges, and conditional dependency tests to determine the orientation of some of those edges. The most popular, and one of the earliest, constraint-based algorithms is the PC algorithm by Spirtes et al. [16, 17]. It starts with a fully connected structure and proceeds to eliminate edges that satisfy a given conditional independence test, and orientate those which satisfy a given conditional dependency test, with some additional directionality rules to help orientate some of the remaining undirected edges. The learning strategies of the PC algorithm have been adopted by many subsequent algorithms within this class of learning, such as PC-Stable by Colombo and Maathuis [18] that partially resolves the sensitivity of PC to the order of the variables as read from data, and the Fast Causal Inference (FCI) algorithm by Spirtes et al. [19] that extends PC to handle partially observed data and latent variables.

ii. **Score-Based** algorithms follow a more traditional approach to structure learning, involving searching for different structures and assigning a score to each visited structure. Unlike constraint-based algorithms, score-based algorithms typically start from an empty – rather than a fully connected - structure. The Hill-Climbing (HC) algorithm is one of the earlier [20] and more popular score-based algorithms. It starts from an empty DAG and proceeds to the neighbouring DAG that maximises a given objective function, through edge additions, reversals and deletions. The search continues until no neighbouring DAG is found that further improves the objective score. The model-selection Bayesian Information Criterion (BIC), which balances model fitting with model dimensionality to avoid overfitting, is one of the most widely-used objective functions in this area of research. As with the PC algorithm, many subsequent score-based algorithms build upon HC's concept, with the most popular extension being the TABU algorithm [21, 22] which involves occasionally moving to neighbouring graphs that decrease – rather than increase - the objective score, in an attempt to escape some of the local maxima solutions the HC might get stuck in.

iii. **Hybrid** structure learning algorithms combine elements of both constraint-based and score-based methods. While these methods tend to be more complicated, since they combine two classes of learning, they are not necessarily better than algorithms that rely on just one class of learning. One of the first and more popular hybrid algorithms is the Max-Min Hill Climbing (MMHC) by Tsamardinos et al. [23], that first uses constraint-based learning to construct an undirected graph and then applies a hill-climbing search to orientate the edges and obtain a DAG.

Markov Equivalence Classes (MECs) of Directed Acyclic Graphs (DAGs) address the fact that several DAGs can represent the same data dependencies, meaning that some edge directions cannot be determined from observational data alone. Consequently, conditional independencies identify not a single DAG but a set of compatible ones, which algorithms summarise using MECs. These are commonly represented by a Completed Partially Directed Acyclic Graph (CPDAG), where directed edges have the same orientation across all DAGs in the class, and undirected edges indicate conflicting directions. This abstraction allows inference methods to capture shared structural features without committing to uncertain details.

While CPDAGs are widely used in score-based and hybrid methods, the PC algorithm, which is a constraint-based approach, uses a related structure called a Partially Directed Acyclic Graph (PDAG). PDAGs evolve during the algorithm's execution as independence tests refine causal links, especially around $v$-structures, where two parent nodes independently point to a common child, making them conditionally dependent when the child or its descendants are observed. Unlike CPDAGs, PDAGs are intermediate representations that may not correspond to any valid DAG structure. Their main limitation is that they cannot always be converted into a CPDAG representing a unique MEC, which poses difficulties when parameterising the graph into a BN; a challenge also encountered in this study (see subsection 5.1).

For a visual illustration of the concepts discussed above, see Figure 2 which presents three DAGs along with their corresponding PFAG and CPDAG structures. For a detailed introduction to causal or BN structure learning from data, see [15, 24, 25].

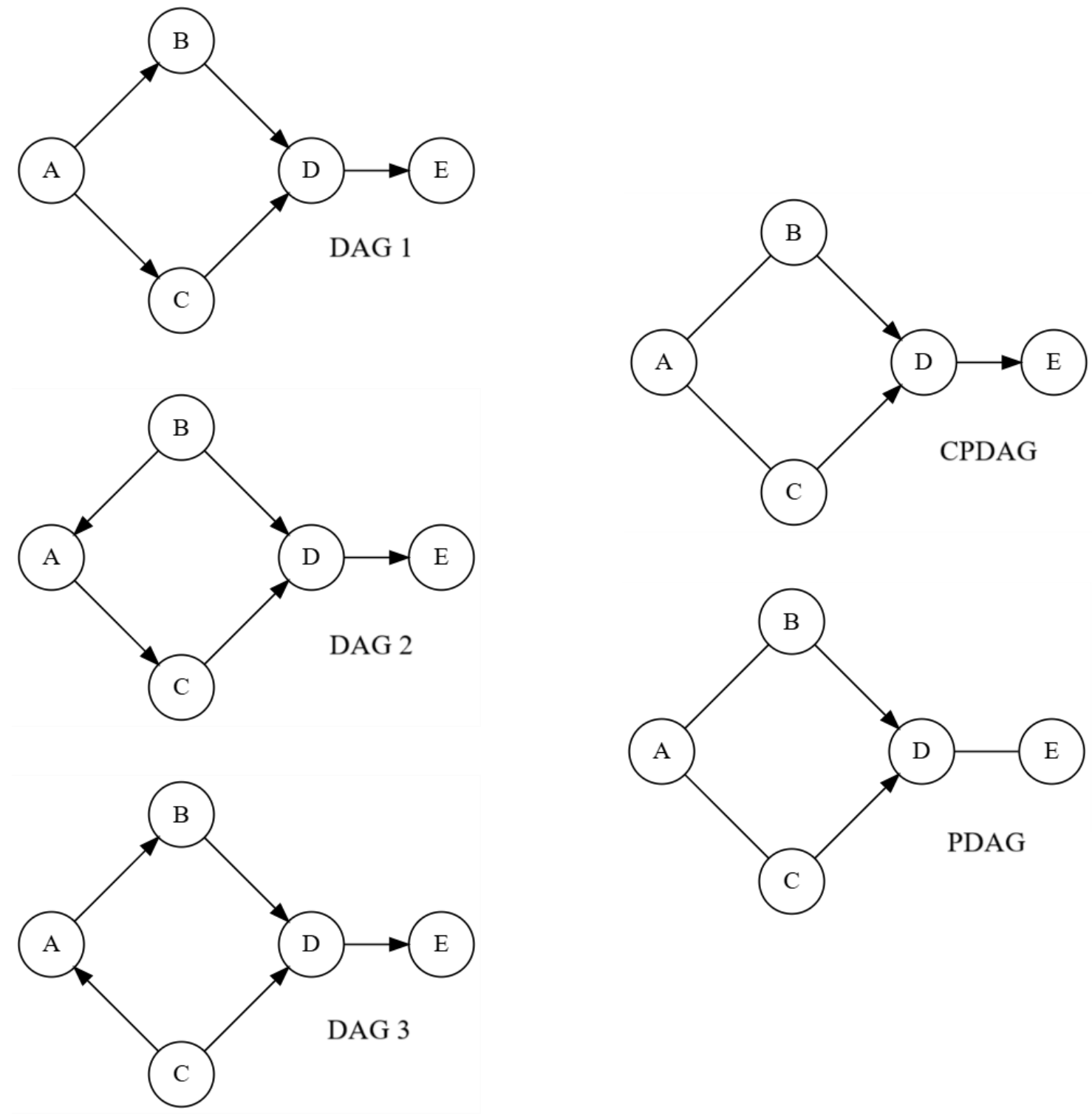

**Figure 2.** An illustration of the three DAGs that make the equivalence class CPDAG, and the corresponding PDAG containing $v$-structures only, based on an example in Verma and Pearl [26] and Kitson et al. [15].

## 3. Dataset and Methodology

The methodology is separated into two main parts. Firstly, subsection 3.1 describes how we use structure learning algorithms and apply them to a comprehensive dataset collated for this study, to learn and parameterise BN structures which we later use for predictive inference. Then, subsection 3.2 describes how we use domain expertise, as well as the game's manual and community knowledge, to fully specify two knowledge-based BN models.

### *3.1. Learning Bayesian networks from data*

We collated data via the Certified Hattrick Product Provider (CHPP) API [27], which is provided by Hattrick to support the development of third-party applications that enhance game experience. Data collection was therefore conducted in accordance with the platform's terms of service. We have carefully organised a dataset containing 250 variables and 1 million samples, where each sample represents a match between two teams. Since the dataset is based on in-game observations generated from real operation of Hattrick and by Hattrick's simulation engine during actual gameplay, it is automatically generated by the software and is therefore noise-free and complete.

The selected data samples were restricted to matches that met a minimum HatStats[1] rating threshold of 333, to ensure that the model is trained on competitive games played in higher divisions across countries. This threshold helps exclude lower-division matches, where teams are often focused on player training rather than competition, or where AI-managed (bot) teams may be present due to lower human participation. In other words, lower-rated matches, whether due to less skilled managers or teams in a 'training mode', tend to be passive rather than competitive, involve less tactical variation, and are therefore less valuable for model training and analytical purposes. In contrast, focusing on higher-rated matches ensures that the predictive models are trained to capture the dynamics of competitive, tactically rich games; rather than the passive, strategically irrelevant ones that are not of interest to model or analyse.

[1] HatStats is a widely-used in-game metric that estimates team strength based on selected ratings. It is calculated as: $\text{HatStats} = (3 \times \text{Midfield rating}) + \text{Attack rating} + \text{Defence rating}$.

Table A.1 provides a description of the variables. We make two versions of this dataset publicly available online through the Bayesys repository [28]; one with continuous and integer values for each variable, and another in which continuous variables are discretised into meaningful categories based on expert knowledge. In this study, we use the discrete dataset for structure learning. This is because all tested algorithms work with discrete data, but not all handle continuous data. Moreover, previous studies have shown that structure learning from mixed or continuous data often produces denser structures with spurious edges [29]. We therefore used discrete data to ensure consistency across all algorithms tested. Later, we use both discrete and continuous datasets to parameterise the learnt structures and convert them into BN models, learning both discrete and hybrid BNs, to capture all variable types in the case study, including continuous, integer, and categorical ones.

Further to the discussion in Section 2, structure learning algorithms are known to have scalability limitations. *Exact* learning algorithms, which guarantee the discovery of a network that maximises a given objective function, are often restricted to datasets containing only tens of variables. In contrast, *approximate* learning algorithms enable scaling up to hundreds of variables, and some can support thousands, but they typically achieve this by prioritising the recovery of sparse networks.

Because the data used in this study is large both in terms of the number of variables and sample size, it further exacerbates these scalability challenges. As a result, we introduce a 48-hour runtime limit for structure learning, per algorithm, and we focus on algorithms that prioritise efficiency, to maximise the probability to obtain a result within the runtime limit. We test multiple structure learning algorithms and disregard those[2] that fail to return a graph within the 48-hour time frame. Table 1 provides details of the algorithms ultimately used. These algorithms are described below.

**FGES:** The Fast Greedy Equivalence Search (FGES) algorithm, implemented in TETRAD, is a score-based method for learning causal structures from data [30]. It is a faster variant of GES by Chickering [31] that operates on MECs of DAGs, and uses a greedy approach to optimise a scoring function such as the BIC. FGES proceeds in two phases: a forward phase that adds edges to maximise the score, and a backward phase that removes redundant edges. It is efficient and scalable, making it suitable for high-dimensional datasets.

**GS:** The Grow-Shrink (GS) algorithm by Margaritis and Thrun [32] also traverses the search-space of MECs in a two-phase efficient approach. It starts with the Grow phase where the algorithm incrementally adds variables to the set of potential neighbouring nodes (parents and children) of each variable, and a variable is added if it shows a statistical association with the target, given the current set of neighbours. Once the initial set of neighbours is identified, GS enters the Shrink phase where it revisits each variable in the set and removes those that are conditionally independent of the variable under assessment, given the remaining neighbours. This step ensures that only the relevant variables are retained. At the end of these two phases, the GS algorithm identifies a minimal set of neighbours that are directly related to each variable, and construct the overall structure of the network.

**HC:** The Hill-Climbing (HC) algorithm is a score-based method for learning BN structures. One of the earliest works that describe this algorithm is the one by Heckerman et al. [22]. HC begins with an initial graph, which can take different forms such as an empty, random, or predefined graph (in this study, we assume an empty graph). It then iteratively improves its structure by making small changes, such as adding, removing, or reversing edges, to maximise an objective function such as the BIC. The process stops when no further score improvement is possible and returns a locally optimal structure. HC is the simplest and one of the most efficient structure learning algorithms, making it a popular choice.

[2] Some of the algorithms tested but did not complete learning within the 48-hour runtime limit include causal insufficiency algorithms such as FCI, GFCI and RFCI available in TETRAD, score-based MAHC available in Bayesys, and hybrid SaiyanH available in Bayesys.

**HC-Stable:** The HC-Stable algorithm is a variant of HC by Kitson and Constantinou [33, 34], designed to eliminate sensitivity to the arbitrary ordering of variables in the dataset. This is because traditional HC algorithms, such as the one described above, can produce different network structures when the order of the variables read from data changes, leading to instability in the learnt models. HC-Stable addresses this issue by ensuring that the search process is invariant to variable order, resulting in consistent and reliable network structures. This development is particularly important in applications where reproducibility and robustness are critical.

**PC-Stable:** The PC-Stable algorithm by Colombo and Maathuis [18] is a variant of the original PC algorithm by Spirtes et al. [17], designed to address the issue of order-dependence in constraint-based causal structure learning. In the standard PC algorithm, the outcome can vary depending on the order in which variables are processed during conditional independence tests. The PC-Stable variant mitigates, but does not fully eliminate, this by ensuring that the set of adjacencies remains consistent throughout the algorithm's execution, leading to more reliable and reproducible results, especially in high-dimensional settings.

**MMHC:** The Max-Min Hill-Climbing (MMHC) algorithm, by Tsamardinos et al. [23], is a hybrid structure learning algorithm that combines constraint-based and score-based approaches to balance efficiency and accuracy. It first identifies the parents and children of each variable using conditional independence tests through constraint-based learning, to reduce the search space of graphs, by focusing only on variables that are likely to be directly connected. MMHC then performs HC search on the restricted search space from the first phase to refine the structure, adding, removing, or reversing edges to maximise an objective function.

**TABU:** The TABU structure learning algorithm, with one of the earliest versions described by Bouchaert [21], is a score-based algorithm that improves upon traditional greedy search approaches, such as the HC algorithm described above, by incorporating a memory-based mechanism to escape local optima. It iteratively explores possible network structures, modifying edges (adding, removing, or reversing) to optimise an objective function. A *tabu* list tracks recently visited solutions to prevent revisiting them, enabling the algorithm to explore a broader search space. This approach allows temporary acceptance of suboptimal moves, facilitating escape from local optima and increasing the likelihood – relative to HC - of finding a globally optimal structure. While less efficient than HC, TABU search remains particularly effective for large datasets and complex networks due to its flexibility and efficient exploration.

**Table 1.** Structure learning algorithms that completed within the 48-hour runtime limit were used to learn BN structures. Implementations were sourced from the bnlearn R package [35], the TETRAD causal discovery software [36], the Bayesys BN structure learning tool [37], and author-specified repositories [34]. Runtime values reflect the computational time-complexity of the specific implementations.

| Algorithm | Learning class | Package | Runtime (hours) |
|---:|:---:|:---:|:---:|
| FGES | Score-based | TETRAD | 1.6 |
| GS | Score-based | bnlearn | 2.7 |
| HC | Score-based | Bayesys | 2.7 |
| HC-Stable | Score-based | Author code | 2.3 |
| MMHC | Hybrid | bnlearn | 0.7 |
| PC-Stable | Constraint-based | bnlearn | 10.4 |
| TABU | Score-based | Bayesys | 4.5 |
| TABU-Stable | Score-based | Author code | 2.3 |

**TABU-Stable:** The TABU-Stable algorithm by Kitson and Constantinou [33, 34] is an improvement of the traditional TABU described above, designed to address the same issue of variable ordering sensitivity as HC-Stable does for HC. In standard TABU search, the sequence in which variables are processed can influence the resulting network structure, leading to inconsistencies. TABU-Stable eliminates this sensitivity by implementing strategies to ensure the learning process remains invariant to variable order, producing reliable and reproducible structures.

### *3.2. Constructing Bayesian networks with expert knowledge*

In addition to learning BN structures and parameterising their conditional distributions using structure learning algorithms, we also construct two knowledge-based BN models informed by existing knowledge. These models integrate information from both the official manual and community-driven resources, which offer insights not formally documented but inferred through player experience and analysis over the years. Specifically, due to the hidden game mechanics of Hattrick, the community has been analysing its game-engine for decades, leading to the development of the *Unwritten Manual*, a continuously revised resource since 2011. The knowledge-based BN models are built using the expertise of two authors of this paper who together have 39 years of experience playing the game, the official game manual which provides a high-level overview of the game-engine, and lastly the *Unwritten Manual* which offers in-depth user-compiled insights.

The two knowledge-based BN models are hybrid BNs, meaning they capture both discrete and continuous variables, as required to accurately represent the structural relationships in the Hattrick data. The game mechanics involve a mix of variable types. For example, team sector ratings are continuous, while tactical choices and match outcomes are discrete. To jointly model these different variable types within a unified probabilistic framework, a hybrid BN is necessary. This approach aligns with the example in Section 2.1 (see Figure 1), where interactions between mixed variable types are captured within a single graphical model. The two knowledge-based BN models are:

a. **KB-regression:** Many of the relationships within the KB-regression model are defined using non-linear regression and other equations derived from the Hattrick community's accumulated knowledge as detailed in the *Unwritten Manual*, which has developed over decades of gameplay and analysis.

b. **KB-probabilistic:** This is a hybrid BN incorporating both discrete and continuous variables, as detailed in Table A.1. This model builds on the KB-regression approach by retaining known game rules and verified information but replaces the community-derived regression equations from the *Unwritten Manual*, with conditional probability distributions learnt from data. These are modelled using the Beta-Binomial framework, which better aligns with BN principles and is well suited for outcomes based on repeated trials (e.g., goals scored, chances converted), where the success probability is uncertain.

   In Hattrick, many outcomes (e.g., turning midfield control into scoring chances) can be viewed as probabilistic results of multiple latent attempts. The Beta-Binomial model captures this by representing both the frequency and uncertainty of such events, allowing for more flexible and realistic conditional dependencies than deterministic, fixed-form equations. This approach maintains consistency with probabilistic modelling, improves interpretability, and better reflects the game's simulated dynamics.

   While the BN models learnt from data contain 250 nodes, the KB-probabilistic model contains 363 nodes, and this increase arises purely from its hybrid nature and different types of variables. For example., each Beta-Binomial representation consists of two nodes, one for Beta and one for Binomial, and this increase in the number of nodes allows for a more detailed simulation without incorporating additional data beyond that used in the data-driven models. A technical description is provided in Appendix B, and its decision-making benefits are illustrated in subsection 6.4.

Table 2 provides a concise summary of the input (observed) and output (inferred) variables of interest. An input variable is one that the user can control, such as the attack rating of their team (usually at the expense of something else), or the number players of a particular speciality to field. An output variable, on the other hand, is a possible outcome of the game, such as attacks generated or goals scored, determined by the input values. Specifically, the KB-probabilistic model incorporates the following variable types:

a. **Discrete variables:** with a finite set of possible values learnt from data.
b. **Continuous variables:** Beta or Binomial distributions used for probabilistic representations of Beta-Binomial processes. The Beta distribution is learnt from data and serves as the parameter $p$ of a Binomial distribution. The parameter $n$ in a Binomial distribution often represents the number of attacks or goals, with prior attack parameters informed by the game's manual or supplementary information verified by game developers.
c. **Equation:** Defines relationships between variables using mathematical formulas derived from in-game manuals, as described in the subsections that follow within this section.
d. **Uniform:** Used exclusively for input (observed) variables, assuming all values within a given range are equally likely.

**Table 2.** A concise summary of the observed input and estimated output variables of interest for both teams. A detailed list and description of these nodes is provided in Table A.2. The input (observed) and output (inferred) variable sets are discussed in Section 4.

| Variable sets | Variable summary | Description |
| --- | --- | --- |
| Attack/Defence inputs | Left, middle, and right attack ratings. Left, middle, and right defence ratings. | Attack and defence ratings for each sector, used to determine the probability of scoring an attack from a given section, based on the corresponding defence rating of the opposing team. |
| Midfield inputs | Midfield ratings. | Ratings that determine the probability of a generated attack being allocated to a team. |
| Set-pieces inputs | Indirect set-piece attack and defence ratings. | Ratings for set-piece attacks and defences, used to determine the probability of scoring from set-pieces. These probabilities are divided into different types of set-pieces and are partially influenced by the defence ratings of the opposing team. |
| Tactic inputs | Tactic and tactic skill. | The playing style chosen from seven possible tactics, and the associated skill level in executing that tactic. |
| Specialities inputs | Players with *Unpredictable* (offensives, special action, long pass, mistake, and own-goal player positions), *Quick* (offensive and defensive positions), Technical (offensive and defensive positions), and Head (corner counts and defensive positions) specialities. | The number of players with *Unpredictable*, *Quick*, *Technical*, and *Head* specialities in different positions, who may trigger either positive or negative special events. |
| PNF/PDIM inputs | Powerful Normal Forwards (PNFs). Powerful Defensive Inner Midfielders (PDIMs). | PNFs can generate an extra attack immediately after a missed attack, while PDIMs can press an opponent's attack. |
| Counterattack goals outputs | Counterattack goals | Goals scored in each sector or from set-pieces as a result of a counterattack. |
| PNF goals outputs | PNF goals | Goals scored through a PNF event. |
| Special event goals outputs | Special event goals | Goals scored through special events. |
| Normal goals outputs | Normal goals | Goals scored from standard attacking opportunities, excluding counterattacks, special events, and PNF events. |
| Total goals output | Goals scored | The total number of goals scored in the match. |
| HDA output | HDA (Home, Draw, Away) | The match outcome in terms of a home win, draw, or away win. |

The next section focuses on describing the KB-probabilistic model which, as we later demonstrate in Section 5, outperforms not only the KB-regression model but also all BN models learnt from data. The KB-regression model, which we developed based on existing equations specified by the Hattrick community in the *Unwritten Manual*, is less central to this study, whose focus is on the KB-probabilistic model which incorporates verified information, with the remaining relationships learnt from data. Therefore, the functions of the KB-regression model that come from the *Unwritten Manual*, and which are not adopted by the KB-probabilistic BN, are outlined in Appendix C.

## 4. Review of the Hattrick mechanisms as learnt by the main, KB-probabilistic, BN model

This section provides expert review and validation of the Hattrick mechanisms learnt by the main knowledge-based model (KB-probabilistic), comparing against known game rules and developer-released documentation, conducted by two authors of this paper who collectively have 39 years of experience playing the game and engaging with the Hattrick community. The section begins with an overview of the model, followed by expert reviews of how the model simulates attacks (Section 4.1), goal scoring (Section 4.2), match tactics (Section 4.3), and other special match events (Section 4.4). This section is written primarily for readers familiar with, or who play, the game. General readers may skip to the next section without loss of continuity.

While many of these mechanisms are known to the Hattrick community, some deviate from those specified in the *Unwritten Manual*, and these discrepancies may be explained by the structural inference approach used in models such as BNs. Where this is the case, further details are provided either in the main text or in a footnote. For clarity, we summarise these mechanisms in subsections that are of particular interest to the readership. Because Hattrick is a complex game, as reflected in the extensive KB-probabilistic model containing 363 nodes, we use simplified notation to describe the mechanisms, rather than referring to the actual node names in the BN model. The model, which was manually constructed using the BN software GeNIe [13], a commercial BN software available for free academic use, is detailed in Appendix A. It is also made publicly available via the Bayesys repository [28]. A screenshot of the KB-probabilistic BN model is provided in Figure 3. The subsections that follow describe how the model simulates attack and defence mechanisms, the influence of match tactics on the game-engine, and the generation of special events.

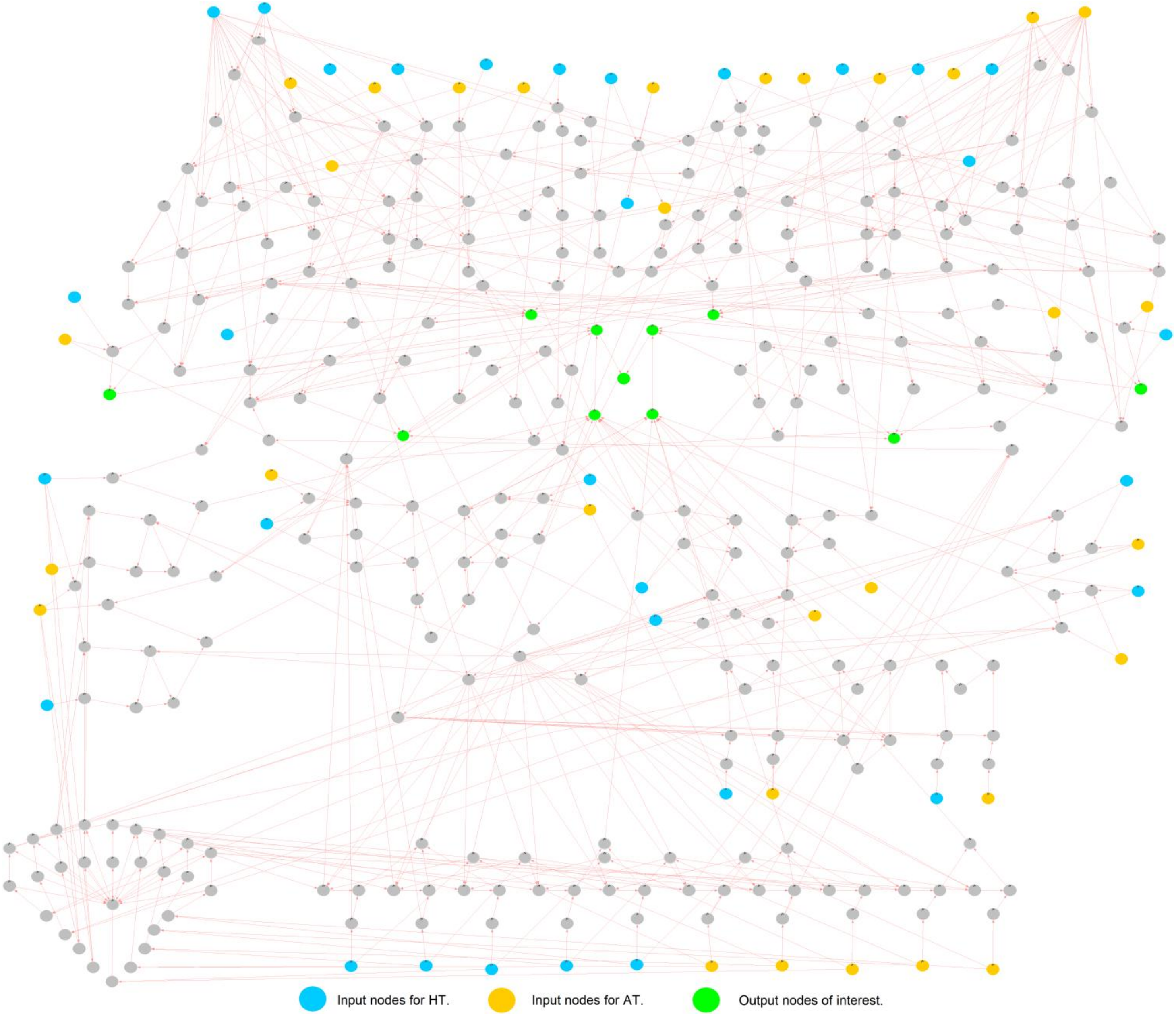

**Figure 3.** The knowledge-based graphical structure of the KB-probabilistic BN model, containing 363 nodes learnt from data, visualised in the GeNIe BN software.

### *4.1. Generating attacks*

The first key mechanism of the knowledge-based model involves capturing how scoring opportunities are generated. An attack can only be generated if a team wins possession of the ball, and team possession depends on the midfield inputs described in Table 2. The difference in ratings (i.e., defence, attack, and midfield) between teams has a non-linear effect on performance. We adopt Equations 1, 2, and 3 from the *Unwritten Manual* which, although never officially confirmed, are thought to be correct given that they match the probabilities generated by Hattrick's in-built simulator, commonly used by players to plan against opponents' past rating performances, in estimating (a) the probability of winning an attack (i.e., possession), (b) the total number of potential attacks, and (c) the probability of scoring from an attack. Specifically, to win an attack, we assume:

$$POS_{HT} = \frac{(R_{HT}^{M} \times 4 - 3)^3}{(R_{HT}^{M} \times 4 - 3)^3 + (R_{AT}^{M} \times 4 - 3)^3} \tag{1}$$

where $R_{HT}^{M}$ and $R_{AT}^{M}$ are the midfield ratings for the home team and the away team respectively. The possession for AT is the complement of HT, $POS_{AT} = 1 - POS_{HT}$.

The official game manual states that each team receives five exclusive attack chances, with an additional five shared chances that both teams compete for. The POS probability determines whether an attack is allocated to a team. If a team fails to secure an exclusive attack chance, the opportunity is lost. However, if a team fails to secure a shared chance, the opportunity is allocated to the opposing team. The game-engine generates a maximum of 15 *normal* attacks, with a maximum of 10 allocated to a single team, comprising 5 exclusive ($E$) attacks and up to 5 shared ($S$) attacks. Equation 2 specifies how these attacks are generated:

$$NM_{HT} = E_{HT} \sim Binomial(5, POS_{HT}) + S_{HT} \sim Binomial(5, POS_{HT}) \tag{2}$$

where $NM$ represents a *normal* attack ($NM_{HT}$ for HT in this example), 5 is the number of trials $n$ in the Binomial distribution, and $POS_{HT}$ is the probability of success $p$ in the Binomial distribution for generating an attack. For $AT$, the probability of success in $E_{HT}$ is $1 - POS_{HT}$, while $S_{AT} = 5 - S_{HT}$.

When a team is allocated an attack, it can be generated on the left, middle, or right, or as a set-piece. A set-piece attack is further classified into a Direct Free-Kick (DFK), an Indirect Free-Kick (IFK), or a Penalty Kick (PK). Equation 3 specifies how these attacks are distributed for both teams in the absence of tactics:

$$AD = \begin{cases} \sim Binomial(NM, 0.2565), & \text{Left} \\ \sim Binomial(NM, 0.3615), & \text{Middle} \\ \sim Binomial(NM, 0.2565), & \text{Right} \\ \sim Binomial(NM, 0.0586), & \text{DFK} \\ \;\; Binomial(NM, 0.0418), & \text{IFK} \\ \;\; Binomial(NM, 0.0251), & \text{PK} \end{cases} \tag{3}$$

where $AD$ is Attack Distribution, $NM$ is the distribution of attacks generated by Equation 2, and $p$ is the probability for each type of attack, summing up to 1 in $AD$. The values for $p$ in Equation 3 are derived from data, and slightly differ from the probabilities stated in the official manual, although are consistent with those stated in the *Unwritten Manual*[3].

---

[3] The official game manual specifies that an attack is generated on the left, middle, right, or as a set-piece with probabilities of 0.25, 0.35, 0.25, and 0.15, respectively. However, those reported in Equation 3 are found to be closer to those stated in the *Unwritten Manual* and the in-game forum post [post=17572304.15].

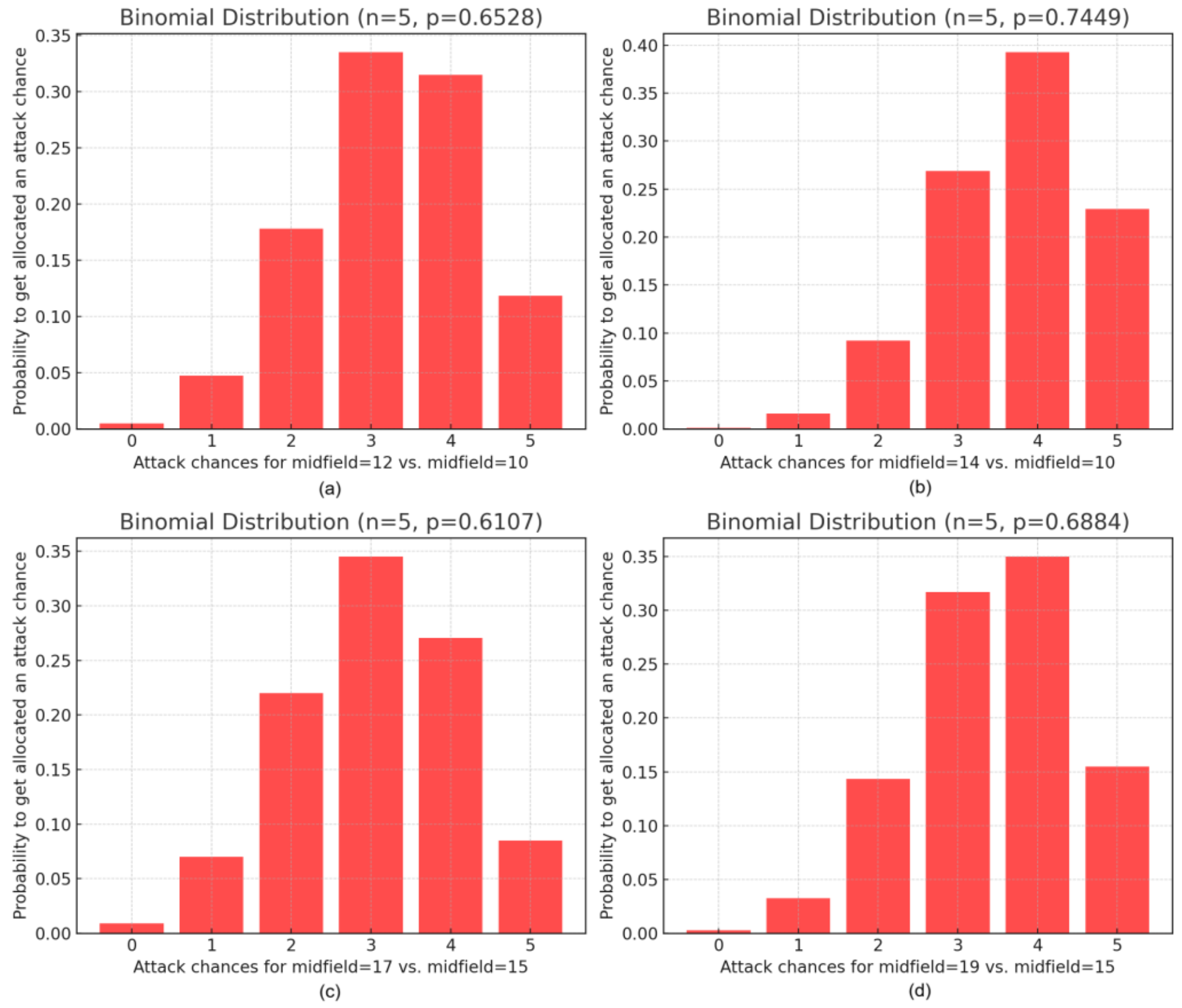

**Figure 4.** Binomial distributions of attack chances generated for each of the four sets of competing midfield ratings specified. The illustration demonstrates that the same rating difference has a more pronounced effect at lower midfield ratings, as described by Equation 1.

Figure 4 illustrates how the Binomial distribution of chance allocation shifts with small differences in midfield ratings. It shows that, and with reference to Equation 1, these rating differences have a more pronounced effect at lower midfield ratings than at higher ones. Figure 5 also presents the fragment of the KB-probabilistic BN model that captures the relevant variables used to compute the attack Binomial distributions, with the midfield rating values corresponding to the example shown in Figure 4b.

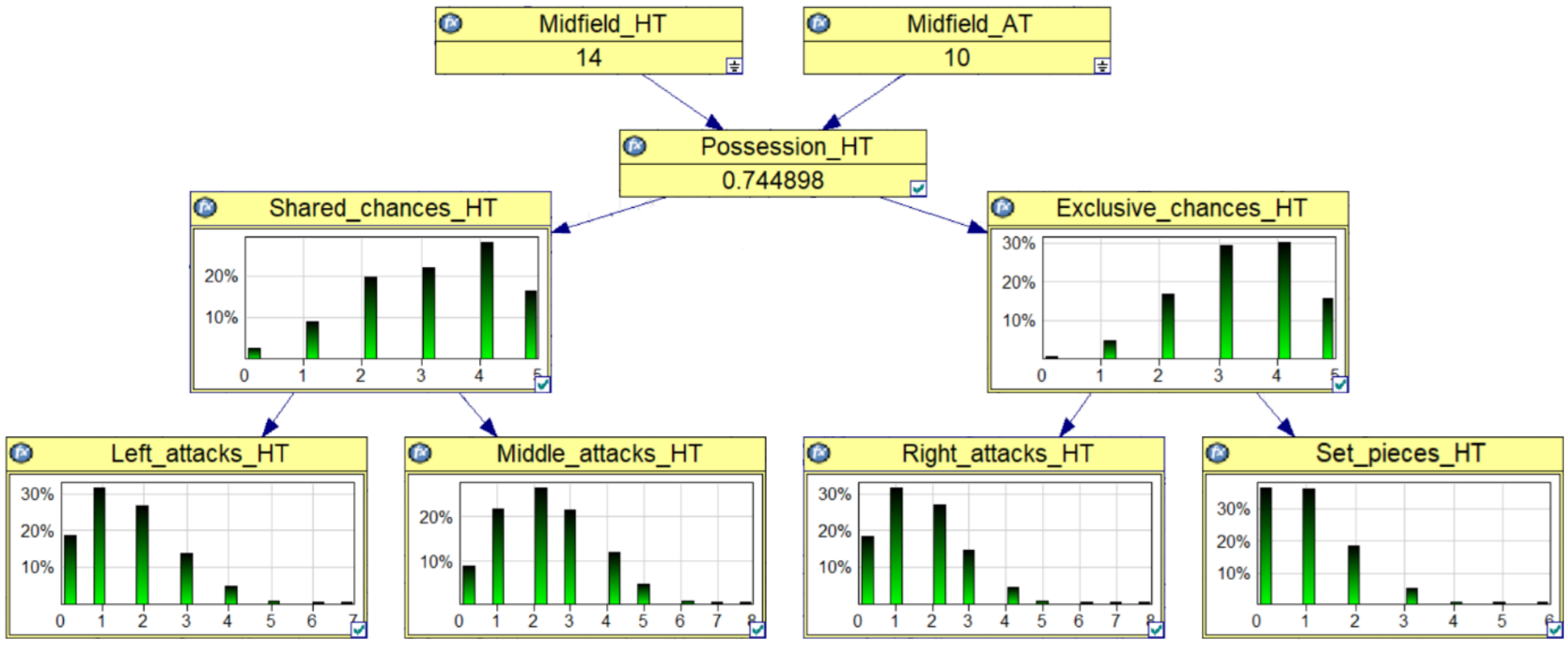

**Figure 5.** The BN fragment that models the allocation of the distribution of attacks across different sectors or as set-pieces, based on the chance distribution in Figure 4b.

### *4.2. Scoring attacks*

Scoring an attack involves comparing the attacking team's attack rating for a given sector with the corresponding defence rating of the opposing team, and relates to the attack/defence and set-pieces inputs in Table 2. Equation 4 presents a well-established Hattrick function that specifies the probability of scoring an attack from the left, middle, or right, as derived from the *Unwritten Manual* forum [post=17372624.15]:

$$SCR_{LMR_{HT}} = \frac{0.92.\left(R_{HT}^{A} \times 4 - 3\right)^{3.5}}{(R_{HT}^{A} \times 4 - 3)^{3.5} + (R_{AT}^{D} \times 4 - 3)^{3.5}} \quad (4)$$

where SCR represents the probability to score an attack from the Left, Middle or Right (LMR), $R_{HT}^{A}$ denotes the home team's attack rating, and $R_{AT}^{D}$ the corresponding away team's defence rating. The same concept applies when the $AT$ is attacking and the $HT$ defending. Specifically, attack and defence ratings always form opposing pairs: right attack versus left defence, left attack versus right defence, and central attack versus central defence. Figure 6 (left) illustrates the probability of being allocated an attack based on Equation 1, while Figure 6 (right) shows the probability of scoring an attack from LMR based on Equation 4, across varied possession and attack ratings, assuming a fixed opponent rating of 15 for both possession and defence.

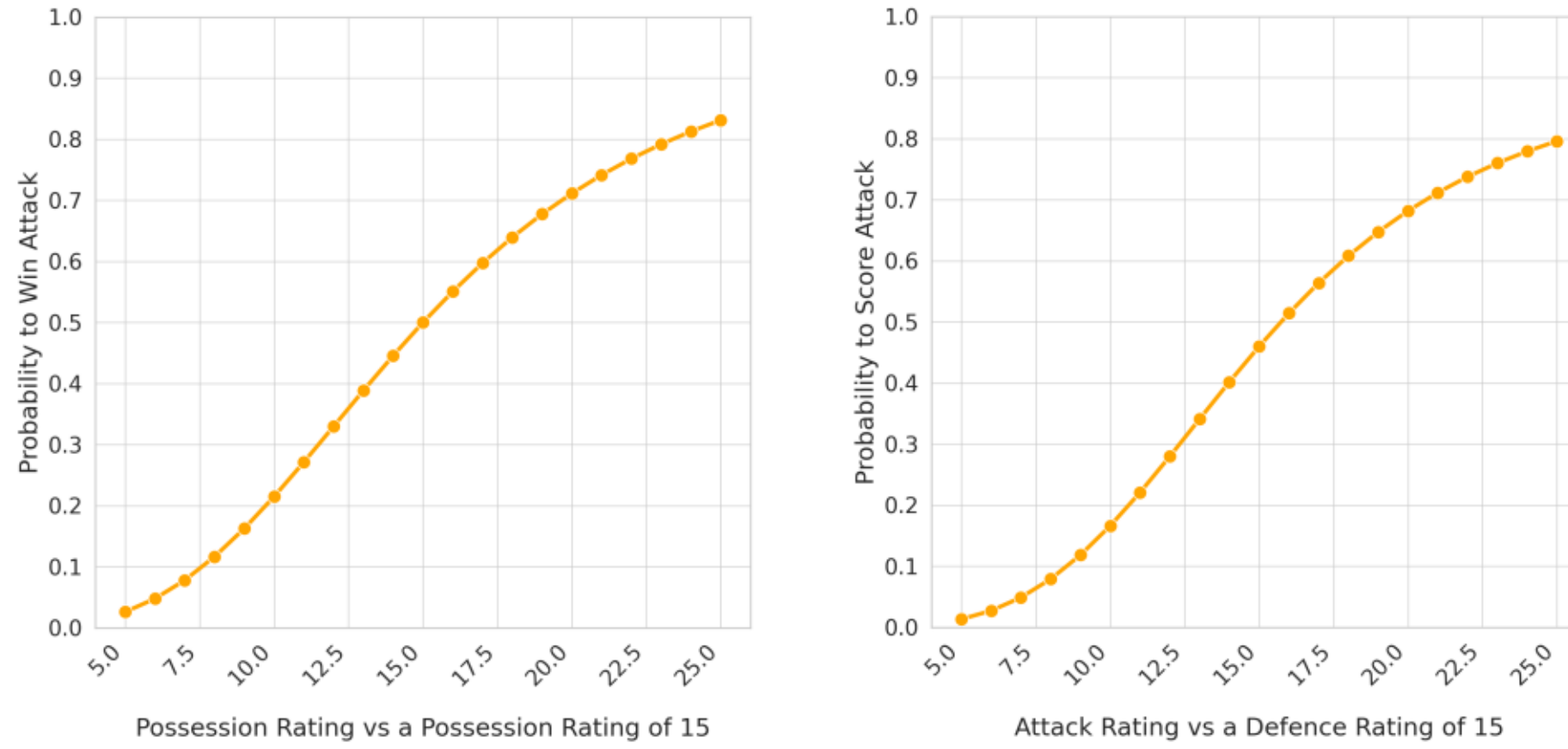

**Figure 6.** The probability of being allocated a generated attack as a function of midfield vs midfield ratings (left chart) based on Equation 1, and the probability of scoring an attack as a function of attack vs defence ratings (right chart) based on Equation 4.

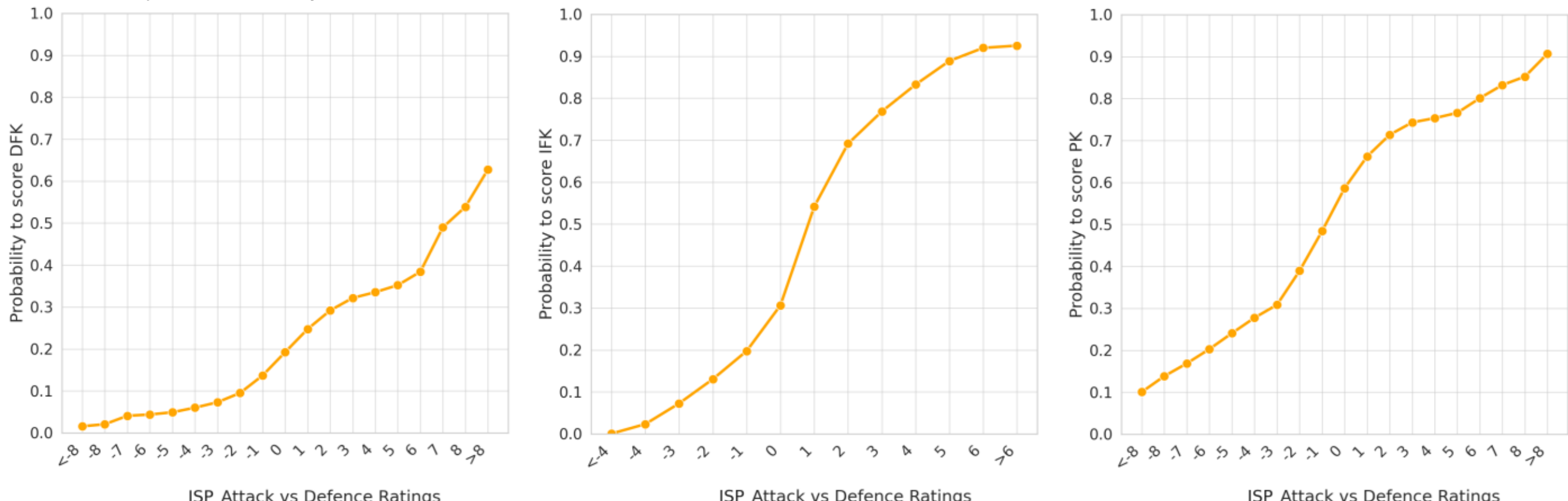

**Figure 7.** The probability to score a Direct Free Kick (DFK; left chart), an Indirect Free Kick (IFK; middle chart), and a Penalty Kick (PK; right chart), based on the difference between ISP attack vs ISP defence ratings.

This attack and defence concept extends to set-pieces, with different scoring probabilities depending on the type of set-piece. Figure 7 illustrates how these probabilities are derived from data, based on the difference between attack and defence ISP ratings, which function similarly to the LMR attack and defence ratings for open play. The figure includes scoring probabilities for DFKs, IFKs, and PKs. Unlike Figure 6, the relationship between rating differences and scoring probability is less smooth, as set-piece outcomes often depend on additional hidden player-specific parameters not captured in the data. One such factor is the set-piece skill of the taker, which remains unknown since individual player skills are not visible to other users and therefore cannot be recorded in the dataset.

### *4.3. Simulating match tactics*

There are seven different match tactics, including the *Normal* tactic, which assumes the default game-engine parameters as described in Subsection 4.1. This subsection relates to the tactic inputs in Table 2. The other six tactics are described below.

Attack in the Middle (AiM): Exchanges 20-35%[4] of wing attacks to the middle, depending on tactic skill, while applying a small penalty to central defence ratings. Figure 8 (left) illustrates the conversion rate assumed by the KB-probabilistic model between tactic rating and attack distribution rates. Playing AiM increases the proportion of attacks through the middle to 47–55%, depending on tactic level, compared to 36.15% when using the *Normal* tactic (see Equation 3).

Attack on the Wings (AoW): Exchanges 34-52%[5] of central attacks to the wings, depending on tactic skill, while applying a small penalty to wing defence ratings. Figure 8 (right) presents the conversion rate assumed by the KB-probabilistic model, between tactic rating and attack distribution rates. Playing AoW increases the share of wing attacks to 63–70%, depending on tactic rating, compared to 51.3% with the *Normal* tactic (see Equation 3).

Counterattack (CA): Reduces the midfield rating by 7% in exchange for the ability to generate a counterattack for each missed Normal chance by the opponent, provided that the team playing CA has a lower midfield rating than their opponent before the 7% penalty is applied. Counterattacks can occur through the left, middle, right, or from set-pieces, but not as a penalty kick (PK). Figure 7c presents the conversion rate assumed by the KB-probabilistic model given the tactic rating. It shows that playing CA increases the proportion of the opponent's missed chances converted into counterattacks to 4–45%[6], up from the non-tactical counterattack rate of 3.8% when using the Normal tactic. The non-tactical counterattack rate is sensitive to the number of defenders, with the model assuming rates of 1.75%, 3.63%, 6.04%, and 8.03% when playing with 2, 3, 4, or 5 defenders, respectively, as learnt from data.

Long shots (LS): exchanges 6-43%[7] of LMR attacks to long shots in exchange for a small penalty to attack and midfield ratings. Figure 9 (left) illustrates the relationship assumed by the KB-probabilistic model between tactic rating and the conversion rate of LMR attacks into long shots. Figure 9 (right) shows the relationship between scoring a long shot and the difference between tactic rating and the opponent's average defence rating, which is used as a proxy for the opponent's hidden goalkeeper's skill, with the scoring rate ranging from 11% to 100%.

[4] This range differs slightly from the conversion rate range reported in the official game manual as 15-30%, and in the *Unwritten Manual* as 30-40% based on forum post 17493341.20.

[5] This range differs from the conversion rate range reported in the official game manual as 20-40%, and slightly from that reported in the *Unwritten Manual* as 40-55% based on forum post 17493341.20.

[6] This conversion rate range is consistent with that reported in the *Unwritten Manual* based on forum post 17493341.6

[7] This moderately differs from the rate of 5-33% reported in the *Unwritten Manual*, forum post 17493341.8.

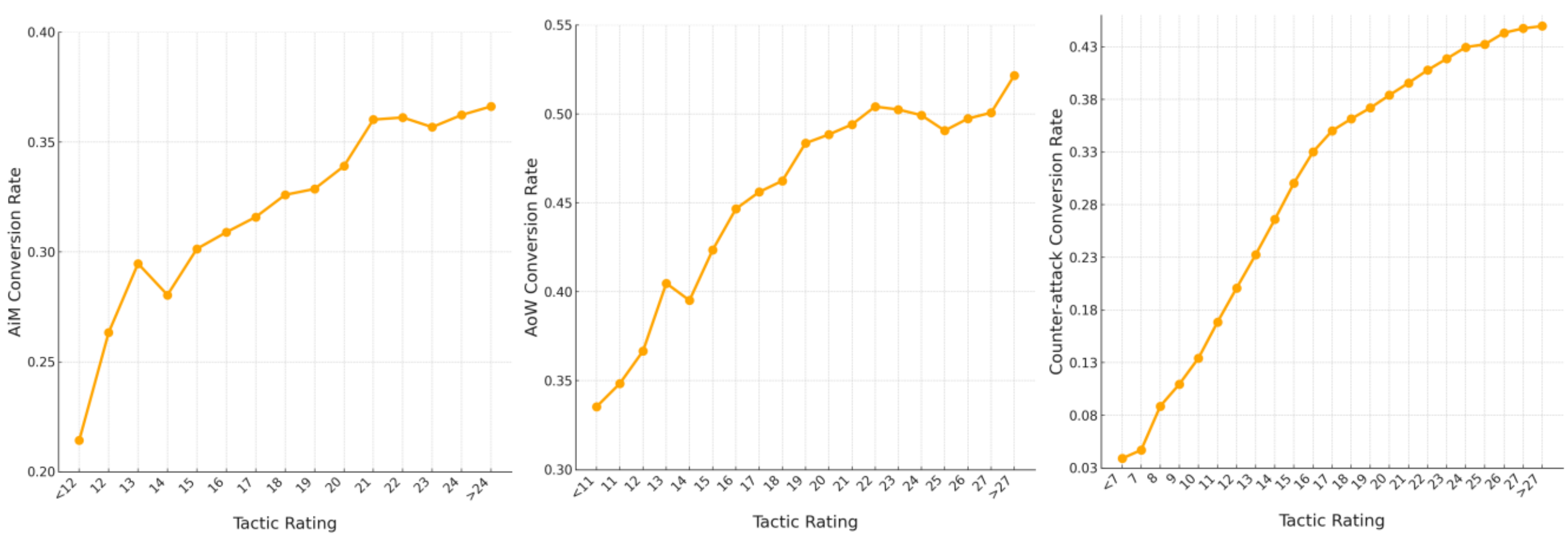

**Figure 8.** Conversion rates for tactics: Attack in the Middle (AiM), shifting wing attacks to the middle (left chart); Attack on the Wings (AoW), shifting middle attacks to the wings (middle chart); and Counterattack (CA), converting an opponent's missed normal chance into a counterattack (right chart).

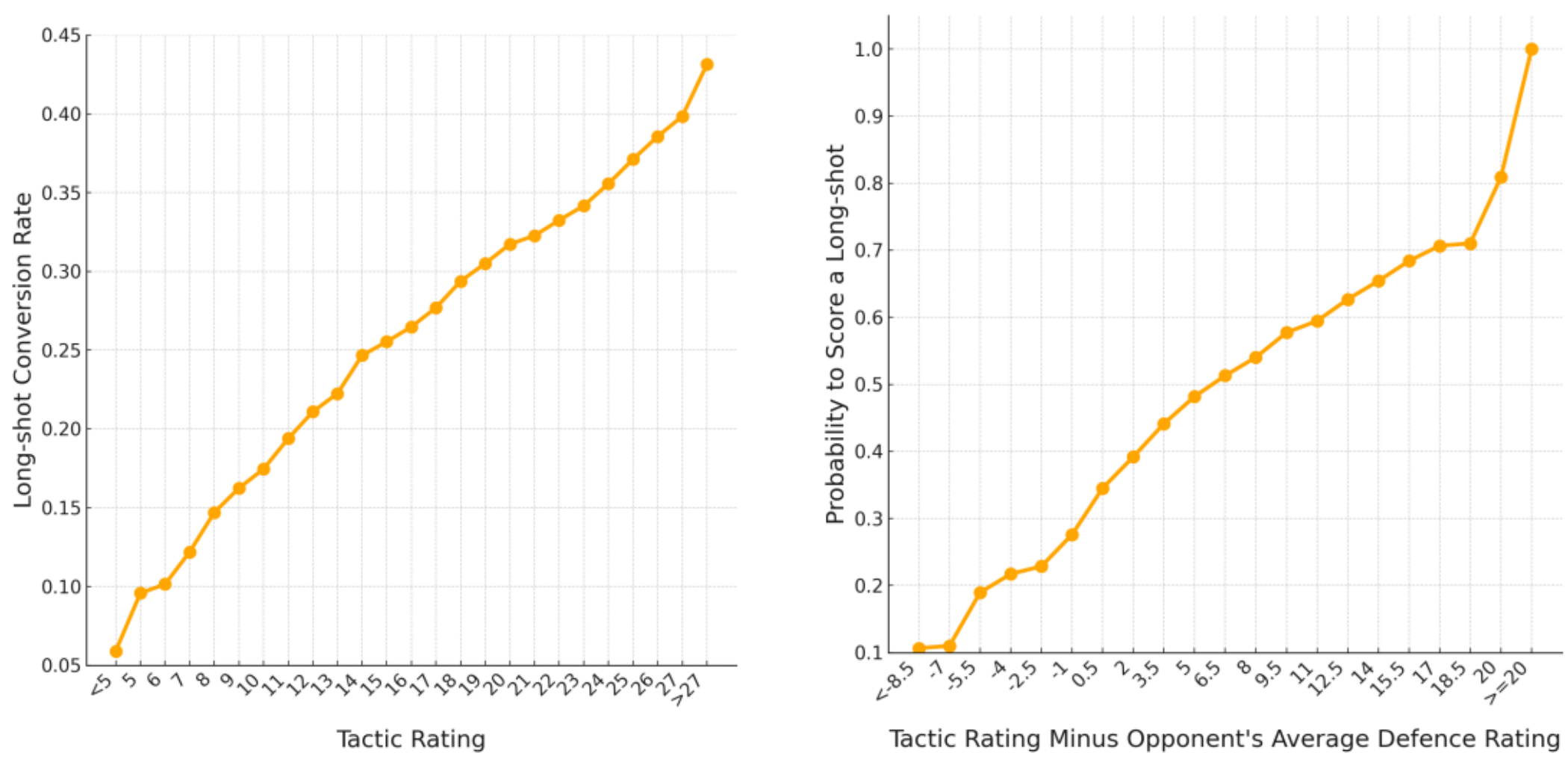

**Figure 9.** Conversion rate for the Long Shots (LS) tactic, showing the probability to convert an LMR attack into a long-shot (left chart) based on the tactic rating, and the probability to score a long shot (right chart) as a function of the tactic rating minus the opponent's average defence rating.

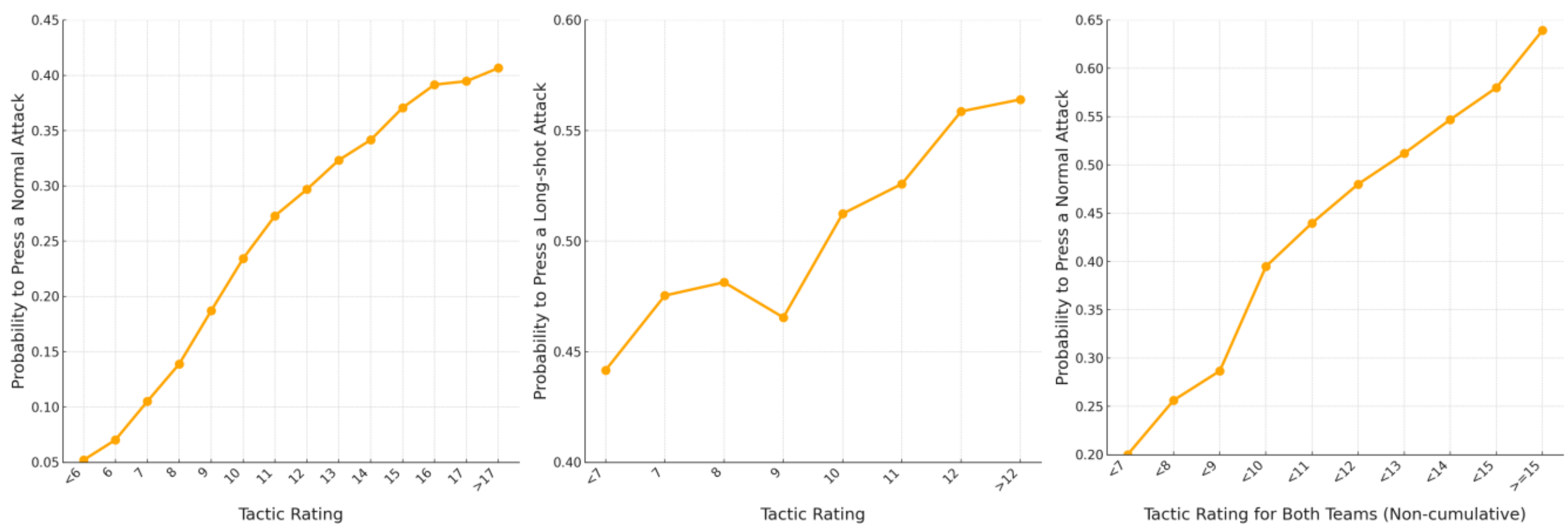

**Figure 10.** Conversion rate for Pressing (PS) when pressing a *Normal* attack (left chart), a long-shot attack (middle chart), and a normal attack when both teams play PS (right chart). The non-cumulative intervals of the $x$-axis in the right chart do not incorporate data from previous intervals.

Pressing (PR): Reduces the number of attacks generated by both teams. As illustrated[8] in Figure 10, 5%-41% of *Normal* attacks are suppressed when one team is playing PS (left chart), depending on the tactic level. When a team using PS faces an opponent playing LS, 44–66% of long-shot attacks are suppressed (middle chart). Additionally, when both teams play PR, 20–64% of *Normal* attacks are suppressed (right chart).

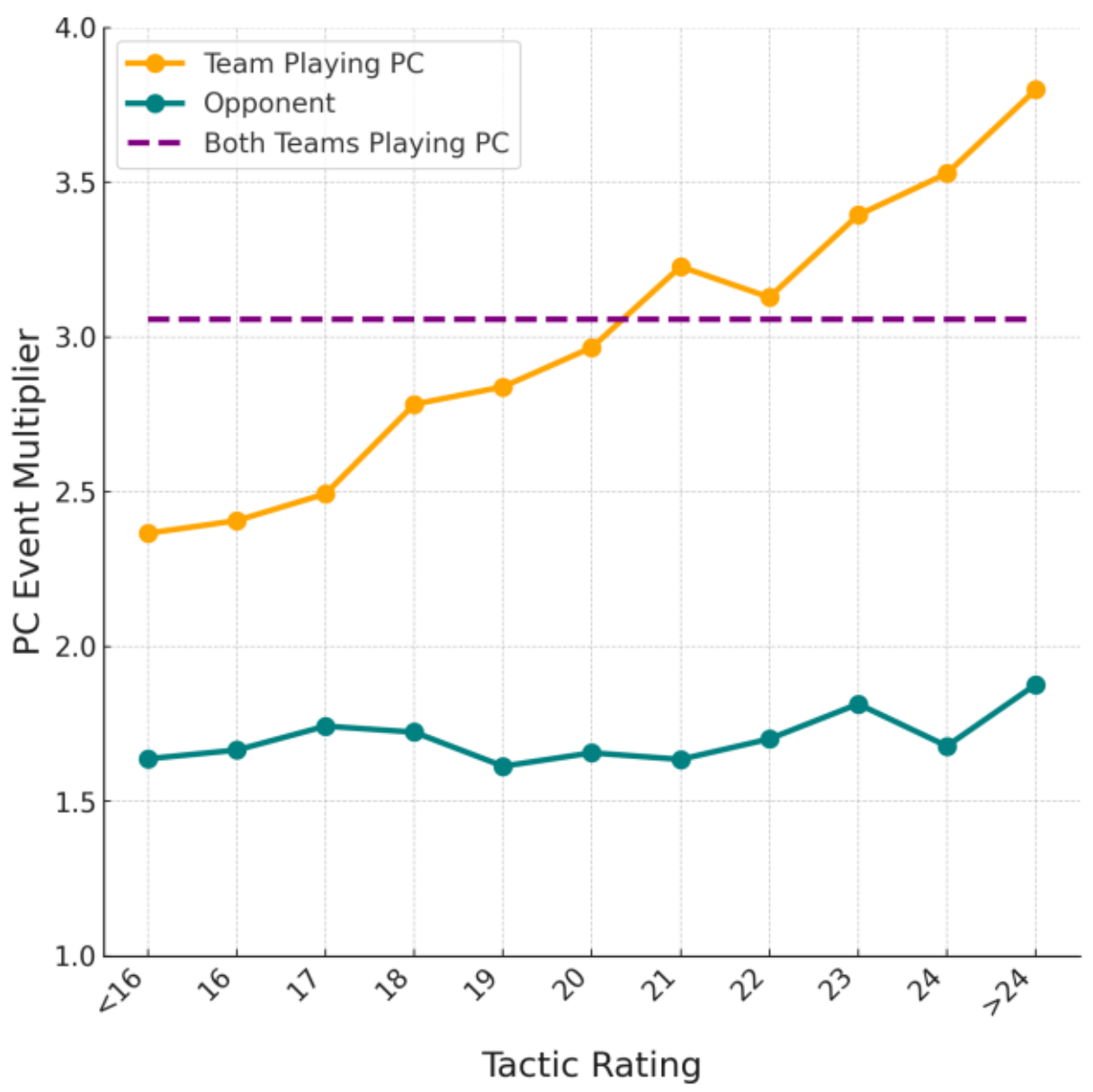

**Figure 11.** Multiplier event rate for the Playing Creatively (PC) tactic, assuming an average of 0.42 special events per team when using the default *Normal* tactic. To adjust for event bias, results are constrained to matches where all possible player specialities capable of triggering a special event appear at least once per team (excluding the player *Support* speciality, which is not considered in this study; see subsection 4.4).

Play Creatively (PC): This tactic influences one of the most complex aspects of the game-engine; the simulation of special events, which is detailed in subsection 4.4. Playing PC increases the maximum possible number of special events per match by two[9] events and hence, the average number of events occurring for both teams, although the team using this tactic benefits more, as illustrated in Figure 11. Assuming that the average number of special events per team is 0.42, as learnt from data when using the default *Normal* tactic in matches that include all possible player specialities, a team playing PC experiences an increase of 2.37× (i.e., 2.37 times more) to 3.8× in its average number of special events, depending on tactic level. Meanwhile, the opponent – not playing PC - sees a smaller increase of 1.63× to 1.88×, though the relationship between tactic skill and this effect remains uncertain. Furthermore, when both teams play PC, the increase in special events is estimated at 3.05× for both teams, regardless of tactic level, due to sample limitations.

[8] The conversion rates for this tactic are not directly comparable with the information provided in the *Unwritten Manual*, forum post 17342010.653 onwards, as the manual reports the average number of attacks pressed rather than the conversion rate.

[9] According to the *Unwritten Manual*, the maximum number of possible special events increase by one; but in this study we assume by two as learnt from data.

### *4.4. Simulating special and other events*

The mechanisms described in this subsection relate to the specialities and PNF/PDIM inputs in Table 2. Special events are divided into player-based and team-based events. Player-based special events require players to possess a specific speciality to trigger an event related to that speciality, whereas team-based special events generally do not require players to have a special skill. A player may have no speciality or one of the seven available specialities, each capable of triggering a different event. However, two[10] of these specialities are considered non-beneficial by the game community and are rarely used by players. For simplicity, we focus only on the five commonly used specialities.

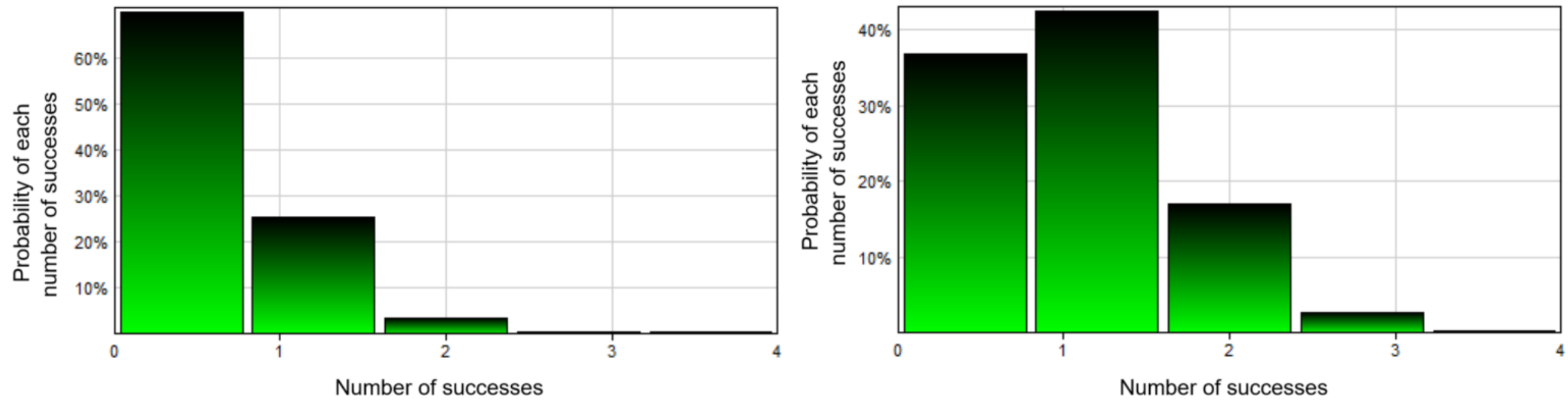

**Figure 12.** $\sim Binomial(n, p)$ distributions in the KB-probabilistic BN model for nodes Team-based special events (left) and Player-based special events (right), where $n = 5$ and $n = 4$, and $p = 0.372$ and $p = 0.841$, respectively. The charts are taken from the actual BN model, developed using GeNIe BN software [13].

Table 3 outlines the positive and negative events that each team-based special event or player-based special skill can trigger. Much of this information is derived from the Hattrick developer blog by HT-Tasos [38] and a series of related forum posts[11], with some modifications made due to BN modelling choices or data inconsistencies. The KB-probabilistic BN model implements the special events described in Table 3 and incorporates key assumptions derived from the *Unwritten Manual*. Hyperparameters are learnt from data constrained to matches where both teams play the default *Normal* tactic and include players with all available specialities and player positions capable of generating these events.

Up to five team-based and four player-based events can be generated in a match, with an average of 0.372 and 0.841, respectively. In the KB-probabilistic BN model, the simulation of these events is represented as a $\sim Binomial(n, p)$ where $n = 5$ and $n = 4$ for team-based and player-based events, respectively, and $p = 0.372$ and $p = 0.841$, as illustrated in Figure 12.

---

[10] The two specialities excluded from this study are: (a) Regainer, which enables players to recover from injuries twice as fast, and (b) Support, which, when triggered, provides a 10% performance boost to neighbouring players but also carries the risk of triggering a negative event known as 'team disorganisation', resulting in a decline in overall team performance.

[11] This information is based on in-game forum posts 17192231.1 and 17192231.3 for team-based events, as well as a long series of forum posts on player-based events, beginning with post 17095629.1 and continuing across 11 additional threads, as listed in post=17537676.1.

**Table 3.** Positive and negative player-based special events implemented in the knowledge-based BN model.

| Special event | Positive event/s | Negative event/s |
|---|---|---|
| Corner (team) | a. The ball receiver may score a goal, with players possessing the *Head* speciality having an increased chance of converting the opportunity. | n/a |
| Winger (player) | a. The ball receiver may score a goal, with players possessing the *Head* speciality having an increased chance of converting the opportunity. | n/a |
| Head (player & team) | a. Increases the probability of scoring a header goal from a corner event.<br>b. Increases the probability of scoring a header goal from a winger event.<br>c. Decreases the probability of conceding a header goal from a corner event | a. Non-forwards may be dribbled by an opposing *Technical* winger, midfielder, or forward, increasing the risk of conceding a goal. |
| Quick (player) | a. Wingers, midfielders, and forwards may break through on their own and score.<br>b. Wingers, midfielders, and forwards may break through on their own and pass the ball.<br>c. Central defenders and wing-backs may intercept an opponent's quick attack. | a. Players lose 5% of their performance in two out of the four possible weather conditions (modelled indirectly). |
| Powerful (player) | a. Can be instructed to man-mark a winger, midfielder, or forward, impacting team ratings (modelled indirectly).<br>b. Gains a 5% performance boost in one of the four possible weather conditions, impacting team ratings (modelled indirectly).<br>c. A Powerful Defensive Inner Midfielder (PDIM) may block an opponent's attack.<br>d. A Powerful Normal Forward (PNF) may generate an extra attack for each unsuccessful attempt. | a. Lose 5% of their performance in one of the four possible weather conditions, impacting team ratings (modelled indirectly).<br>b. Fielding more than one Powerful Defensive Inner Midfielder (PDIM) or Powerful Forward (PFW) applies an increasing penalty to their skills, impacting team ratings (modelled indirectly). |
| Technical (player) | a. Wingers, midfielders, and forwards may dribble past an opponent's non-forward player with the *Head* speciality and score.<br>b. Forwards receive a percentage boost in their passing skill, impacting team ratings (modelled indirectly).<br>c. Central defenders and wing-backs may generate a technical counterattack. | a. Wingers, midfielders, and forwards are 8% more affected by the opponent's man-marking effect, impacting team ratings (modelled indirectly). |
| Unpredictable (player) | a. The keeper, central defenders, and wing-backs may generate an unpredictable long pass that could lead to a goal. | a. Wingers and forwards may generate an unpredictable own goal.<br>b. Central defenders, wing-backs, or midfielders may make an unpredictable |
| | b. All players (except keepers) may generate an unpredictable special action that could result in a goal.<br>c. Wingers, midfielders, and forwards may score an unpredictable goal on their own.<br>d. Wingers, midfielders, and forwards are 8% less affected by the opponent's man-marking effect, impacting team ratings (modelled indirectly). | mistake that leads to an attack for the opponent.<br>c. A failed unpredictable special action may create a counterattack for the opponent.<br>d. A failed unpredictable solo scoring attempt may result in a counterattack for the opponent. |

Player-based special events are assumed to be mutually exclusive, with their occurrence rates presented in Table 4. That is, if a special event cannot be triggered in a match due to the absence of certain player specialities in specific positions, its probability is redistributed amongst the remaining special events. The redistribution is weighted based on the base frequency of the residual events that can still be triggered.

**Table 4.** Overall frequencies of player-based special events, restricted to match instances where all events are possible, and where both teams play the default *Normal* tactic. Quick events account for the risk of being stopped by an opposing *Quick* defender.

| Player-based event | Frequency | Event rate per match | $p$ to score event (avg) |
|---:|---|---|---|
| Winger event (any) | 25.72% | 21.63% | 49.51% |
| Tech over head | 15.19% | 12.77% | 29.37% |
| Quick rush | 15.29% | 12.86% | 36.70% |
| Quick pass | 14.49% | 12.19% | 43.87% |
| Unpredictable long pass | 08.17% | 06.87% | 40.90% |
| Unpredictable score on his own | 06.37% | 05.36% | 58.22% |
| Unpredictable special action | 06.66% | 05.60% | 42.41% |
| Unpredictable mistake | 03.45% | 02.90% | 18.16% |
| Unpredictable own goal | 04.66% | 03.92% | 17.25% |

When an event is triggered, it is allocated with a higher probability to the team that has the greater number of players capable of supporting that event. For example, if a *Quick Rush* event is generated and Team A has two offensive *Quick* players while Team B has one, the allocation probability is assumed to be 0.67 for Team A and 0.33 for Team B. This assumption aligns with the possession model described in in-game forum post 17537676.367.

If the special event is team-based, a linear difference in midfield ratings is used to determine event allocation, rather than the non-linear difference described in Equation 1. Equation 5 defines this linear difference:

$$LinearPOS_{HT} = \frac{(R_{HT}^{M} \times 4 - 3)}{(R_{HT}^{M} \times 4 - 3) + (R_{AT}^{M} \times 4 - 3)} \quad (5)$$

where $LinearPOS$ represents linear possession. Table 5 lists the team-based special events, along with their frequencies and goal rates.

**Table 5.** Overall frequencies of team-based special events, averaged across all matches.

| Team-based event | Frequency | Event rate per match | $p$ to score event (avg) |
|---|---|---|---|
| Experienced Forward | 10.76% | 04.00% | 37.04% |
| Inexperienced Defender | 10.54% | 03.92% | 10.50% |
| Tired Defender | 00.12% | 00.04% | 34.32% |
| Corner | 78.58% | 29.22% | 48.49% (to-anyone)<br>55.03% (to-head) |

We modelled the first two events listed in Table 5, *Experienced Forward* and *Inexperienced Defender*, independent of possession, contradicting the official information shared in in-game post 17537676.4. This is because the initial model we built, making them dependent on possession, showed no relationship or influence between the two. Instead, in the final version of the knowledge-based model, we chose to link these events to the number of forwards and defenders, identifying a clear linear relationship, which we modelled as illustrated in Figure 13. While it is likely that these events are influenced by overall team experience, as suggested in in-game forum post 17192231.14, this could not be verified in our study due to the absence of relevant data. Therefore, future work should investigate this further when the relevant data becomes available.

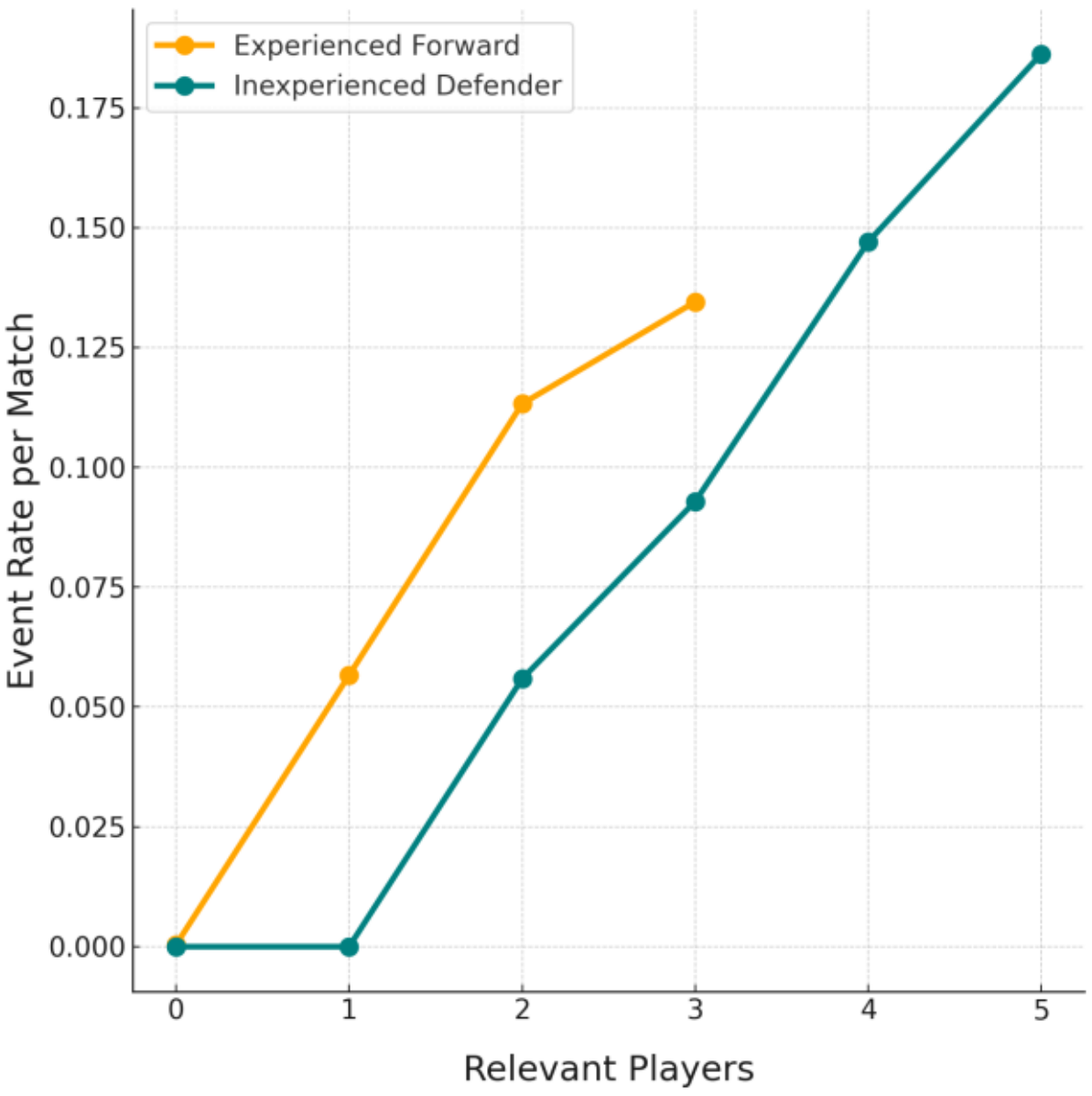

**Figure 13.** Probability of observing *Experienced Forward* and *Inexperienced Defender* events in a match for a given team, based on the number of relevant players; up to three forwards for the former and five defenders for the latter event captured.

Moreover, corners are divided into two sub-events: corner-to-anyone and corner-to-head, with the latter generally having a higher probability of resulting in a goal. Figure 14 (left) illustrates how the model determines the probability of scoring from a corner-to-anyone event, based on the difference between ISP attack and defence ratings. Figure 14 (right) presents the probability of scoring from a corner-to-head event according to Equation 6, which we implemented, compared to the empirical probability reflected in bar heights:

$$\text{score_CtH}(a,b) = \begin{cases} \dfrac{a}{a+b}, & \text{if } a > b \\ \dfrac{a}{a+b} - 0.05 \times \left(1 + 0.1 \times (b-1)\right), & \text{if } a = b \\ \dfrac{a}{a+b} \times \dfrac{a}{b}, & \text{if } a < b \end{cases} \tag{6}$$

where $\text{score_CtH}$ represents the probability of scoring a corner-to-head event, $a$ denotes the number of offensive headers in the attacking team, and $b$ denotes the number of defensive headers in the defending team. The comparison is based on the difference between offensive and defensive players with the *Head* speciality. The highest error (tallest bar) of 13.63% between estimated and empirical probabilities is observed when the matchup is 4 vs 5 headers, followed closely by 3 vs 5 headers at 12.78%. Overall, Equation 6 is found to capture this relationship reasonably well, except when the defending team has five headers, where the error ranges between 2.85% and 13.63%. Lastly, the probability of a corner event being a corner-to-head rather than a corner-to-anyone increases with the number of offensive headers. Specifically, the probability is 27% with one offensive header, 42% with two, 51% with three, 59% with four, and 65% with more than four offensive headers.

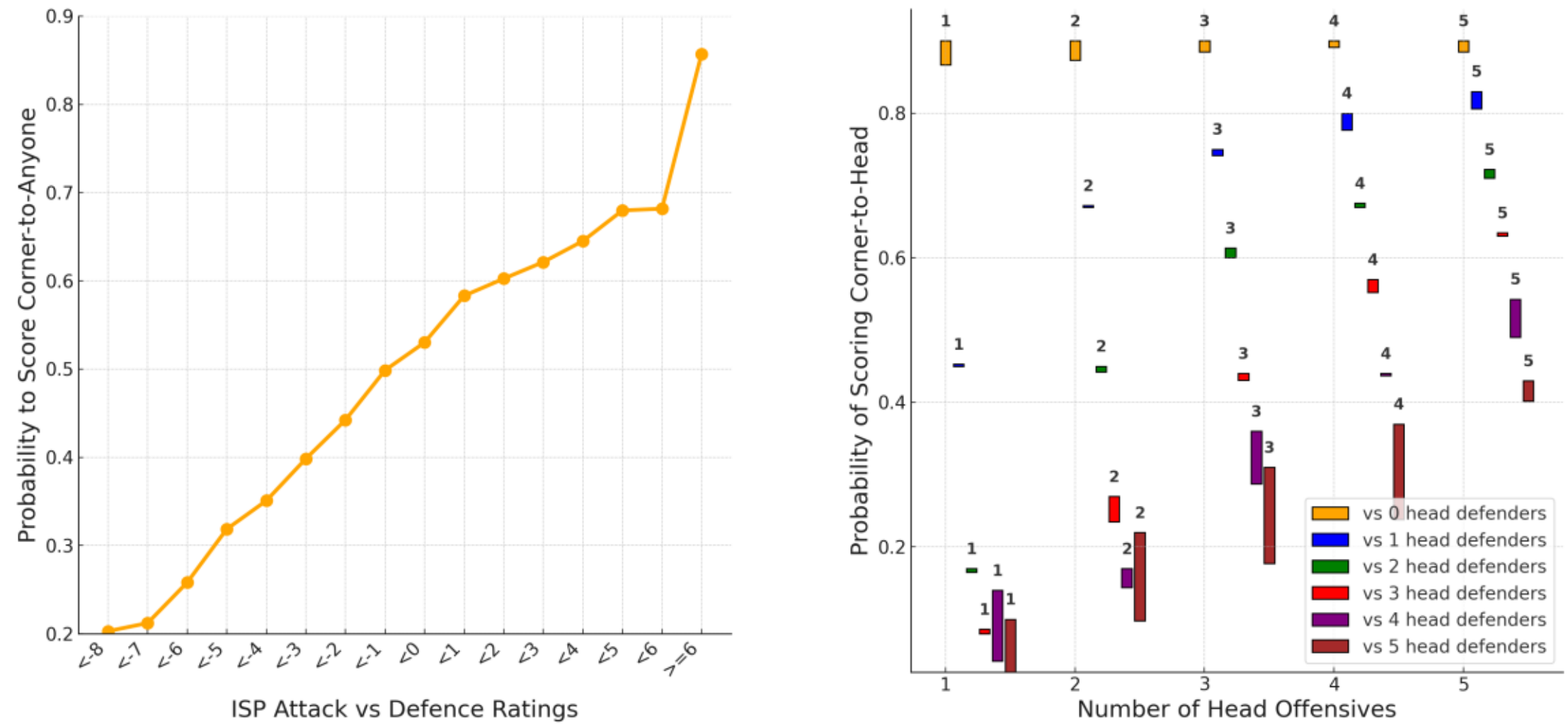

**Figure 14.** The left figure presents the probability of scoring from a corner-to-anyone event based on the difference between Indirect Set-Pieces (ISP) attack and defence ratings. The right figure presents the probability of scoring from a corner-to-head event based on the number of offensive headers (ranging from 1 to 5, as indicated above the bars) versus head defenders (ranging from 0 to 5, as indicated by colour), where the bar heights represent the difference between the estimated probability derived from Equation 6 and the empirical probability.

Winger events can only occur for teams that field at least one winger, while winger-to-head events require the presence of at least one offensive player with the *Head* speciality in addition to the winger. Since the probability of scoring from this event is said to depend on the winger's winger skill relative to the opposing defender's defending skill, we use the probability of scoring from each wing-side attack as a proxy for estimating the scoring probability of a winger-to-anyone event. Figure 15 illustrates this relationship.

Similar to corner events, having more offensive headers increases the probability of a winger event being a winger-to-head rather than a winger-to-anyone event. This probability increases with the number of offensive headers as follows: 15% with one offensive header, 29% with two, 35% with three, 38% with four, and 42% with more than four offensive headers. Since it was not possible to extract the probability of scoring a winger-to-head event directly from the data, we assume that it has, on average, a +11% higher success rate relative to a winger-to-anyone event for a given team, based on in-game forum posts 17342010.442 and 17342010.443.

A Powerful Defensive Inner Midfielder (PDIM) has a chance to press an opponent's *Normal* attack, while a Powerful Normal Forward (PNF) has a chance to generate an extra attack for each missed *Normal* attack. Figure 16[12] illustrates how the probability of pressing an opponent's *Normal* attack varies with the number of PDIMs fielded, and how the probability of generating an extra attack for each missed normal attack is influenced by the number of PNFs fielded and the number of opposing central defenders (CDs).

Lastly, Technical central defenders or wing-back have a chance to generate a counterattack for each opponent's missed normal attack. Specifically, the conversion rate is 0.84% for two defenders, 1% for three, and 3.11% for more than three.

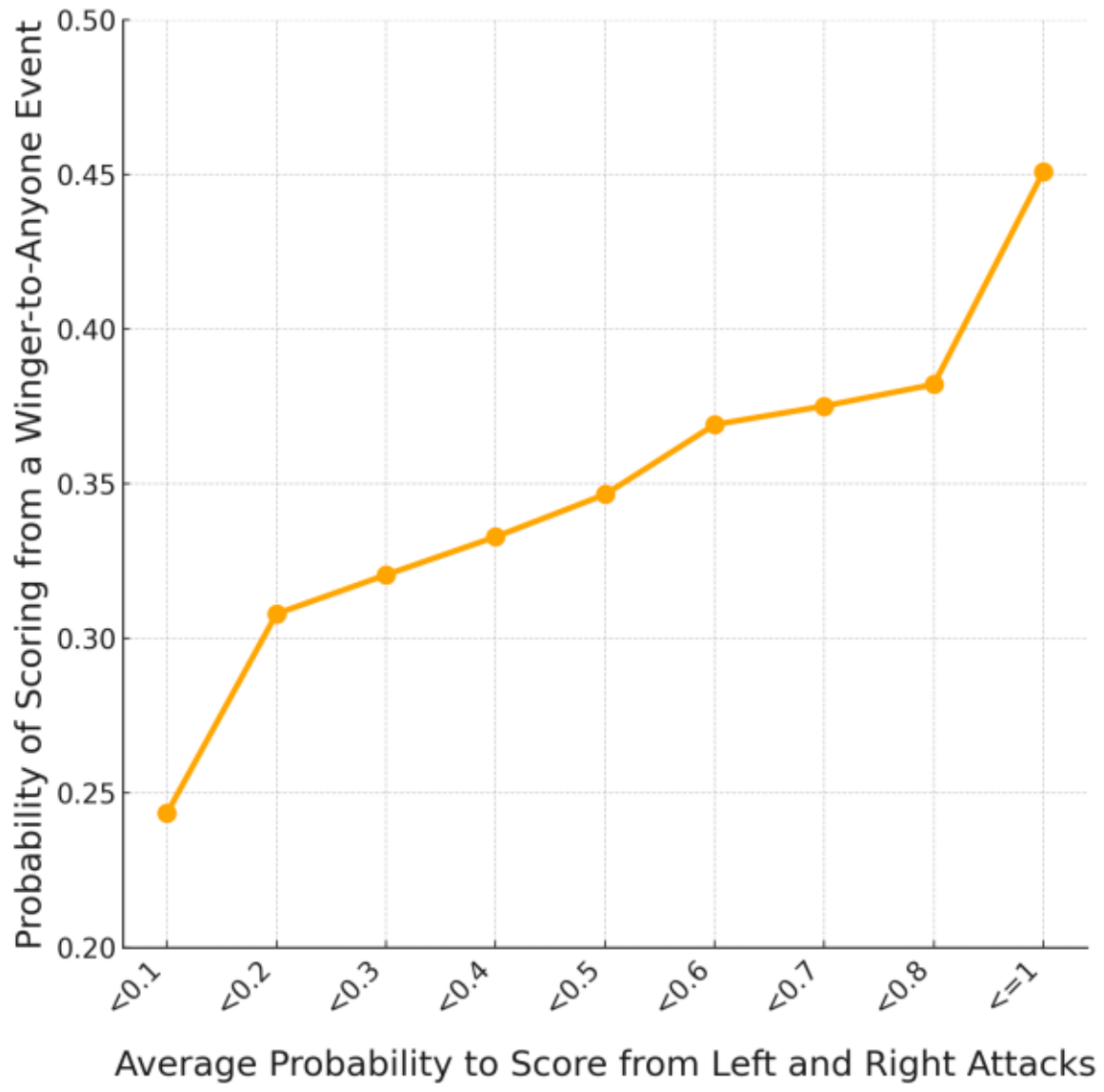

**Figure 15.** Probability of scoring from a corner-to-anyone event based on the difference between Indirect Set-Pieces (ISP) attack and defence ratings (left figure). The non-cumulative intervals of the $x$-axis in the right chart do not incorporate data from previous intervals.

[12] These conversion rates do not align with those reported in the official game manual, which states that a single PDIM has a 10% chance to block an opponent's attack, two have a 16% chance, and three have a 20% chance. The same percentages are mentioned for PNFs for creating a new attack from a previously missed attack. This discrepancy may be attributed to hidden confounding factors, such as player skills not captured by data. However, the conversion rates reported in this study are consistent with those documented in the *Unwritten Manual*, based on in-game forum posts 17342010.652 and 17342010.13.

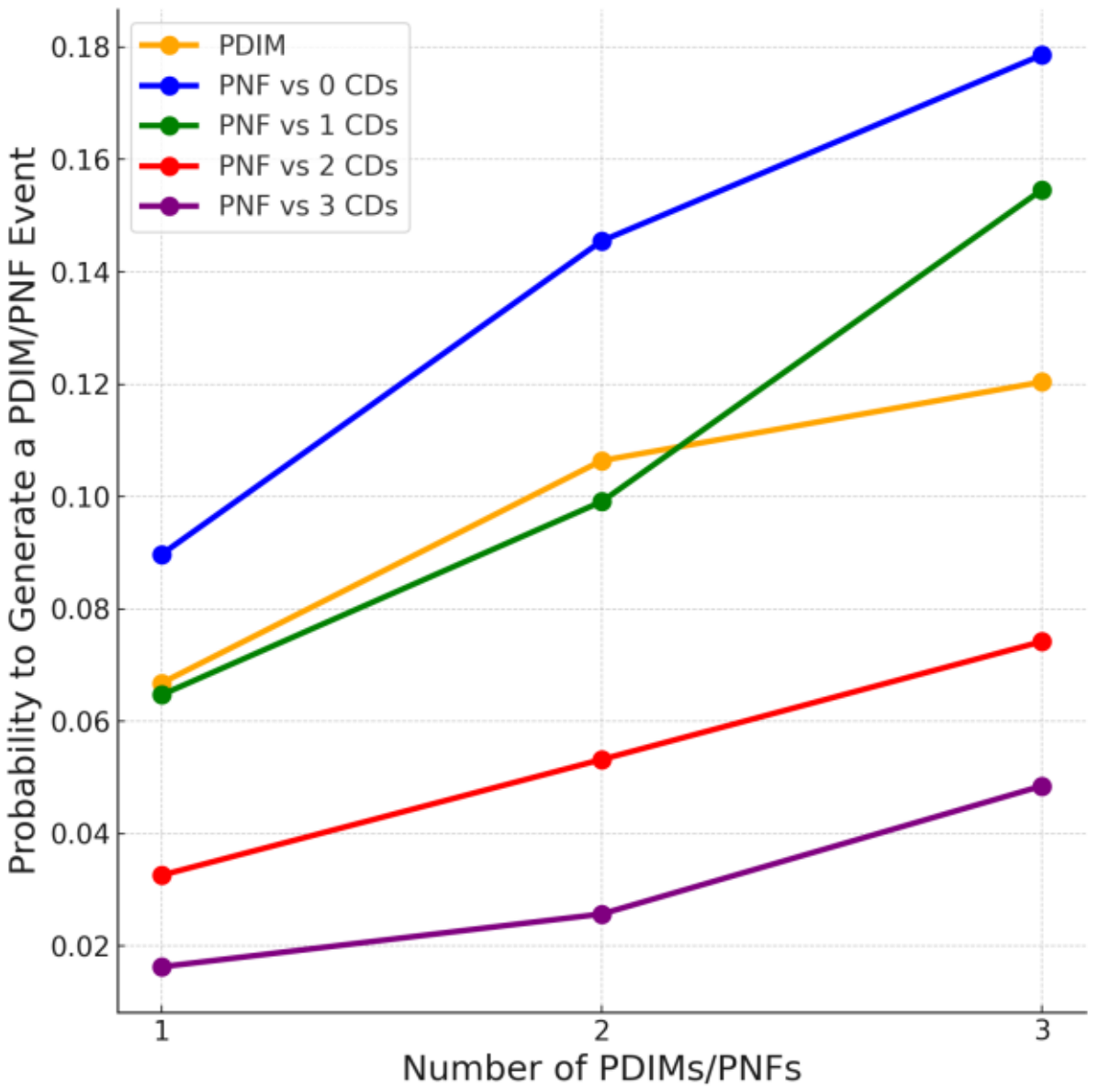

**Figure 16.** Probability of pressing an opponent's normal attack based on the number of Powerful Defensive Inner Midfielders (PDIMs), and probability of generating an extra attack for each missed *Normal* attack based on the number of Powerful Normal Forwards (PNFs) against the number of opposing Central Defenders (CDs).

## 5. Results Part A: Graphical structures

This section focuses on evaluating the graphical structures (i.e., relationships) learnt by the structure learning algorithms. It presents both quantitative and qualitative evaluations, assessing the learnt structures using model selection scores and by comparing learnt relationships with those elicited from domain experts, respectively. Section 5.1 describes the graphical metrics and model selection scores used to compare BN structures learnt from data with expert-elicited structures, while Section 5.2 presents the comparative evaluation results.

### *5.1. Preliminaries*

For structural evaluation, we focus on graphical metrics, objective functions, and other statistical properties related to these graphical representation models. We begin by outlining the three graphical metrics employed, which are commonly used in the structure learning community to assess the similarity between two graphical structures. Firstly, we use the F1 metric, which produces a score ranging from 0 to 1, representing the harmonic mean between Precision (P) and Recall (R), as defined by Equation 7

$$F1 = 2\frac{P \times R}{P + R}, \quad \text{where} \quad P = \frac{TP}{TP + FP} \quad \text{and} \quad R = \frac{TP}{TP + FN} \tag{7}$$

where $TP$ represents *true positive* edges; i.e., edges that are present and identically orientated in both the learnt and ground truth graphs, $FP$ represents *false positive edges*; i.e., edges present in the learnt but absent in the ground truth graph, and $FN$ represents *false negative* edges; i.e., edges present in the ground truth but missing from the learnt graph.

Secondly, we employ the Balanced Scoring Function (BSF) by Constantinou [40], which generates a score ranging from -1 to 1 by incorporating all four parameters of the confusion matrix. This function balances the score such that both an empty graph and a fully connected graph yield a score of 0. This is achieved by adjusting the reward function based on the relative difficulty of identifying a correct present edge versus a correct absent edge. This adjustment is proportional to the number of edges present in the ground truth compared to the number of edges absent. The BSF is defined by Equation 8:

$$BSF = \left( \frac{\frac{TP}{|E|} + \frac{TN}{|M|} - \frac{FP}{|M|} - \frac{FN}{|E|}}{2} \right), \quad \text{where} \quad |M| = \frac{N \times (N-1)}{2} - |E| \tag{8}$$

where $TN$ represents *true negative* edges; i.e., edges absent in both the learnt and ground truth graphs, $N$ is the total number of nodes, and $E$ and $M$ denote the number of edges present and absent, respectively, in the ground truth graph.

Thirdly, we use the Structural Hamming Distance (SHD), which quantifies the number of edge modifications, such as edge additions, deletions, and reversals, required to make the learnt graph identical to the ground truth [23]. We consider the special, but commonly adopted, variant of SHD in which edge reversals incur an error of 0.5 instead of 1, to acknowledge that an arc reversal signifies the correct identification of a dependency but with an incorrect direction. The SHD score effectively represents the summation of FP and FN edges described above, and is known to heavily favour algorithms that produce sparse structures and limit propagation of evidence in practice [40]. However, its simplicity and interpretability continue to make it a widely used measure in this area of research.

For objective scores, we report the Bayesian Information Criterion (BIC) score, as it serves as the objective function used for optimisation by the score-based and hybrid learning algorithms employed in this study, as previously described in subsection 3.1. In the context of structure learning, the BIC score is utilised as a model selection function to evaluate both model complexity and data fit, aiming to balance the two to prevent overfitting. The structure learning algorithms in this study aim to maximise the BIC function, as defined by Equation 9:

$$BIC = LL(G|D) - \frac{\log_2 n}{2} k, \quad \text{where} \quad \mathrm{k} = \sum_{i}^{N} \left( (s_i - 1) \prod_{j}^{|\pi_{N_i}|} q_j \right) \tag{9}$$

where $G$ represents the input graphical structure, $n$ is the sample size of dataset $D$, $k$ is the number of free parameters, $s_i$ is the number of states of node $N_i$, $\pi_{N_i}$ is the parent set of $N_i$, $|\pi_{N_i}|$ denotes the size of set $\pi_{N_i}$, and $q_j$ is the number of states of node $N_j$ in the parent set $\pi_{N_i}$.

Lastly, we also evaluate the statistical properties of the graphical structures, including the Log-Likelihood (LL) from Equation 9 and the number of free parameters $k$, also from Equation 9, which represents the number of independent parameters or additional parameters introduced through edge additions. Additionally, we consider the average and maximum in-degree, which represent the number of edges pointing into a node (i.e., parents), as these typically drive model complexity.

### 5.2. *Comparison between the structures learnt from data and the knowledge-based structures elicited from domain experts*

Table 6 presents the results of structure learning based on results from each algorithm, with the knowledge-based structure containing the same number of variables serving as a reference. We present the knowledge graph in Figure 17, and provide the graphical structures learnt by each of the algorithms in Appendix D. In terms of fit and complexity, all algorithms - except MMHC and GS, which are known to produce sparse structures - generated graphical structures that improved both the BIC and $LL$ scores relative to the knowledge-based structure. While it is expected that data-driven algorithms would optimise these functions better than manually constructed structures, part of the discrepancy in this study arises from the fact that the knowledge-based structure does not explicitly link related input variables that are always observed during inference. For instance, we would expect a team with high side attack ratings to also have high middle attack ratings, and structure learning algorithms would likely detect such relationships between input variables. While modelling these dependencies is not incorrect, we deliberately omitted them from the knowledge-based model to better manage model complexity without affecting key model outputs. This decision was based on the assumption that these input variables would always be observed (i.e., user input), eliminating the need to estimate them.

As a result, structure learning algorithms identified nearly twice the number of edges compared to those incorporated into the knowledge-based graph, while achieving similar $LL$ and BIC scores. However, MMHC and GS produced far fewer edges. Interestingly, MMHC and GS also resulted in weaker fitted graphical structures relative to the knowledge-based graph, highlighting their strong preference for sparsity towards optimising for SHD (we further discuss this below in relation to graph-based metrics). Lastly, the output of PC-Stable was a PDAG that could not be converted into a CPDAG or a DAG (refer to subsection 2.2), preventing us from transforming its learnt structure into a parameterised BN model.

In terms of graph structural metrics, MMHC and GS achieve the best SHD score, reflecting their preference for maximising this metric [23, 32]. However, the graphs learnt by these two algorithms are not particularly useful for inferential capability, as their sparse structures limit the propagation of evidence within the models. This is also evident in the number of graphical fragments in Table 6, where MMHC and GS generate an unreasonably high number of disjoint subgraphs, preventing inference propagation between most nodes. Therefore, our interest in this study focuses on algorithms that maximise the F1 and BSF metrics, as these scores are more indicative of graphical structures that are practically useful, as we later demonstrate in subsection 5.2.

Figure 18 presents a model averaging graph, generated using the Bayesys tool (see Table 1 caption), based on the number of times an edge is learnt by each algorithm. The averaging approach follows that used by Constantinou et al. [29] to combine COVID-19 causal structures learnt by different structure learning algorithms, and aligns with the averaging principles discussed by Cugnata et al. [39]. To be able to visualise the model averaging graphs, we present only the node families (parents and children highlighted in yellow) of selected special events that remain a topic of debate within the Hattrick community, associated with *Quick*, *Technical*, *Head*, and *Unpredictable* player specialties. Each edge has a count, with its width increasing based on how often it appears in graphs learnt by the algorithms for home or away teams. Acyclicity is enforced by first adding edges that maximise edge counts, and then reversing or dropping edges that violate acyclicity. The maximum possible count for an edge in this exercise is 16 (eight algorithms applied to home and away teams).

The model-averaging graph successfully captures some of the key relationships governing these special events. However, some high-count relationships challenge existing understanding and merit further investigation. For example, Figure 18 suggests that playing a counterattack tactic (*Event_CA* and *CA_tactic* nodes) may affect the number of *Technical* counterattacks generated by defenders (*Event_SEZ_TechCA* node). One possible explanation is that the tactic reduces an opponent's missed chances that could trigger a *Technical* counterattack, under the assumption of a hierarchical check where a tactical counterattack is prioritised, and only if it fails is a technical counterattack considered.

Another finding worth highlighting is that the corner-to-head event (*Event_TSE_CornerHead* node) is influenced by the Creative tactic (*Creative_tactic* node), as expected, but also by the variable which includes all tactics (*TacticType* node). This suggests that the Creative tactic alone may not fully explain this event.

Interestingly, the algorithms also identified spurious relationships between specialities, such as (a) an unpredictable special action (*Event_PSE_UnpredSpecial* node) influencing unpredictable score-on-his-own event (*Event_PSE_UnperScoreOwn* node), and (b) a quick rush event (*Event_PSE_QuickRush* node) affecting *Technical* and *Head* events (*Event_TSE_CornerHead* and *Event_TSE_WingHead* nodes). These edges are spurious because, as confirmed by Hattrick developers [38], the game-engine generates and allocates special events based on the number of players with the corresponding specialty on the field. If a certain speciality is absent, other special events become more frequent. Thus, the relative number of players with a given specialty, rather than direct dependencies between events as learnt by some structure learning algorithms, should be their causes.

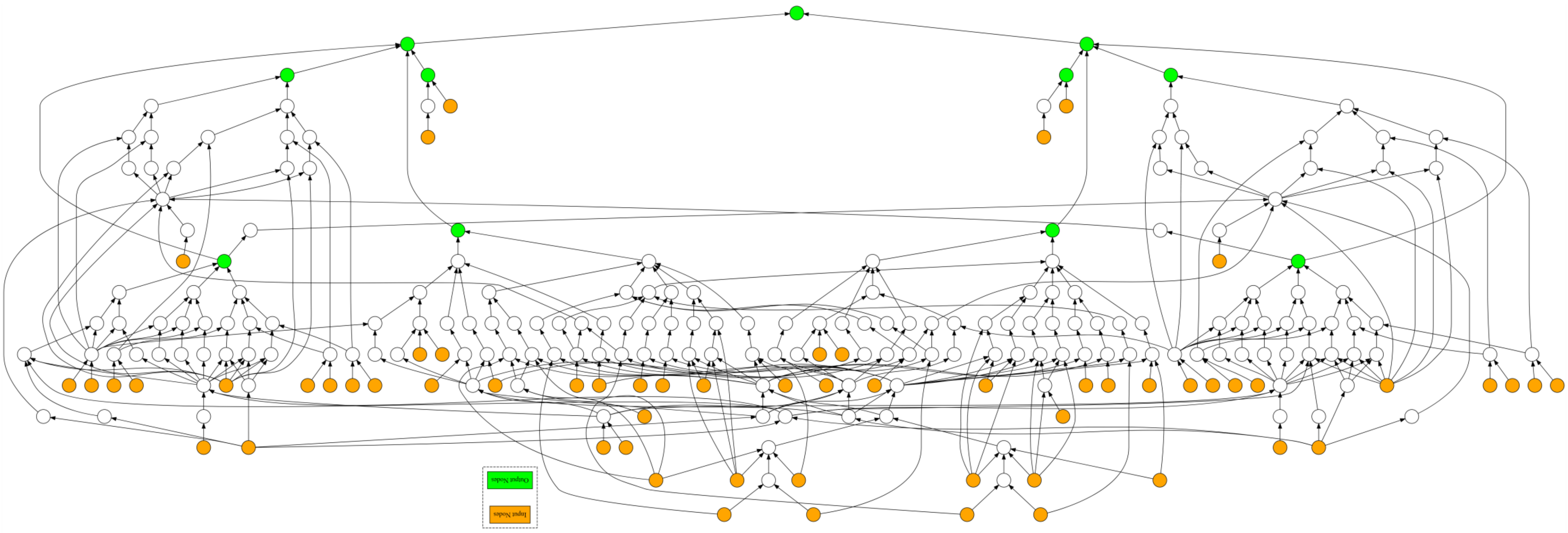

**Figure 17.** The knowledge-based graph with input (observed) nodes in orange and output (inferred) nodes of interest in green.

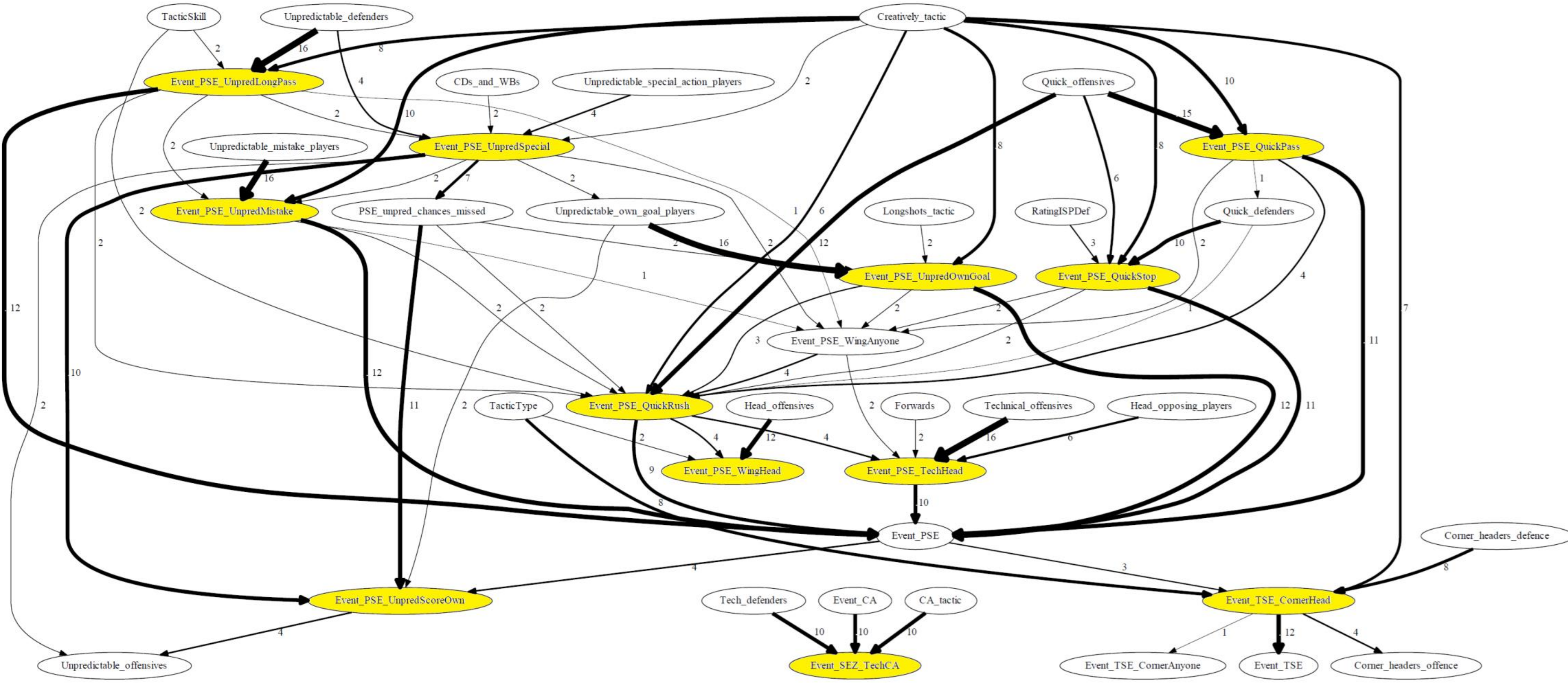

**Figure 18.** A model averaging graph of structure learning algorithms, highlighting node families linked to *Quick*, *Technical*, *Head*, and *Unpredictable* special events. Edge width reflects edge counts across graphs.

**Table 6.** The graphical properties and results of the structural models learnt from data and constructed by expert knowledge. The SHD, F1, P (Precision), R (Recall), and BSF scores assume CPDAG comparisons between graphs.

| Learnt or knowledge-based BN | LL (million) | BIC (million) | Edges | Free param | In-degree (avg) | In-degree (max) | Graphical fragments | SHD | F1 | P | R | BSF |
|---|---|---|---|---|---|---|---|---|---|---|---|---|
| Knowledge | -197.76 | -202.58 | 386 | 483,349 | 1.544 | 5 | 1 | n/a | n/a | n/a | n/a | n/a |
| HC | -150.58 | -156.21 | 681 | 564,990 | 2.724 | 10 | 1 | 552.0 | 0.446 | 0.349 | 0.617 | 0.603 |
| HC-Stable | -150.04 | -156.74 | 693 | 672,084 | 2.772 | 10 | 1 | 561.0 | 0.449 | 0.349 | 0.627 | 0.613 |
| Tabu | -150.58 | -156.21 | 680 | 564,970 | 2.72 | 10 | 1 | 550.5 | 0.447 | 0.351 | 0.618 | 0.605 |
| Tabu-Stable | -150.02 | -156.74 | 695 | 674,185 | 2.78 | 10 | 1 | 563.0 | 0.448 | 0.348 | 0.627 | 0.613 |
| FGES | -150.28 | -156.97 | 720 | 671,313 | 2.88 | 9 | 1 | 604.0 | 0.418 | 0.321 | 0.598 | 0.584 |
| MMHC | -269.14 | -269.17 | 108 | 3,518 | 0.432 | 2 | 143 | 363.0 | 0.190 | 0.435 | 0.122 | 0.121 |
| GS | -275.66 | -276.31 | 71 | 64,597 | 0.284 | 3 | 180 | 381.0 | 0.127 | 0.408 | 0.075 | 0.074 |
| PC-Stable | n/a | n/a | 283 | n/a | n/a | n/a | 31 | 312.0 | 0.448 | 0.530 | 0.389 | 0.386 |

Regarding the F1 score, all algorithms, except MMHC and GS which generate high *precision* and low *recall* due to the limited number of edges they learn, produce similar and reasonably good F1 scores, suggesting a fair alignment with the knowledge-based graph despite the large number of additional edges discovered by the algorithms. The BSF scores suggest even more promising results, particularly for the HC and Tabu algorithms and their variants, with scores exceeding 0.6, indicating even greater alignment. In the following subsections, we further investigate the inferential capabilities of these algorithms (excluding PC-Stable), alongside other in-game models and examine how these graphical scores and metric values relate to the inferential power of the resulting models.

## 6. Results Part B: Inferential Performance

Section 6 evaluates the learnt models across multiple aspects of predictive performance, including goal distributions by team tactic and match outcome prediction (home win, draw, and away win). Subsection 6.1 describes the evaluation metrics and the existing Hattrick models used as benchmarks. Subsection 6.2 presents the evaluation results, which are then discussed and interpreted in subsection 6.3. Subsection 6.4 illustrates decision support by simulating the impact of various decision options using the learnt models. Lastly, we provide additional findings that suggest directions for future research in Appendix E.

### *6.1. Preliminaries*

In this subsection, we evaluate the inferential capability of the BN models with reference to a set of output nodes of interest. We focus on predicting the number of goals scored in the game across four different methods: (a) PNF (Powerful Normal Forwards), (b) counterattacks, (c) special events, (d) *Normal* attacks, and (e) the total number of goals scored. Beyond goal predictions, we also assess (f) the accuracy of home-win, draw, and away-win (HDA) distributions. Outcomes (a) to (e) are evaluated based on the mean absolute difference between predicted and observed goals scored in a match (e.g., if we observe match score 2-1 with a goal difference of +1, and we predict a goal difference of -1, that would result in a goal difference error of 2), whereas outcome (f) is measured using the Rank Probability Score (RPS), which is based on the Brier score [41], and extends the assessment of binary forecasts to multinomial forecasts.

RPS has been extensively used over the past decade to evaluate probabilistic football match forecasts in the literature and international football predictive competitions [42, 43, 44]. Given its relevance, we adopt it for assessing the accuracy of predictions in this football simulation game as well. Specifically, RPS quantifies the difference between two cumulative ordinal distributions, in this case the predicted and observed HDA distributions, as defined by Equation 10:

$$\text{RPS} = \frac{1}{r-1}\sum_{i=1}^{r-1}\left(\sum_{j=1}^{i}(p_j - e_j)\right)^2 \tag{9}$$

where $\mathrm{r}$ is the number of categories captured by each distribution (three in our case, for home win, draw, and away win), $\mathrm{p_j}$ represents the predicted probability at position j such that $\mathrm{p_j} \in [0, 1]$ for $\mathrm{j} = 1, 2, 3$, with $\mathrm{p_1} + \mathrm{p_2} + \mathrm{p_3} = 1$. Similarly, $\mathrm{e_j}$ represents the observed outcome at position j such that $\mathrm{e_j} \in [0, 1]$ for $\mathrm{j} = 1, 2, 3$, with $\mathrm{j_1} + \mathrm{j_2} + \mathrm{j_3} = 1$.

The evaluation considers a total of 14 models, four of which are models commonly used by Hattrick users and serve as benchmarks. These are:

a) **BNs learnt entirely from data:** A set of seven BN models learnt exclusively from data using the structure learning algorithms outlined in Table 6 (excludes PC-Stable). These models are referred to by the name of the structure learning algorithm used: **HC**, **HC-Stable**, **Tabu**, **Tabu-Stable**, **FGES**, **MMHC**, and **GS**. All seven BN models are discrete and trained on discretised data.

b) **BNs learnt from data and expert knowledge:** A set of three BN models where the structure was constructed based on expert knowledge, while parameters were learnt either from data or expert knowledge. These models include (a) **Knowledge**: A discrete BN model with parameters learnt from data, (b) **KB-regression**: A hybrid BN model with parameters specified from expert knowledge, using regression functions, discrete variables, and other pre-defined equation-based relationships derived from in-game official and unofficial manuals, and (c) **KB-probabilistic**: A hybrid BN model with parameters specified from both data and officially verified knowledge, but represented using Beta-Binomial statistical distributions instead of the regression-based functions used in (b), in addition to the discrete variables and the pre-defined relationships.

c) **Established Hattrick models:** A set of four models developed and actively used by the Hattrick community, most of them for decades, serving as established benchmarks. These models have not been published in the literature, and there is limited information available about them. Below is key information for each of these models:

   i. **HT-ML:** A machine learning model that is based on the LightGBM algorithm [45], that uses gradient boosting and decision trees, to estimate the probability of winning for each team in a Hattrick match, trained on over 2 million matches [46]. There are two key differences between HT-ML and the BN models learnt in this study. Firstly, HT-ML does not account for special event variables (see Subsection 4.4) in its training, as they found that "*the predictions are more accurate without this feature*" [46]. Secondly, HT-ML recalculates team ratings in the event of a *pullback*, an in-game mechanism that increases defence ratings at the expense of attack ratings when a team is winning by more than one goal, with the effect intensifying as the goal difference increases. In contrast, this study assumes static team ratings across all trained models, a limitation discussed in Section 6.

   ii. **HTMS:** A model that estimates match outcomes by implementing and simulating the game rules, team ratings, and formulas widely assumed by the Hattrick user community [47]. HTMS is also available as a browser add-on, enabling users to seamlessly integrate model outputs into the Hattrick online browser game for real-time analysis. Like HT-ML, HTMS does not consider special event variables in its calculations.

iii. **DHTH:** Another Hattrick match simulator that, like HTMS, implements game rules and formulas assumed by the community [48]. However, unlike HT-ML and HTMS, DHTH also simulates special events. Similar to HTMS, it is seamlessly integrated with the Hattrick browser game for analysis. Like HT-ML, DHTH accounts for the dynamic effect of the *pullback* event, but also considers two additional dynamic factors. The first factor is *underestimation*; a penalty applied to the midfield rating of significantly stronger teams based on their difference in accumulated league points (larger discrepancies increase the occurrence of this event). The second factor is *stamina*; a drop in player performance towards the end of a match, proportional to each player's stamina level. As with the *pullback* event, the dynamic effects of *underestimation* and *stamina* are not considered in the models trained in this study (see limitations in Section 6)

iv. **Nickarana:** A rule-based model built on in-game equations only, to estimate match outcomes. The model's elements are derived from empirical studies using large datasets that capture over 20 million matches, aimed at confirming or correcting community assumptions. The model is available online [49] for completed matches and through an Excel file for user-entered parameters

### *6.2. Results – Predicting goals scored*

Of the four Hattrick models, we were only able to apply the Nickarana model to all 1 million samples, alongside the 10 BN models discussed in the previous subsection. We were unable to automate the application of HT-ML, HTMS, and DHTH models to the full dataset, requiring us to manually apply them to a small set of matches; a process that was highly time-consuming. However, as we later show, this is not a significant concern because the Nickanara model is the best-performing Hattrick model. We, therefore, split the evaluation into two parts: (a) we test the 10 BN models, plus the Nickarana model, on the entire 1 million sample dataset, and (b) we test all four Hattrick models on a smaller subset of 422 samples, comparing them against the three knowledge-based BN models which, as we later show, outperform all fully data-driven BNs. The 422 matches were randomly selected from the dataset; initially, we sampled approximately 1,000 matches, but ultimately considered 422 due to DHTH failing to generate outputs for older[13] matches. Each match is associated with a unique ID, which we used to extract predictions from all four Hattrick models.

Generating 1 million predictions per model proved to be highly time-consuming. Each BN model, except for MMHC and GS which were sparse, required up to approximately three days of continuous runtime to generate inferences using GeNIe's inference engine, equivalent to approximately four BN models inferred per second. This process involved entering evidence into the BN model, running inference, and recording the set of outputs of interest, resulting in a total runtime exceeding one month on a single computer.

This computational burden introduced severe limitations in our ability to perform cross-validation, leading us to compare models based on non-cross-validated performance. However, to ensure that the knowledge graph does not lead to overfitting, we conducted a hold-out validation on the Knowledge BN model. Specifically, we trained the model on 90% of randomly selected data samples and evaluated it on the remaining unseen 10%. The results showed only a trivial difference in predicting the HDA distribution, with the training score of 0.18945 increasing (worsening) marginally to 0.18946, suggesting no risk of model overfitting. This result is further supported by the findings in Table 6, which show that the Knowledge graph contains approximately half the number of edges, with around half the average and maximum in-degrees, and fewer free parameters, compared to all other BN models learnt through structure learning; except for MMHC and GS, which, as we later demonstrate, the resulting BN models underfit the data.

[13] Some information is lost in the API for older matches. For example, when a player is removed from the game, some of their attributes (e.g., specialities such as Head or Quick) are no longer available through the API and cannot be retrieved from the data. This causes the DHTH simulation to fail. It is difficult to predict which older matches will be affected, since managers can remove (i.e., fire) players at any time. The older the match, the greater the risk that at least one fielded player has been removed, leading to DHTH prediction failure.

Figure 19 presents six bar charts capturing the performance of the 11 models specified (the 10 BN models and the Nickarana model). Each chart corresponds to one of the outcomes (a) to (f) discussed in subsection 5.2.1. Specifically, the top four charts report performance across the four different types of goals scored, the bottom left chart focuses on total goals scored, and the bottom right chart evaluates the models' ability to predict overall match outcomes in terms of the HDA distribution.

There are two key takeaways from these results. Firstly, we observe an unusual outcome where GS outperforms all other models in predicting PNF and SE goals, and MMHC in SE goals, despite generating identical outputs for nearly all 1 million instances. The high variance in their prediction error arises due to the sparsity of their networks, which do not allow inference to propagate, resulting in the same prediction for each match. Specifically, in the GS model, the PNF goal outcomes (for HT and AT) had 0 parents and 0 to 1 children, while in MMHC they had 1 parent and 0 children. Similarly, for SE goal outcomes, the GS model had 0 parents and 0 children, whereas MMHC had 0–1 parents and 0 children. This structural sparsity prevents meaningful dependencies from being established, leading to the observed behaviour.

The key takeaway is that GS and MMHC rank highly in these cases not because they effectively learnt the causal mechanisms behind these events, but rather because their sparse structures left those variables largely unconnected, simply generating the average expectation for all matches. A plausible explanation for why this average expectation results in the lowest error is that PNF and SE goals are rare. This observation also aligns with the claim made by the author of the HTMS Hattrick model, who stated that "*the predictions are more accurate without this feature* [referring to special events]" [47].

On the other hand, for *Normal* and counterattack goals, which are more frequent making up around 80% and 6% (but 44% for teams playing counterattack strategy) of total goals respectively, GS and MMHC perform significantly worse than the other models. This trend is also evident in the *All goals* scored category and in the RPS for overall match outcomes (HDA). Notwithstanding the observations discussed above, the top-ranked models are dominated by either one of the three BN models incorporating expert knowledge or the Nickarana model.

Figure 20 replicates the analysis for All goals and HDA distributions on the 422-match sample discussed earlier, comparing the three BN models built using expert knowledge with the four Hattrick models. For this sample, Nickarana and KB-probabilistic perform best in terms of both accuracy and variability for HDA outcomes (Figure 20, right). While HT-ML exhibits the lowest error in goal difference (Figure 20, left), this reduced average error comes at the cost of higher variance, suggesting a less desirable outcome.

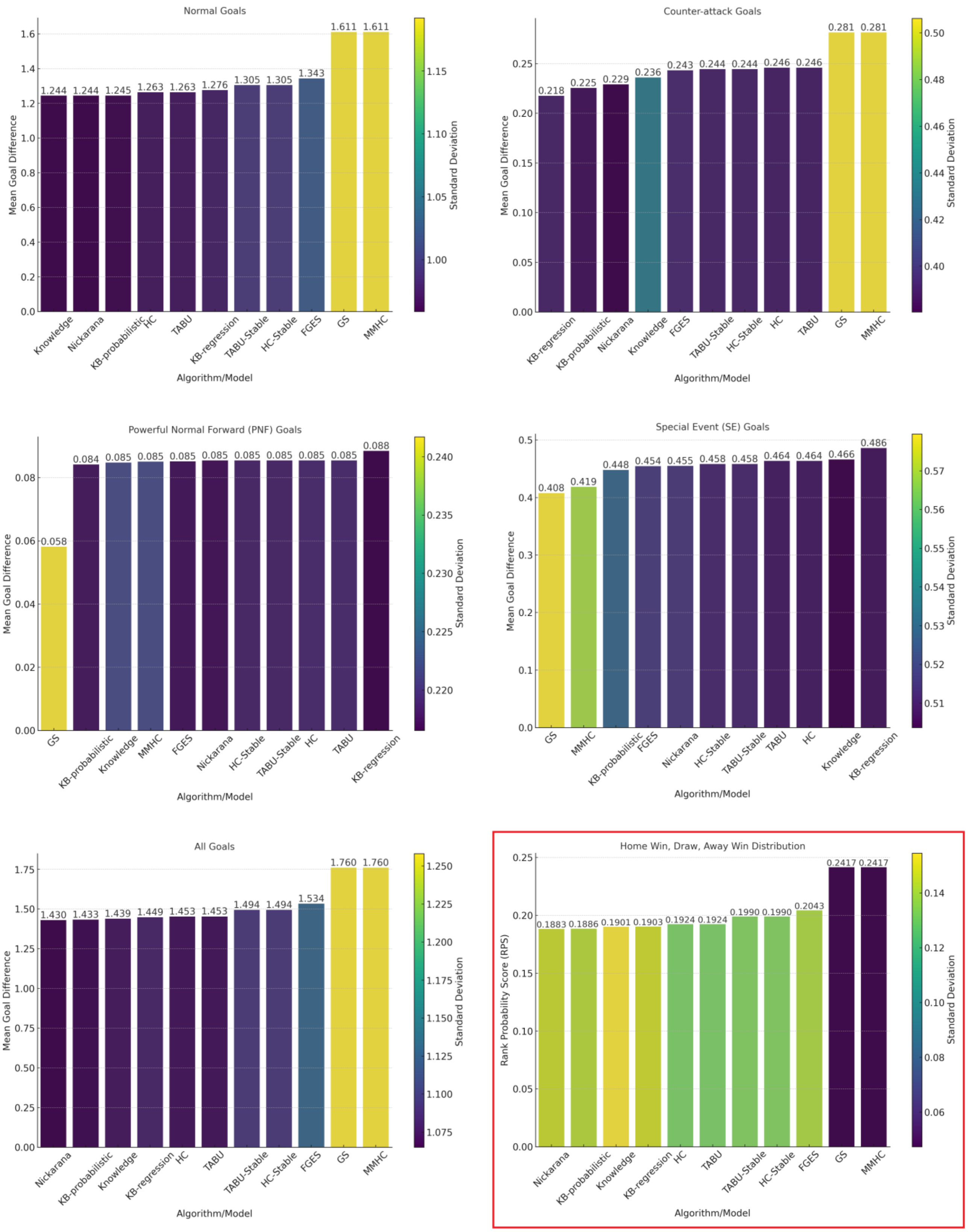

**Figure 19.** Inferential performance of the BN models constructed from data or incorporating expert knowledge, benchmarked against the best performing Hattrick model Nickarana. The top four charts illustrate how well the models predict the four different types of goals scored in terms of absolute goal difference error (i.e., predicted difference vs observed difference), the bottom left chart represents the difference in total goals scored, and the bottom right chart - highlighted with a red border - evaluates the models' ability to infer the overall match outcomes in terms of home win, draw, or away win (HDA distribution). These performance comparisons are based on all 1 million matches. Darker coloured bars represent lower standard deviation.

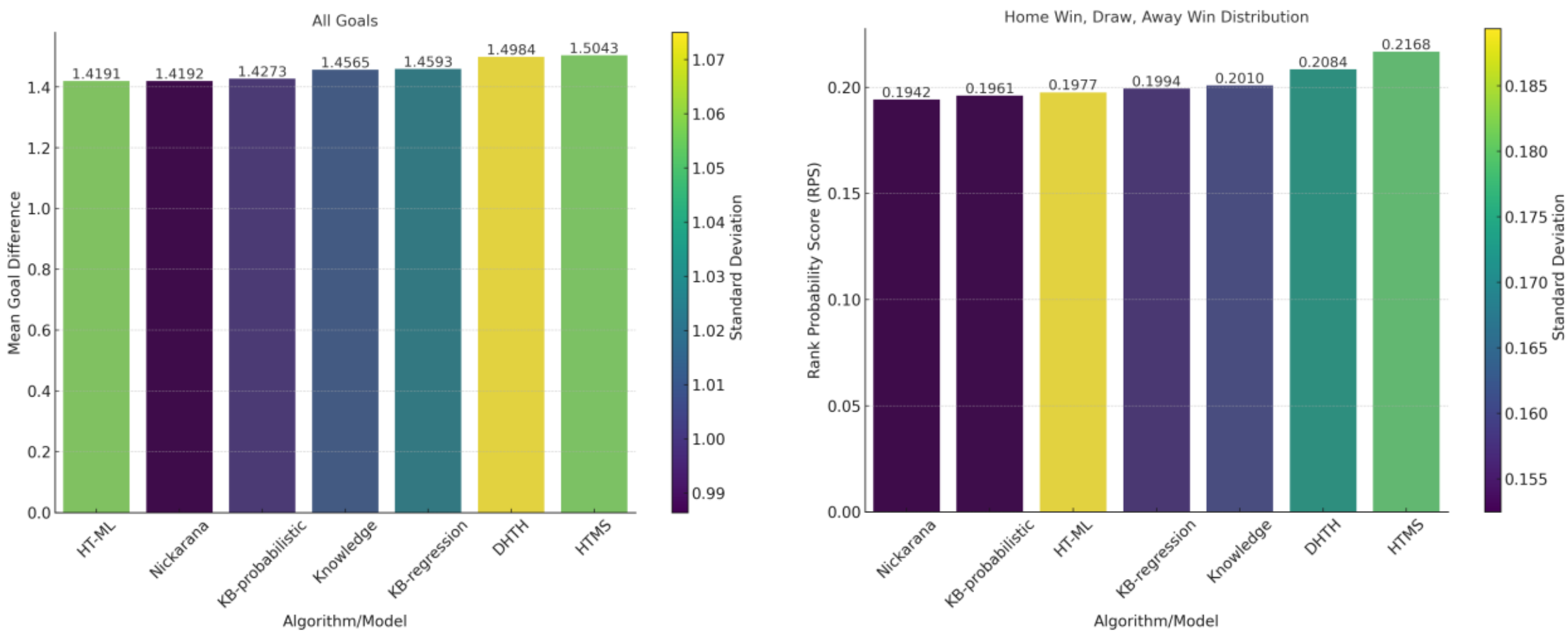

**Figure 20.** Inferential performance of the three best-performing BN models constructed from knowledge and data, compared against the four Hattrick models. The analysis evaluates both total goals scored (All) and home win, draw, and away win (HDA) distributions. These performance comparisons are based on a subset of 422 matches due to three, out of the four, Hattrick models requiring manual application. Darker coloured bars represent lower standard deviation.

### *6.3. Explanation of the results*

One might wonder about the significance of these results in the context of the case study, the accuracy of the models, and whether the models tested exhibit similar behaviour. First, it is important to clarify that the average number of goals scored per game in Hattrick is 3.76, which is considerably higher than the goal averages observed in real-life football matches. For instance, Li and Zhao [50] report that, over a ten-year period, the real-world average number of goals scored per team in the top five[14] European football leagues ranges from 1.259 to 1.457, and in [51] the match average was 2.69 goals across major leagues worldwide. The average scoring rate in Hattrick naturally results in a much larger mean absolute goal difference (Figures 18 and 19) compared to real-life football matches. Consequently, this subsection will focus on the HAD (Home-Draw-Away) distribution to interpret the results effectively, with reference to the RPS.

To begin, we use a set of hypothetical predictions in Table 7 to illustrate how the RPS behaves. The illustration demonstrates the conditions under which the RPS value is 0 or 1; i.e., 0 when the predictive distribution perfectly matches the observed distribution and 1 when they are as far apart as possible. Additionally, it shows the RPS values generated for various predictions, highlighting what impact the difference between distribution outcomes has on RPS. Specifically, if the observed outcome is $H$, prediction values for outcome $D$ will produce a lower error than prediction values for outcome $A$, and this is because $A$ is farther from $H$ than $D$ in the ordered distribution.

[14] English Premier League (England), La Liga (Spain), Bundesliga (Germany), Serie A (Italy), and Ligue 1 (France).

**Table 7.** RPS values for hypothetical match outcomes and match forecasts.

| Match outcome | Observed distribution $y(H)$, | $y(D)$, | $y(A)$ | Predictive distribution $p(H)$, | $p(D)$, | $p(A)$ | RPS |
|---|---|---|---|---|---|---|---|
| H | 1 | 0 | 0 | 1 | 0 | 0 | 0 |
| H | 1 | 0 | 0 | 0 | 1 | 0 | 0.5 |
| H | 1 | 0 | 0 | 0 | 0 | 1 | 1 |
| H | 1 | 0 | 0 | 0.75 | 0.25 | 0 | 0.03125 |
| H | 1 | 0 | 0 | 0.75 | 0.15 | 0.1 | 0.03625 |
| H | 1 | 0 | 0 | 0.5 | 0.3 | 0.2 | 0.145 |
| H | 1 | 0 | 0 | 0.5 | 0.2 | 0.3 | 0.17 |
| A | 0 | 0 | 1 | 1 | 0 | 0 | 1 |
| A | 0 | 0 | 1 | 0 | 1 | 0 | 0.5 |
| A | 0 | 0 | 1 | 0 | 0 | 1 | 0 |
| A | 0 | 0 | 1 | 0.75 | 0.25 | 0 | 0.78125 |
| A | 0 | 0 | 1 | 0.75 | 0.15 | 0.1 | 0.68625 |
| A | 0 | 0 | 1 | 0.5 | 0.3 | 0.2 | 0.445 |
| A | 0 | 0 | 1 | 0.5 | 0.2 | 0.3 | 0.37 |
| D | 0 | 1 | 0 | 0 | 1 | 0 | 0 |
| D | 0 | 1 | 0 | 0.25 | 0.5 | 0.25 | 0.0625 |
| D | 0 | 1 | 0 | 0.1 | 0.5 | 0.4 | 0.085 |

Now that we have a clear understanding of how the RPS functions, we can compare the RPS values obtained in this study with those from real-life football match forecasts to assess the significance and the importance of RPS discrepancies between models. Figure 21 presents the RPS values generated by the BN models and Hattrick predictors, applied to (a) the full set of 1 million samples as previously shown in Figure 19, and (b) the subset of 422 samples as previously shown in Figure 20. These scores are compared with the RPS values obtained by the models that participated in the 2017 International Machine Learning Soccer Competition, hosted by the *Machine Learning* journal. The comparison includes three baseline models: (a) Ignorant Prior models which assume a uniform distribution for predictions, (b) Global Prior models which reflect the average distribution of outcomes, and (c) a League Prior model applicable only to the international ML competition, which incorporate the average distribution per football league.

Figure 21 provides a useful performance comparison between predicting Hattrick matches and real-life matches. For instance, it is now evident that MMHC and GS perform almost identically to the Global Priors models; i.e., models that always predict the average distribution, confirming their poor performance due to the absence of edges in the BN models between variables of interest. Moreover, most models tested in this study produce lower RPS values than those observed in the real-life football forecasting models as part of the 2017 competition, suggesting that Hattrick match outcomes may be easier to predict than real-life football outcomes. Lastly, Figure 21 also facilitates the visualisation of model performance in clusters and highlights differences in performance between those clusters. For example, in the case of the 1M-sample dataset, the Nickarana and KB-probabilistic models form one cluster, followed by the Knowledge and KB-regression models, then the HC and TABU models, then the HC-Stable and TABU-Stable models, amongst others.

Lastly, the KB-regression model is shown to outperform all structural models learnt solely from data. This suggests that the information derived from the *Unwritten Manual*, on which the KB-regression model is based, is sufficiently accurate to provide a performance advantage. However, the KB-regression model is still outperformed by the KB-probabilistic model, which relies only on officially verified game mechanics, with the remaining relationships learnt from data. This result implies that the regression equations specified in the *Unwritten Manual* may be less accurate than the conditional probability distributions derived from data using the Beta-Binomial framework in the KB-probabilistic model.

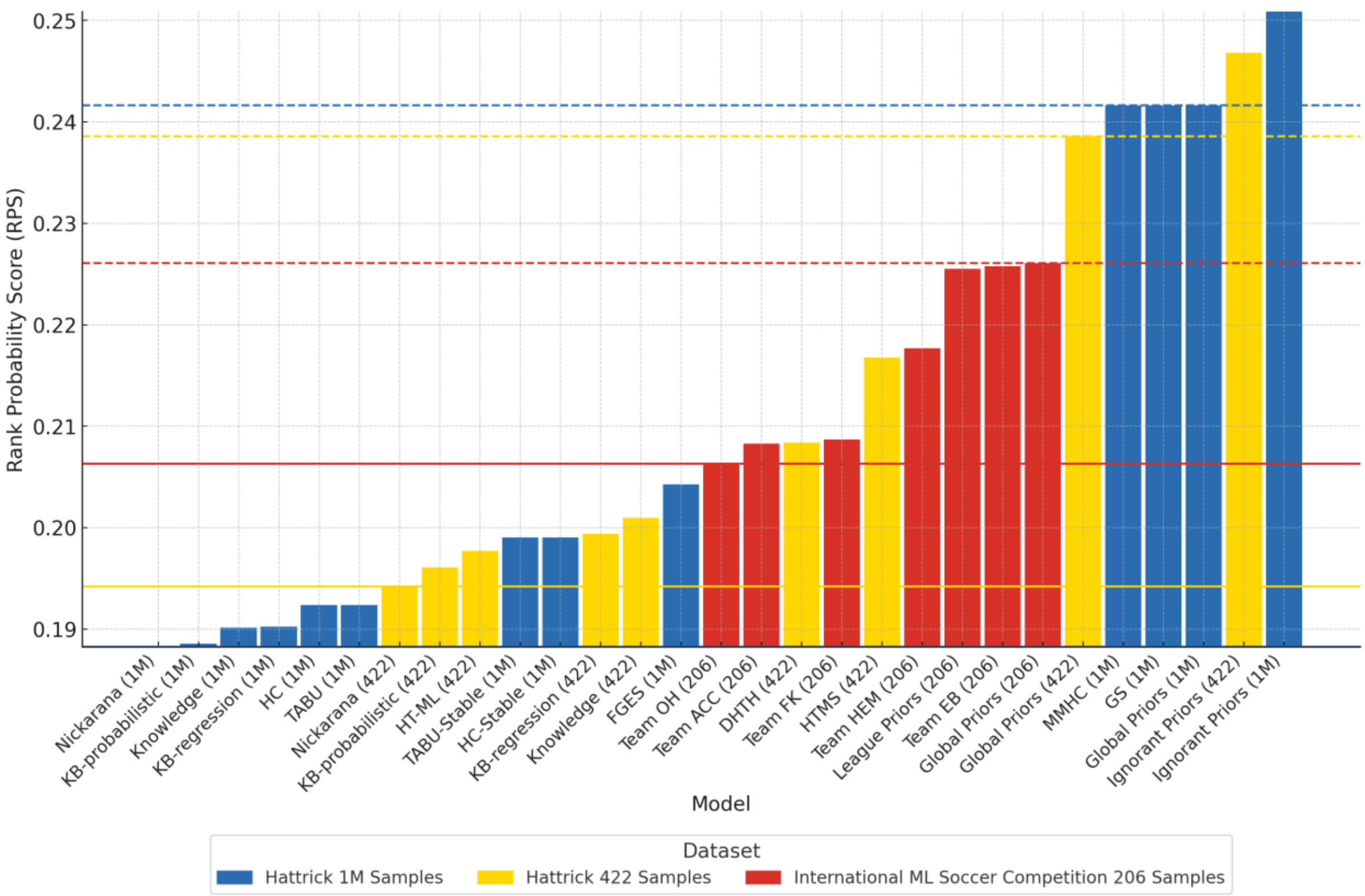

**Figure 21.** RPS values generated by BN and Hattrick models for the 1-million and 422-sample datasets, compared with RPS values from real-life football models tested during the 2017 Machine Learning for Soccer Competition, hosted by the *Machine Learning* journal. The *Ignorant Prior* models assume a uniform distribution for predictions, *Global Priors* assume the average distribution of outcomes for prediction, and *League Priors* (applicable only to the 206-sample dataset) assume the average distribution per football league. Dashed horizontal lines indicate the *Global Priors'* RPS value for each dataset, while solid lines represent the best-performing model within each dataset.

### *6.4. Using the BN model for decision support*

In this subsection, we illustrate how the KB-probabilistic model can be used for decision support by simulating the impact of various common decision options frequently debated within the Hattrick community. We begin with a brief theoretical explanation of how BNs can be used to estimate the effects of different decisions, and then proceed with a focused decision analysis.

In the context of BNs, decision analysis can be based on estimating what would happen if a particular action or intervention were taken. This relies on causal modelling assumptions, which allow the BN to represent not just correlations amongst variables but the direction of influence, and impact of decisions, between them. If we treat a BN model as an influence diagram, we are assuming that the network represents causal relationships, and we can use it to evaluate and choose optimal actions. However, given that we do not have access to the true game engine, we cannot be sure the KB-probabilistic model captures all causal relationships accurately. Thus, these decision analysis results should be interpreted as showing how the model can support decision under the assumed causal structure.

Building on the discussion in subsection 2.1, this process involves treating certain input variables as interventions rather than observations, and essentially simulating the effect of setting a variable to a specific value and observing how the rest of the network responds. We model this using the GeNIe BN software, which performs interventions by rendering the intervention or action on a node independent of their causes (parent nodes). In causal inference, this operation is formally represented using Pearl's $do$-operator, written as $do(X = x)$, which is a mathematical framework that asks the model to imagine that variable $X$ is externally fixed to value $x$, breaking any causal links that would normally determine $X$. The network can then propagate this change to estimate the resulting effects on downstream variables. Because the structure of the knowledge-based BNs tested in this subsection is manually constructed, the input nodes are already defined as background variables without parents, and therefore naturally act as decision options. As a result, the outcomes remain identical whether the model is regarded as a standard BN or as a causal BN with explicit interventions.

To begin, we define a set of three opponent profiles as case studies for decision support. These are described in Table 8 and represent three typical teams employing different strategies including *Normal* (NM), Counterattacks (CA), and Long shots (LS), commonly seen at higher levels of Hattrick play. The NM tactic is the most widely used by managers. In contrast, the CA and LS tactics, while less frequently chosen, are often the focus of strategic discussions within the Hattrick community, particularly around how to use and counter them effectively. One might ask how these profiles are represented within the BN model. As an example, and with reference to the 30 input variables described in Table A.2 for each team, if we enter the values for the NM profile as the away team, the resulting 30 variable assignments would appear as shown in the accompanying footnote[15].

Figure 22 illustrates how the probability of winning (left charts) and losing (right charts) changes when the same opponent profiles from Table 8 are used, but specific variable values, such as team ratings or tactic skills, are varied. The top two charts focus on matches between two teams using the NM profile. When both teams have identical ratings, the probability of winning or losing is approximately 42%, as shown at the point where the $x$-axis value is 0 (indicating no rating change for the user's team). However, these probabilities shift depending on changes to specific rating levels. In this scenario, the results show that increasing the *Midfield* rating by +2 leads to a significantly higher probability of winning (~57%) compared to increasing either the *Attack* or *Defence* ratings by the same amount (~49%), relative to the baseline of ~42%. This demonstrates that, under this tactical scenario, *Midfield* improvements yield the highest marginal benefit, offering t insight into which ratings to prioritise when making trade-offs.

[15] *Tactic_AT=Normal, Tactic_skill_AT=0, Left_defence_rating_AT=15, Middle_defence_rating_AT=15, Right_defence_rating_AT=15, Left_attack_rating_AT=15, Middle_attack_rating_AT=15, Right_attack_rating_AT=15, Midfield_rating_AT=15, IFK_attack_rating_AT= 10, IFK_defence_rating_AT=15, Unpredictable_SA_players_AT=3, Unpredictable_LP_players_AT=3, Unpredictable_Mistake_players_AT=2, Unpredictable_owngoal_players_AT=2, Quick_defenders_AT=One, Technical_defenders_AT=No, PDIM_AT=No, Quick_offensives_AT=2, Technical_offensives_AT=1, Head_defenders_or_IMs_AT=Yes, Head_offensives_AT=Two, Corner_Head_defensives_AT=1, Corner_Head_offensives_AT=2, Unpredictable_offensives_AT=2, PNF_AT=1, Central_defenders_AT=One, CDs_and_WBs_AT=3, Wingers_AT=2, Forwards_AT=2.*

**Table 8.** Three commonly used team profiles frequently debated within the Hattrick community, *Normal* (NM), counterattack (CA), and long shots (LS), which are used for illustration in the decision analysis.

| Input | Assumed values for Profile NM | Assumed values for Profile CA | Assumed values for Profile LS |
|---|---|---|---|
| Tactic | Normal | Counterattacks | Long shots |
| Tactic skill | n/a | 20 | 20 |
| Defence | 15 | 20 | 20 |
| Attack | 15 | 15 | 5 |
| Midfield | 15 | 10 | 15 |
| ISP-defence | 15 | 15 | 20 |
| ISP-attack | 10 | 10 | 15 |
| Specialities | 1x Unpred. defender,<br>1x Powerful defender,<br>1x Quick defender,<br>0x Technical defender,<br>0x PDIM,<br>0x Head defender,<br>2x Quick offensives,<br>1x Technical offensives,<br>1x Head inner midfielder,<br>2x Unpred. offensives,<br>1x PNF. | 1x Unpred. defender,<br>1x Powerful defender,<br>1x Quick defender,<br>1x Technical defender,<br>0x PDIM,<br>0x Head defender,<br>2x Quick offensives,<br>1x Technical offensives,<br>0x Head offensives,<br>2x Unpred. offensives,<br>1x PNF. | 1x Unpred. defender,<br>1x Powerful defender,<br>1x Quick defender,<br>1x Technical defender,<br>1x PDIM,<br>1x Head defender,<br>1x Quick offensives,<br>0x Technical offensives,<br>0x Head offensives,<br>2x Unpred. offensives,<br>0x PNF. |
| Player positions | 1x Central defender,<br>2x Wing Backs<br>2x Wingers,<br>3x Midfielders,<br>2x Forwards. | 2x Central defenders,<br>2x Wing Backs<br>2x Wingers,<br>2x Midfielders,<br>2x Forwards. | 3x Central defenders,<br>2x Wing Backs<br>2x Wingers,<br>3x Midfielders,<br>0x Forwards. |

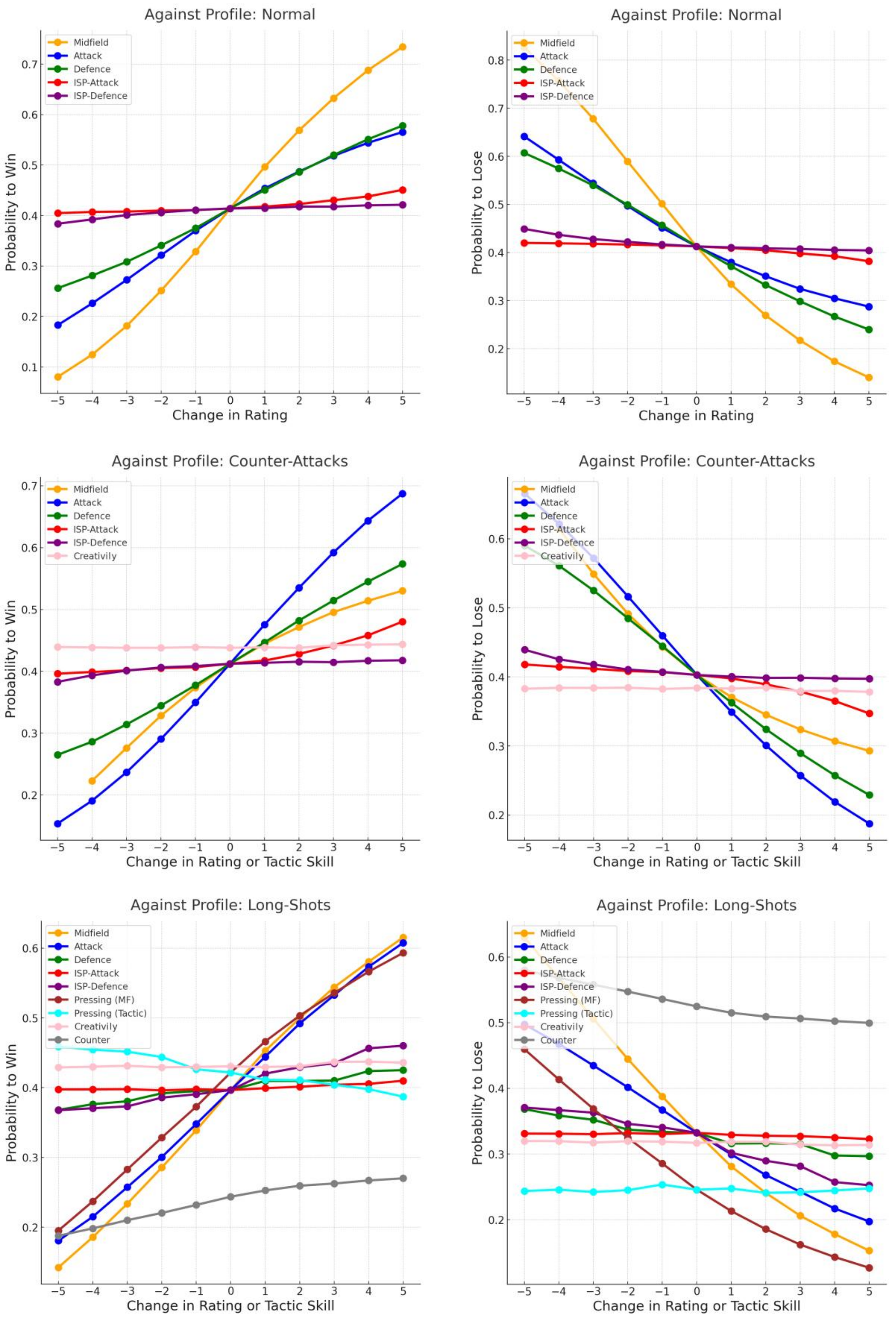

**Figure 22.** Results from simulated decision analyses based on the three team profiles described in Table 8: Normal (NM) tactic profile (top left), Counterattack (CA) profile (top right), and Long Shots (LS) profile (bottom). The results illustrate the impact of varying specified team ratings or tactic skills for each team profile. The default tactic skill levels (i.e., at x-axis value 0) for pressing, counterattacks, and creatively are 10, 15, and 15, respectively.

Similarly, the middle charts focus on opponents using a counterattacking profile, while the controlled ratings still refer to the team playing the NM tactic, as in the previous example, based on the NM and CA profiles in Table 8. However, in this case, we also include the PC (Play Creatively), meaning that the $x$-axis in this case represents either changes in ratings or tactic level, depending on the variable being analysed. Specifically, for the PC tactic, it is the tactic skill that varies along the $x$-axis, from a level of 10 to 20 (default assumed is 15). Under this scenario, we observe that attack ratings are by far the most influential, followed by defence ratings, which have a tendency to be underestimated by managers, and then midfield ratings. Additionally, ISP attack ratings show a meaningful impact when increased from 10 to 15, whereas increasing ISP defence or playing PC (without considering its negative impact on defence ratings) has only marginal effects.

Lastly, the bottom charts repeat this analysis against LS teams, incorporating a few additional relevant tactics. The results indicate that midfield ratings, attack ratings, and the pressing tactic are the most significant factors influencing a team's probability of winning or losing. Pressing, which is widely considered an effective counter to the LS tactic, is evaluated in two ways: (a) Pressing (MF), which assumes a fixed pressing skill of 10 with the midfield rating varying, and (b) Pressing (Tactic), which assumes the default midfield rating of 15 as static, with the pressing skill varying. The results suggest that increasing attack or midfield ratings, with or without pressing (i.e., playing *Normal*), has the greatest impact on improving the chances of winning. Interestingly, while increasing pressing skill generally improves the probability of winning, very high pressing values reduce the win probability below the baseline of 40% for this scenario. This is because pressing substantially increases the probability of a draw, as shown in the bottom right figure, where any level of pressing skill keeps the opponent's win probability just below 25%, which is significantly lower than the baseline of around 33% for this scenario. However, it is important to note that this pressing simulation does not account for its negative impact on *Stamina*, which is a dynamic factor that influences how quickly players fatigue and lose ratings earlier in a match (see Limitations in Section 6). As a result, the impact of pressing is somewhat overestimated in the context of this analysis.

## 7. Discussion and Limitations

### *7.1. The Hattrick case study*

This is the first academic study to investigate the game-engine of Hattrick, a football management simulation that has existed since 1997 and currently has an active user base of 200,000. Many of these users have been engaged for decades, demonstrating a strong interest in probability and statistics and a desire to better understand the game mechanics. This research provides insights that can help players enhance their decision-making as they manage digital football teams, which take real-life years to build and develop.

We use both manually constructed and automated BN models to disentangle the game-engine, providing interpretable graphical fragments that explain its different components. These models highlight the relationships between key nodes, offering users a clearer understanding of how various aspects of the game-engine interact, through visual illustrations and performance analyses. We use the learnt models to infer different outcomes of interest and evaluate their performance in predicting these outcomes. Additionally, we compare the four most widely used predictive models in the Hattrick community against the BN models learnt in this study, assessing their effectiveness in predicting both goals scored and final match outcomes

A total of 11 BN models were evaluated on a dataset with 250 variables and 1 million samples. Excluding the time required for structure learning and parameterisation (discussed below), most BN models took approximately three days of processing (each BN model) on a single computer to generate 1 million inferences, which resulted in over a month of continuous computation on a single computer. The BN models perform poorly in terms of efficiency compared to the Nickarana model, which generated 1 million predictions in just 2 minutes. This result exposes the high computational complexity of exact Bayesian inference engines in probabilistic graphical models, at least when compared with rule-based and equation-based models that operate on point values rather than distributions, which often perform well for prediction, as is the case in this study.

The structure of the best-performing structural model was used to conduct decision analysis on common strategic choices debated within the Hattrick community. This allows us to estimate the impact of decisions involving rating exchanges between sectors and tactic skills. We illustrate this capability by focusing on three common adversary rating profiles, demonstrating how different decision options affect a team's chances of winning or losing a match. As a result, this part of the study generates new, previously unseen decision-analytic insights.

For decades, the Hattrick community has debated whether the game's match engine produces results that are more unpredictable than real-world football. The findings of this study, however, decisively refute that claim. This is because the models tested here generated RPS scores with lower error rates than those typically achieved in football forecasting literature. Consider the best-performing Hattrick and BN models examined in this study: 0.1883 and 0.1886 RPS scores, respectively (see Figure 19). These compare favourably to the top-performing football forecasting models in international machine learning competitions, which achieve RPS scores just below 0.21 [43, 51]. If the Hattrick game-engine were inherently more chaotic than real-life football, the Hattrick models assessed in this study would not achieve considerably higher predictive accuracy than real-life football forecasting models. In fact, this outcome is inevitable given that the average number of goals in Hattrick is known to be higher than in real football. This reduces the likelihood of draws while proportionally increasing home and away wins, making it easier to predict winners.

Why, then, does the perception of unpredictability persist within the Hattrick community? The answer appears to lie in the mathematics of large numbers. In real-world football, the number of elite teams genuinely capable of winning the most prestigious trophies is relatively small. In Hattrick, however, with hundreds of thousands of teams, there are typically hundreds of contenders for the top honours. This disparity intensifies across different brackets of team strength, leading to an abundance of matches that are, in statistical terms, close to a coin toss. The issue, therefore, is not with the game-engine itself, since the engine functions as intended. The perception of excessive randomness arises from the sheer volume of competitive teams, which naturally amplifies the frequency of unpredictable results. Perhaps a more dynamic managing system, one that rewards users based on how accurately they predict their opponent's style of play, would empower users to make more strategic decisions. Granting users greater control over match outcomes would not only improve the tactical landscape, but also encourage greater diversity in playstyles and likely higher user engagement.

### 7.2. Structure learning and Bayesian networks

With respect to structure learning, this study makes three key contributions. First, we assess the ability of commonly used constraint-based and score-based structure learning algorithms to recover the game-engine of the Hattrick football manager game, using a manually constructed graph based on expert knowledge as a reference.

The results reveal that only a few structure learning algorithms could be applied to such a large dataset, as most we tested failed to complete within the 48-hour runtime limit. This highlights the practical limitations of many existing algorithms when dealing with high-dimensional, large-scale data. In particular, structure learning algorithms with high computational complexity become infeasible under such constraints. These findings reinforce the ongoing need for more scalable structure learning methods that can maintain accuracy without excessive computational cost. Interestingly, and somewhat counterintuitively, part of the structure learning literature suggests that some algorithms can be both more efficient and more effective than others [52, 53]. Specifically, early versions of HC and Tabu Search, along with their improved stable variants (HC-Stable and Tabu-Stable [33, 34]), stand out for achieving efficient and state-of-the-art performance. The results from this study support this view and challenge the common assumption in structure learning that greater accuracy typically comes at the cost of efficiency. In contrast, our results show that algorithms such as MMHC and GS, which optimise for the SHD score and tend to produce sparser graphs, did not perform well in practice at this scale, consistent with other studies within the structure learning literature [52, 53].

Overall, the structure learning algorithms performed moderately well (but MMHC and GS poorly) in recovering a structure that could achieve inferential performance comparable to the knowledge-based structure. This is particularly noteworthy given that the data are partially synthetic (game-generated) rather than real-world observations, which are typically subject to data noise that further hinders the performance of such algorithms. Although knowledge-based models are found to be superior in the context of this case study, it is important to acknowledge their limitations, particularly the significant effort required to construct them. This challenge becomes even more pronounced when dealing with hundreds of variables, as was the case in this study, where finalising the overall structure took many months of work, even someone with decades of experience playing this game. This highlights the need for further, more pronounced, improvements in structure learning or other forms of automation that would result in a more efficient process of constructing such models.

Second, it is commonly reported in other studies that knowledge graphs perform poorly in terms of Log-Likelihood and objective scores such as BIC, when compared to structures learnt from data, leading to the expectation that inferential performance from knowledge graphs will also be inferior. This expectation is reasonable, as structure learning algorithms are designed to optimise fitting across all nodes, versus dimensionality. While our findings support this general trend, showing that graphs produced by structure learning algorithms generally achieve higher fitting scores than knowledge graphs, they also show that this higher overall network fit did not translate into better inferential performance in practice. In other words, a higher overall network fit does not necessarily lead to more accurate predictions for selected variables of interest. Instead, all three knowledge-based models explored outperformed every single BN model learnt entirely from data across a set of six output nodes of interest. This highlights the trade-off that occurs when optimising across an entire network (of 250 nodes in this case), as it may come at the cost of optimising the specific outcomes of interest.

While this study models outcomes at the team-sector level, in line with Hattrick’s game mechanics, other football management games such as Football Manager capture detailed data at the individual player level. The modelling approach presented here could be naturally extended to represent player-specific attributes, skills, and interactions, enabling more granular analyses of player development, tactical decisions, and match performance. The interpretable structure of BNs makes them particularly well suited to this type of modelling, as they allow the integration of knowledge with data-driven learning, and support both inference and decision analysis. Previous research on football games or sports has explored community behaviour, decision-making strategies, and player or team performance using simulation data [54, 55, 56]. While these studies have largely used statistical or machine learning approaches, BNs and structure learning techniques offer a promising framework for capturing the complex, probabilistic relationships underlying these games.

### *7.3. Limitations*

This study comes with limitations both in relation to the specific case study and the broader application of structure learning methods. A key limitation concerns the nature of the game modelled. That is, the actual gameplay is dynamic, with events occurring in 5-minute intervals over a full 90-minute period. However, the models developed in this study (but also the community models tested) assume a static representation, due to the constraints of the available data reflecting only average team performance across the entire match. While we do not expect this simplification to meaningfully impact the results, under the assumption that a team's average performance over the 90-minute match is a good estimator of its overall performance, it does limit the model's ability to capture the temporal dynamics of the game engine. Access to more granular, dynamic time-stamped data, would enable us to explore whether structure learning can effectively recover such temporal patterns, which remains an open and promising direction for future research [57, 58]. While Dynamic BNs (DBNs) may seem like a natural extension for modelling temporal dynamics [59], they are rarely used in football simulations due to significant modelling and computational limitations, and structure learning algorithms do not natively support them, making DBNs difficult to apply in practice [57]. As a result, past football modelling studies, both in gaming and traditional sports analytics, typically rely on evolving rating systems such as Elo [60], pi-ratings [61], and Glicko [62], combined with probabilistic models to capture time-dependent changes in team or player strength.

Another limitation involves the absence of a ground truth; i.e., access to the actual game engine. To address this, we performed comparative analysis with existing community-developed Hattrick predictors. However, of the four Hattrick predictors considered, only one could be feasibly applied to the full dataset of one million samples. The remaining three required manual implementation, which limited their application to a smaller subset of 422 samples due to time and resource constraints. As a result, we report model performance in two parts: one based on the full dataset and another on the smaller subset. While not ideal, this approach still provides valuable insights into how the models compare, both with each other and with those introduced in this study. Notably, the Hattrick predictor that was applicable to the full dataset (Nickarana), was also the best-performing amongst the four. This partially mitigates the limitation by ensuring that the broadest comparisons involve the strongest available baseline from the Hattrick predictors.

Lastly, the findings presented in this study are specific to the Hattrick football manager and may not directly generalise to other football management games. Hattrick's relative structural transparency and the accessibility of community knowledge provide particularly favourable conditions for this type of modelling and validation. Nevertheless, the methodological framework proposed in this paper, which combines expert-elicited knowledge-based models, data-driven structure learning, and comparative evaluation using structural and predictive metrics, could be adapted to other games with opaque or hidden mechanics, such as football management games or similar simulation-based systems. Such extensions would require domain-specific variable definitions, expert knowledge elicitation, and access to suitable in-game observational data, as well as careful consideration of game-specific mechanics and constraints.

## 8. Concluding remarks

This study presents the first academic investigation of the game-engine behind Hattrick, a free, web-based probabilistic football manager game launched in Sweden in 1997 as part of an MSc project. With over 200,000 users, Hattrick's slow-paced design has promoted a loyal community and long-term engagement.

We combine existing user knowledge with structure learning to provide a rigorous, data-driven methodology for reverse-engineering the Hattrick game-engine, using interpretable probabilistic BNs, offering interpretable insights and predictive performance that match or exceed those of the most widely used community-developed predictors. This is an approach applicable to other games or simulation-based environments, where internal mechanics are hidden but large-scale observational data is available. In doing so, we not only benchmark predictive accuracy but also demonstrate how the optimal BN model supports a deeper understanding of the game-engine through graphical representations of dependencies and decision-support analyses.

The results demonstrate how structure learning algorithms perform when applied to complex gaming datasets at scale, containing hundreds of variables and millions of samples, adding empirical evidence to ongoing discussions in the causal discovery literature. A key relevant finding of this study involves the assessment of structure learning algorithms relative to knowledge-based structures. While machine-learnt models do achieve better global fit, as expected, we find that knowledge-based models offer superior prediction for targeted outcomes, highlighting a trade-off between overall model optimisation and task-specific performance.

Our findings also contribute to an ongoing debate within the Hattrick community regarding the perceived unpredictability of match results. While some users argue the game is overly random, our results suggest that Hattrick matches carry less uncertainty than real-world football. This is largely due to higher average goal counts in Hattrick, which reduce draw frequency and increase the predictability of match winners.

Finally, to support future research, we make the dataset, the knowledge graph, and all BN models publicly available through the Bayesys repository. These data and graphical structures can serve as a benchmark for structure learning and causal discovery algorithms, facilitating comparative studies in causal modelling and inference. Moreover, the BN models provide Hattrick users with the ability to analyse in-game decisions and offer an opportunity to explore BNs and experiment with the models and data.

### Acknowledgements

The authors acknowledge the use of the GeNIe Bayesian Network software, provided free of charge for academic use by BayesFusion LLP, USA. The software played a key role in facilitating the modelling and analysis conducted in this study.

### CRediT author statement

**Anthony Constantinou:** Conceptualisation, Methodology, Software, Validation, Formal analysis, Data Curation, Writing - Original Draft, Visualisation, Supervision. **Nicholas Higgins:** Conceptualisation, Formal analysis, Data Curation, Writing - Review & Editing: **Ken Kitson:** Formal analysis, Writing - Review & Editing.

## **Appendix A:** Data variables for structure learning models, and Bayesian network nodes for knowledge-based models

**Table A.1.** Description of the 250 data variables used for structure learning, where *HT* is *Home Team* and *AT* is *Away Team,* indicating variables that are repeated for each team. The Input variables represent those which are always observed from data during the experiments, whereas the Output variables represent the inferred variables.

| **Variables** | **Description** | **# of states in discretised form** | **I/O** |
|---|---|---|---|
| **Core game-engine variables** | | | |
| RatingLeftAtt_HT/AT | The defence, midfield, and attack ratings for each team per sector. | 33 | Input |
| RatingLeftDef_HT/AT | | 33 | |
| RatingMidAtt_HT/AT | | 33 | |
| RatingMidDef_HT/AT | | 33 | |
| RatingMidfield_HT/AT | | 29 | |
| RatingRightAtt_HT/AT | | 33 | |
| RatingRightDef_HT/AT | | 33 | |
| RatingISPAtt_HT/AT | The attack and defence ratings for indirect set-pieces. | 22 | Input |
| RatingISPDef_HT/AT | | 22 | |
| Event_Normal_ChanceLeft_HT/AT | The number of attacks generated, and goals scored, in each sector. | 6 | Output |
| Event_Normal_ChanceLeft_Goal_HT/AT | | 5 | |
| Event_Normal_ChanceLRM_Goal_HT/AT | | 7 | |
| Event_Normal_ChanceMid_HT/AT | | 7 | |
| Event_Normal_ChanceMid_Goal_HT/AT | | 5 | |
| Event_Normal_ChanceRight_HT/AT | | 6 | |
| Event_Normal_ChanceRight_Goal_HT/AT | | 5 | |
| Event_Normal_Goal_HT/AT | | 7 | |
| Event_Normal_HT/AT | | 11 | |
| Event_Normal_ChanceDFK_HT/AT | The number of indirect and direct set-pieces, and penalty kicks, generated and scored in each sector. | 3 | Output |
| Event_Normal_ChanceDFK_Goal_HT/AT | | 3 | |
| Event_Normal_ChanceIFK_HT/AT | | 3 | |
| Event_Normal_ChanceIFK_Goal_HT/AT | | 3 | |
| Event_Normal_ChanceISP_Goal_HT/AT | | 3 | |
| Event_Normal_ChancePK_HT/AT | | 3 | |
| Event_Normal_ChancePK_Goal_HT/AT | | 3 | |
| Event_Normal_ChanceLSNonTactic_HT/AT | The number of events and goals scored through non-tactical long shots. | 2 | Output |
| Event_Normal_ChanceLSNonTactic_Goal_HT/AT | | 2 | |
| Possession_HT | The probability to win possession of the ball, or score a goal in a given sector or through set-pieces. The rate of possession for AT is the complement of HT. | 31 | Output |
| Score_left_HT/AT | | 31 | |
| Score_middle_HT/AT | | 31 | |
| Score_right_HT/AT | | 31 | |
| Score_ISP_HT/AT | | 20 | |
| Goals HT/AT | The number of goals scored. | 8 | Output |
| HDA | The match outcome in terms of home win, draw, and away win. | 3 | Output |
| **Tactic variables** | | | |
| TacticType_HT/AT | The 7 tactics that influence the style of play; Attack in the Middle (AiM), Attack on the Wings (AoW), Counterattack (CA), Play Creatively (PC), Long-Shots (LS), and pressing (PR). | 7 | Input |
| AiM_AoW_tactic_HT/AT | | 3 | |
| CA_tactic_HT/AT | | 2 | |
| Creatively_tactic_HT/AT | | 2 | |
| Longshots_tactic_HT/AT | | 2 | |
| Pressing_tactic | | 3 | |
| TacticSkill_HT/AT | The skill level of the tactic. | 11 | Input |
| Event_CA_HT/AT | The variables that capture the events related to the CA tactic. | 5 | Output |
| Event_CA_ChanceDFK_HT/AT | | 2 | |
| Event_CA_ChanceDFK_Goal_HT/AT | | 3 | |
| Event_CA_ChanceIFK_HT/AT | | 2 | |
| Event_CA_ChanceIFK_Goal_HT/AT | | 2 | |
| Event_CA_ChanceISP_Goal_HT/AT | | 2 | |
| Event_CA_ChanceLeft_HT/AT | | 2 | |
| Event_CA_ChanceLeft_Goal_HT/AT | | 2 | |
| Event_CA_ChanceLMR_Goal_HT/AT | | 4 | |
| Event_CA_ChanceMid_HT/AT | | 2 | |
| Event_CA_ChanceMid_Goal_HT/AT | | 2 | |
| Event_CA_ChanceRight_HT/AT | | 2 | |
| Event_CA_ChanceRight_Goal_HT/AT | | 2 | |
| Event_CA_Goal_HT/AT | | 4 | |
| Nornal_chances_missed_HT/AT | | 8 | |
| Event_Normal_ChanceLS_HT/AT | The variables that capture the events related to the LS tactic. | 5 | Output |
| Event_Normal_ChanceLS_Goal_HT/AT | | 4 | |
| Event_Normal_ChanceLS_all_Goal_HT/AT | | 4 | |
| **Special events variables** | | | |
| Unpredictable_defenders_HT/AT | The number of players related to the player speciality *Unpredictable*. | 7 | Input |
| Unpredictable_mistake_players_HT/AT | | 4 | |
| Unpredictable_offensives_HT/AT | | 9 | |
| Unpredictable_own_goal_players_HT/AT | | 4 | |
| Unpredictable_players_HT/AT | | 12 | |
| Unpredictable_special_action_players_HT/AT | | 5 | |
| Quick_defenders_HT/AT | The number of players related to the player speciality *Quick*. | 4 | Input |
| Quick_offensives_HT/AT | | 9 | |
| Tech_defenders_HT/AT | The number of players related to the player speciality *Technical*. | 4 | Input |
| Technical_offensives_HT/AT | | 9 | |
| Head_offensives_HT/AT | The number players related to the player speciality *Head*. | 9 | Input |
| Head_opposing_players_HT/AT | | 5 | |
| Corner_headers_defence_HT/AT | | 6 | |
| Corner_headers_offence_HT/AT | | 6 | |
| PDIMs_HT/AT | The number of players playing as Power Defensive Inner Midfielders (PDIM) and Powerful Normal Forwards (PNFs). | 3 | Input |
| PNFs_HT/AT | | 3 | |
| CDs_and_WBs_HT/AT | The number of players playing as Central Defenders (CD), Wing Backs (WBs), forwards, and wingers. | 4 | Input |
| CDs_HT/AT | | 4 | |
| Forwards_HT/AT | | 4 | |
| Wingers_HT/AT | | 3 | |
| PSE_player_count_HT-AT | The number of players with any speciality, and the difference between the total player specialities between the two teams. | 23 | Input |
| PSE_spec_player_count_HT/AT | | 12 | |
| Event_PSE_Unpred_Goals_HT/AT | The events related to the player speciality *Unpredictable*. | 3 | Output |
| Event_PSE_UnpredLongPass_HT/AT | | 2 | |
| Event_PSE_UnpredLongPass_Goal_HT/AT | | 2 | |
| Event_PSE_UnpredMistake_HT/AT | | 2 | |
| Event_PSE_UnpredMistake_OwnGoal_HT/AT | | 2 | |
| Event_PSE_UnpredOwnGoal_HT/AT | | 2 | |
| Event_PSE_UnpredOwnGoal_OwnGoal_HT/AT | | 2 | |
| Event_PSE_UnpredScoreOwn_HT/AT | | 2 | |
| Event_PSE_UnpredScoreOwn_Goal_HT/AT | | 2 | |
| Event_PSE_UnpredSpecial_HT/AT | | 2 | |
| Event_PSE_UnpredSpecial_Goal_HT/AT | | 2 | |
| PSE_unpred_chances_missed_HT/AT | | 2 | |
| Event_PSE_Quick_Goal_HT/AT | The events related to the player speciality *Quick*. | 3 | Output |
| Event_PSE_QuickPass_HT/AT | | 2 | |
| Event_PSE_QuickPass_Goal_HT/AT | | 2 | |
| Event_PSE_QuickRush_HT/AT | | 2 | |
| Event_PSE_QuickRush_Goal_HT/AT | | 2 | |
| Event_PSE_QuickStop_HT/AT | | 2 | |
| Event_PSE_TechHead_HT/AT | The events related to the player speciality *Technical*. | 2 | Output |
| Event_PSE_TechHead_Goal_HT/AT | | 2 | |
| Event_SEZ_TechCA_HT/AT | | 2 | |
| Event_PSE_WingHead_HT/AT | Whether a team fielded defenders or midfielders with the speciality *Head*. | 2 | Output |
| Event_PSE_WingHead_Goal_HT/AT | | 3 | |
| Event_SEZ_PDIM_HT/AT | The events related to the player specialities *Powerful*. | 3 | Output |
| Event_SEZ_PNF_HT/AT | | 2 | |
| Event_SEZ_PNF_Goal_HT/AT | | 3 | |
| Event_PSE_Wing_Goal_HT/AT | The events related to wing-based special events. | 5 | Output |
| Event_PSE_WingAnyone_HT/AT | | 2 | |
| Event_PSE_WingAnyone_Goal_HT/AT | | 3 | |
| Event_TSE_Corner_Goal_HT/AT | The events related to corner-based special events. | 3 | Output |
| Event_TSE_CornerAnyone_HT/AT | | 2 | |
| Event_TSE_CornerAnyone_Goal_HT/AT | | 3 | |
| Event_TSE_CornerHead_HT/AT | | 2 | |
| Event_TSE_CornerHead_Goal_HT/AT | | 3 | |
| Event_TSE_ExperienceFwd_HT/AT | The events related to experience-based special events. | 2 | Output |
| Event_TSE_ExperienceFwd_Goal_HT/AT | | 2 | |
| Event_TSE_InexpDef_HT/AT | | 2 | |
| Event_TSE_InexpDef_Goal_HT/AT | | 2 | |
| Event_TSE_TiredDef_HT/AT | | 2 | |
| Event_TSE_TiredDef_Goal_HT/AT | | 2 | |
| Event_PSE_HT/AT | The number events related to both Player-based Special Events (PSE), and Team-based Special Events (TSE), as well as all Special Events (SE). | 4 | Output |
| Event_PSE_OwnGoal_HT/AT | | 2 | |
| Event_PSE_Goal_HT/AT | | 3 | |
| Event_TSE_HT/AT | | 3 | |
| Event_TSE_Goal_HT/AT | | 3 | |
| Event_SE_Goal_HT/AT | | 4 | |

**Table A.2.** Description of the 363 nodes, grouped into core or mirroring nodes and omitting some auxiliary nodes for simplicity, that constitute the KB-probabilistic model, where *HT* represents the *Home Team* and *AT* represents the *Away Team*. Uniform distributions are assumed for all input variables, which are always observed from data in the experiments. Variable types indicated as *Equations* specify relationships through an in-game equation, as described in Section 4.

| Node/s | Description | Variable type | I/O |
|---|---|---|---|
| **Core game-engine variables** | | | |
| Left_attack_rating_HT/AT<br>Middle_attack_rating_HT/AT<br>Right_attack_rating_HT/AT | The attack ratings for each sector. | $\sim Uniform(a, b)$ | Input |
| Left_defence_rating_HT/AT<br>Middle_defence_rating_HT/AT<br>Right_defence_rating_HT/AT | The defence ratings for each sector. | $\sim Uniform(a, b)$ | Input |
| Midfield_rating_HT/AT | The midfield ratings. | $\sim Uniform(a, b)$ | Input |
| IFK_attack_rating_HT/AT<br>IFK_defence_rating_HT/AT | The attack and defence ratings for indirect set-pieces. | $\sim Uniform(a, b)$ | Input |
| Left_attacks_HT/AT<br>Middle_attacks_HT/AT<br>Right_attacks_HT/AT | The number of attacks generated in each sector. | $\sim Binomial(n, p)$ | Output |
| Set-pieces_HT/AT<br>DFKs_HT/AT<br>IFKs_HT/AT<br>PKs_HT/AT | The number of set-piece attacks, distributed across penalty kicks, and direct and indirect set-pieces, in each sector. | $\sim Binomial(n, p)$ | Output |
| Possession_HT | The rate of possession (for AT it is the complement of HT). | Equation | Output |
| Exclusive_chances_HT/AT<br>Shared_chances_HT/AT | The number of exclusive and shared chances allocated to each team. | $\sim Binomial(n, p)$, Equation | Output |
| Normal_chances_HT/AT | The number of chances classified as *Normal*, allocated to each team. | $\sim Binomial(n, p)$, Equation | Output |
| Score_left_HT/AT<br>Score_middle_HT/AT<br>Score_right_HT/AT | The probability of scoring from an attack in each sector. | Equation | Output |
| ISP_rating_dif_HT/AT | The difference between ISP attack and defence ratings between the teams. | Equation | Output |
| Score_DFKs_HT/AT<br>Score_IFKs_HT/AT<br>Score_PKs_HT/AT | The probability to score from the different set-piece events. | $\sim Beta(a, \beta)$ | Output |
| Left_goals_HT/AT<br>Middle_goals_HT/AT<br>Right_goals_HT/AT | The number of goals scored in each sector. | $\sim Binomial(n, p)$ | Output |
| Set-Pieces_goals_HT/AT<br>DFKs_goals_HT/AT<br>IFKs_goals_HT/AT<br>PKs_goals_HT/AT | The number of goals from the different set-piece events. | $\sim Binomial(n, p)$ | Output |
| Normal_goals_HT/AT | The number of goals scored from chances classified as *Normal*. | Equation | Output |
| Goals_HT/AT | The number of goals scored. | Equation | Output |
| HDA | The match outcome in terms of home win, draw, and away win. | Discrete | Output |
| **Tactic variables** (includes related non-tactical variables) | | | |
| Tactic_HT/AT | The 7 tactics that influence the style of play. | Discrete | Input |
| Tactic_skill_ HT/AT | The skill level of the tactic. | $\sim Uniform(a, b)$ | Input |
| p(AiM)_HT/AT | The probability to convert an attack on the sides into an attack in the middle. | $\sim Beta(a,\beta)$ | Output |
| p(AoW)_HT/AT | The probability to convert an attack in the middle into an attack on the sides. | $\sim Beta(a,\beta)$ | Output |
| p(Longshot)_HT/AT<br>p(non_tactical_Longshot)_HT/AT | The tactical and non-tactical probabilities to convert a normal attack into a long-shot attack. | $\sim Beta(a,\beta)$ | Output |
| p(press)_HT/AT | The probability to press a normal or a long-shot attack. | $\sim Beta(a,\beta)$ | Output |
| p(CA)_HT/AT<br>p(non_tactical_CA)_HT/AT | The tactical and non-tactical probabilities to convert an opponent's normal missed chance into a counterattack. | $\sim Beta(a,\beta)$ | Output |
| Left_attacks_given_[tactic]_HT/AT<br>Middle_attacks_given_[tactic]_HT/AT<br>Right_attacks_given_[tactic]_HT/AT | The number of attacks generated in each sector given the tactic. | $\sim Binomial(n,p)$, Equation | Output |
| Non_tactical_counter_attacks_HT<br>Tactical_counterattacks_HT/AT<br>All_counterattacks_HT/AT | The number of non-tactical, tactical and all counterattacks. | $\sim Binomial(n,p)$, Equation | Output |
| Missed_normal_chances_HT/AT | The number of opponent attacks that did not lead to a goal. | Equation | Output |
| Left_counterattacks_HT/AT<br>Middle_counterattacks_HT/AT<br>Right_counterattacks_HT/AT<br>SPs_counterattacks_HT/AT<br>DFKs_ counterattacks_HT/AT<br>IFKs_ counterattacks_HT/AT | The number of counterattacks generated in each sector. | $\sim Binomial(n,p)$ | Output |
| Left_counter_goals_HT/AT<br>Middle_counter_goals_HT/AT<br>Right_counter_goals_HT/AT<br>SPs_counter_goals__HT/AT<br>DFKs_ counter_goals__HT/AT<br>IFKs_ counter_goals__HT/AT<br>CA_goals__HT/AT | The number of goals scored in each sector, or through set-pieces, as a result of a counterattack. | $\sim Binomial(n,p)$, Equation | Output |
| Longshots_HT/AT<br>Non_tactical_longshots_HT<br>Longshots_given_pressing_HT/AT | The number of tactical and non-tactical long-shot attacks, with and without pressing | $\sim Binomial(n,p)$ | Output |
| p(Long_shot_score)_HT/AT | The probability to score a long shot. | $\sim Beta(a,\beta)$ | Output |
| Long_shot_goals_HT/AT | Number of long shot goals. | $\sim Binomial(n,p)$ | Output |
| **Special events variables** | | | |
| Unpredictable_offensives_HT/AT<br>Unpredictable_SA_players_HT/AT<br>Unpredictable_LP_players_HT/AT<br>Unpredictable_mistake_players_HT/AT<br>Unpredictable_owngoal_players_HT/AT | The number of players with speciality *Unpredictable* who can generate the different types of unpredictable events. | $\sim Uniform(a,b)$ | Input |
| Quick_offensives_HT/AT<br>Quick_defenders_HT/AT | The number of offensive players and defenders with speciality *Quick*. | $\sim Uniform(a,b)$, Discrete | Input |
| Technical_ offensives_HT/AT | The number of offensive players with speciality *Technical*. | $\sim Uniform(a,b)$ | Input |
| Technical_defenders_HT/AT | The number of defenders with speciality *Technical*. | Discrete | Input |
| Corner_Head_offensives_HT/AT<br>Corner_Head_defensives_HT/AT | The number of players with the speciality *Head* who attack and/or defend corners. | $\sim Uniform(a,b)$ | Input |
| Head_offensives_HT/AT | The number of offensive players with speciality *Head*. | Discrete | Input |
| Head_defenders_or_IMs_HT/AT | Whether a team fielded defenders or midfielders with the speciality *Head*. | Discrete | Input |
| PNF_HT/AT<br>PDIM_HT/AT | The number of forwards and midfielders with the speciality *Powerful*. | Discrete | Input |
| Forwards_HT/AT<br>Wingers_HT/AT<br>CDs_and_WBs_HT/AT<br>Central_defenders_HT/AT | The number of forwards, wingers, central defenders and wing backs. | $\sim Uniform(a, b)$, Discrete | Input |
| Event_Unpred_[type]_HT/AT | The five types and numbers of *Unpredictable* events. | $\sim Binomial(n, p)$, Equation | Output |
| p_Score_Unpred_[type]_HT/AT | The probability to score for each of the five *Unpredictable* events. | $\sim Beta(a, \beta)$ | Output |
| Unpred_[type]_goals_HT/AT | The number of goals and own goals scored through *Unpredictable* events. | $\sim Binomial(n, p)$, Equation | Output |
| Unpred_counterattacks_HT/AT | The number of counterattacks generated due to an *Unpredictable* (special action or score on their own) event missed by the opponent. | $\sim Binomial(n, p)$ | Output |
| Unpred_SA_SO_missed_chances_HT/AT | *Unpredictable* (special action or score on their own) events missed. | $\sim Binomial(n, p)$ | Output |
| Event_Quick_HT/AT<br>Events_Quick_stopped_HT/AT | The number of *Quick* events. | $\sim Binomial(n, p)$ | Output |
| p_Event_Quick_HT/AT<br>p_Score_Quick_HT/AT<br>p_Quick_stop_HT/AT | The probability to generate, score, and stop a *Quick* event. | $\sim Beta(a, \beta)$ | Output |
| Quick_goals_HT/AT | The number of goals scored through *Quick* events. | $\sim Binomial(n, p)$ | Output |
| Event_Tech_over_Head_HT/AT | The number of *Technical*-over-*Head* events. | $\sim Binomial(n, p)$ | Output |
| p_Score_Tech_over_Head_HT/AT | The probability to score a *Technical-over*-Head event. | $\sim Beta(a, \beta)$, Equation | Output |
| Technical_goals_HT/AT | The number of goals scored through *Technical*-over-*Head* events. | $\sim Binomial(n, p)$ | Output |
| Technical_CA_HT/AT | The number of counterattacks generated from a player with speciality *Technical*. | $\sim Binomial(n, p)$ | Output |
| p_Technical_CA_HT/AT | The probability to generate a *Technical* counterattack. | $\sim Beta(a, \beta)$ | Output |
| Event_corner_HT/AT | The number of corner events. | $\sim Binomial(n, p)$ | Output |
| p_Event_corner<br>p_Event_corner_to_any_HT/AT | The probability of a corner event, and for the event to be corner-to-anyone (as opposed to corner-to-*Head*). | $\sim Beta(a, \beta)$ | Output |
| p_Score_corner_to_Head_HT<br>p_Score_corner_to_any_HT | The probability to score a corner-to-*Head* or corner-to-anyone event. | $\sim Beta(a, \beta)$, Equation | Output |
| Corner_to_any_goals_HT<br>Corner_to_Head_goals_HT | The number of goals scored through a corner-to-anyone or a corner-to-*Head* event. | $\sim Binomial(n, p)$ | Output |
| Event_winger_HT/AT | The number of winger events. | $\sim Binomial(n, p)$ | Output |
| Unpred_[type]_weight_HT<br>Tech_over_Head_weight_HT<br>Quick_weight_HT<br>Winger_weight_HT<br>Header_weight_HT/AT | Weights determining which team gets allocated a specified special event (for AT it is the complement of HT, if AT is not reported). | Equation | Output |

| p_Event_winger_to_any_HT/AT | The probability of a winger event to be winger-to-anyone (as opposed to winger-to-*Head*). | $\sim Beta(a, \beta)$ | Output |
|---|---|---|---|
| p_Score_winger_to_any_HT/AT | The probability to score a winger-to-anyone event. | $\sim Beta(a, \beta)$ | Output |
| Avg_score_wing_attack_HT/AT | The average probability to score through a wing attack (used as a proxy for determining the probability to score a winger event). | Equation | Output |
| Winger_to_anyone_goals_HT/AT<br>Winger_to_head_goals_HT/AT | The number of goals scored through a winger-to-anyone or winger-to-*Head* events. | $\sim Binomial(n, p)$ | Output |
| Event_IneExp_Def_HT/AT<br>Event_Exp_FW_HT/AT<br>Event_Tired_Def_HT/AT | The numbers of inexperienced defender, experienced forward, and tired defender events. | $\sim Binomial(n, p)$ | Output |
| p_Event_InExp_Def_HT/AT<br>p_Event_Exp_FW_HT/AT<br>p_Event_Tired_Def | The probabilities of inexperienced defender, experienced forward, and tired defender events. | $\sim Beta(a, \beta)$ | Output |
| Score_InExp_Def<br>p_Score_Exp_FW<br>Score_Tired_Def | The probabilities of scoring inexperienced defender, experienced forward, and tired defender events. | $\sim Beta(a, \beta)$ | Output |
| InExp_Def_goals_HT/AT<br>Exp_FW_goals_HT/AT<br>Tired_Def_goals_HT/AT | The numbers of goals scored through inexperienced defender, experienced forward, and tired defender events. | $\sim Binomial(n, p)$ | Output |
| PNF_chances_HT/AT | The number of Powerful Normal Forward events. | $\sim Binomial(n, p)$ | Output |
| p(PNF_Score)_HT/AT | The probability to score a Powerful Normal Forward event. | $\sim Beta(a, \beta)$ | Output |
| PNF_goals_HT/AT | The goals scored through a Powerful Normal Forward event. | $\sim Binomial(n, p)$ | Output |
| Player_SEs<br>Team_SEs | The number of player-based and team-based special events. | $\sim Binomial(n, p)$ | Output |
| Linear_possession_for_TSEs_HT | The possession rate for team-based special events (for AT it is the complement of HT). | Equation | Output |
| SE_goals_HT/AT | The goals scored through special events. | Equation | Output |

**Appendix B:** Description of Beta-Binomial representation

Formally, the Beta-Binomial model consists of two layers. The first layer specifies a Beta distribution as a prior over the probability of success $p$ used in the Binomial distribution. Specifically, $p \sim Beta(a, \beta)$, where $a$ represents the prior pseudo-counts of successes and $\beta$ the prior pseudo-counts of failures. The Probability Density Function (PDF) of a Beta distribution is:

$$f(p|a, \beta) = \frac{p^{a-1}(1-p)^{\beta-1}}{B(a, \beta)}$$

where $B(a, \beta)$ is the Beta function:

$$B(a, \beta) = \int_0^1 p^{a-1}(1-p)^{\beta-1} dp$$

and the mean of Beta is:

$$E[p] = \frac{a}{a+\beta}$$

The $p$ is the sampled from the Beta distribution to serve as the input $p$ in the Binomial distribution. Specifically, the observed number of successes $X$ in n trials follows a Binomial distribution $X|p \sim Binomial(n, p)$ which means:

$$P(X = k|p) = \binom{n}{k} p^k (1-p)^{n-k}, \quad k = 0,1, \dots, n$$

where $n$ is the number of trials and $k$ is the number of successes. However, $X$ is not directly observed in the BN model. Instead, inference is performed over the latent binomial variable using the Beta prior. Since the binomial outcomes are unobserved, the posterior distribution of $p$ is inferred based on other observed nodes in the BN model.

## **Appendix C:** Main specification of KB-regression that differs from KB-probabilistic

This appendix section focuses on key parts of the KB-regression BN model that specifically differ from the KB-probabilistic, typically with regression equations that are either taken directly from the *Unwritten manual*, or from summary data tables within that manual or other widely read in-game forum posts. These key changes between models, and with reference to the implementations within the KB-regression model, are:

a. Scoring from set-pieces (DFKs, IFKs, and PKs): the probability of scoring from set-pieces is estimated using a non-linear regression described by Equation C.1, based on in-game forum table data [post=17493341.24]:

$$ATTACK_{SP_{HT}} = -0.0000380429(R_{HT}^{A_F} - R_{AT}^{D_F})^3 + 0.0000226846(R_{HT}^{A_F} - R_{AT}^{D_F})^2 + 0.0366246(R_{HT}^{A_F} - R_{AT}^{D_F}) + 0.45515 \tag{C.1}$$

where $ATTACK_{SP_{HT}}$ represents scoring a set-piece attack for home team (same concept applies to the away team), $R_{HT}^{A_F}$ is the Indirect Set-Piece (ISP) attack rating for the home-team, and $R_{AT}^{D_F}$ is the ISP defence rating for the away team. This relationship is also depicted in Figure C.1.

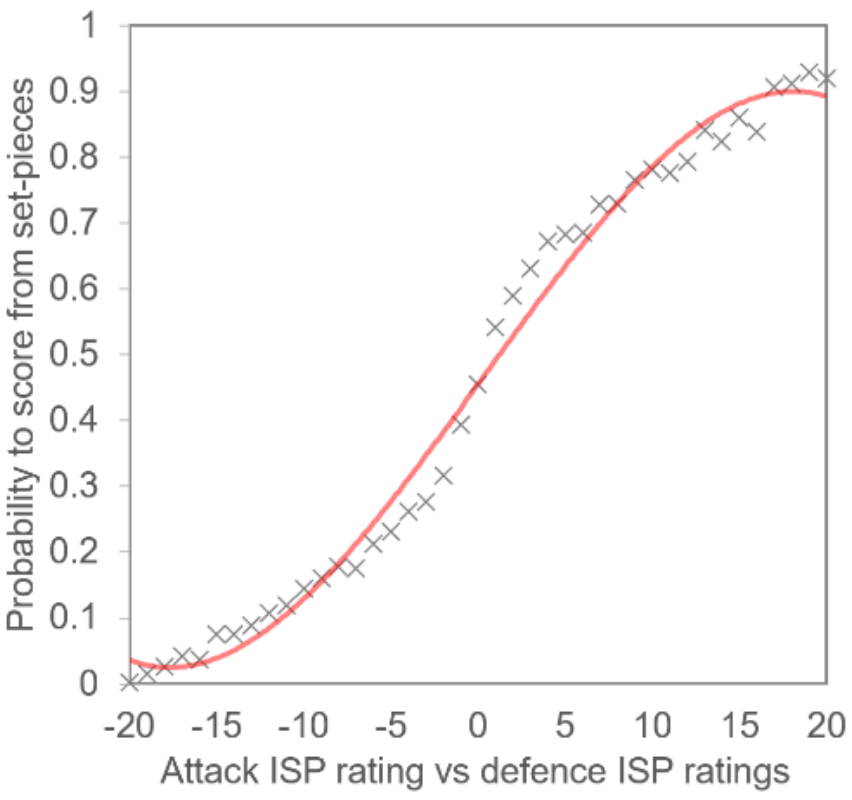

**Figure C.1.** The probability of scoring from set-pieces given the difference in ISPs ratings, as given in Equation C.1, where x are real data points.

b. Equation C.2 describes the regression equations, based on in-game forum posts 17493341.6, 17537676.367, 17493341.20, and 17342010.653, corresponding to each of the five regression lines presented in Figure C.2 that are used to compute each Tactic Conversion Rate ($TCR$) given the tactic rating $R^T$.

$$TCR(R^T) = \begin{cases} -0.617941717072569 + 0.104274398.R^T \\ \quad -0.00358354796.{R^T}^2 + 0.0000434356.{R^T}^3, & \text{Counter} \\ -0.00036765.{R^T}^2 + 0.02180462.R^T + 0.0705084, & \text{AiM} \\ -0.00046569.{R^T}^2 + 0.02894608.R^T + 0.10514706, & \text{AoW} \\ 0.00761935.R^T + 0.07520052, & \text{Long Shot} \\ -0.00780421.{R^T}^2 + 0.471402.R^T - 1.10735, & \text{Pressing} \end{cases} \tag{C.2}$$

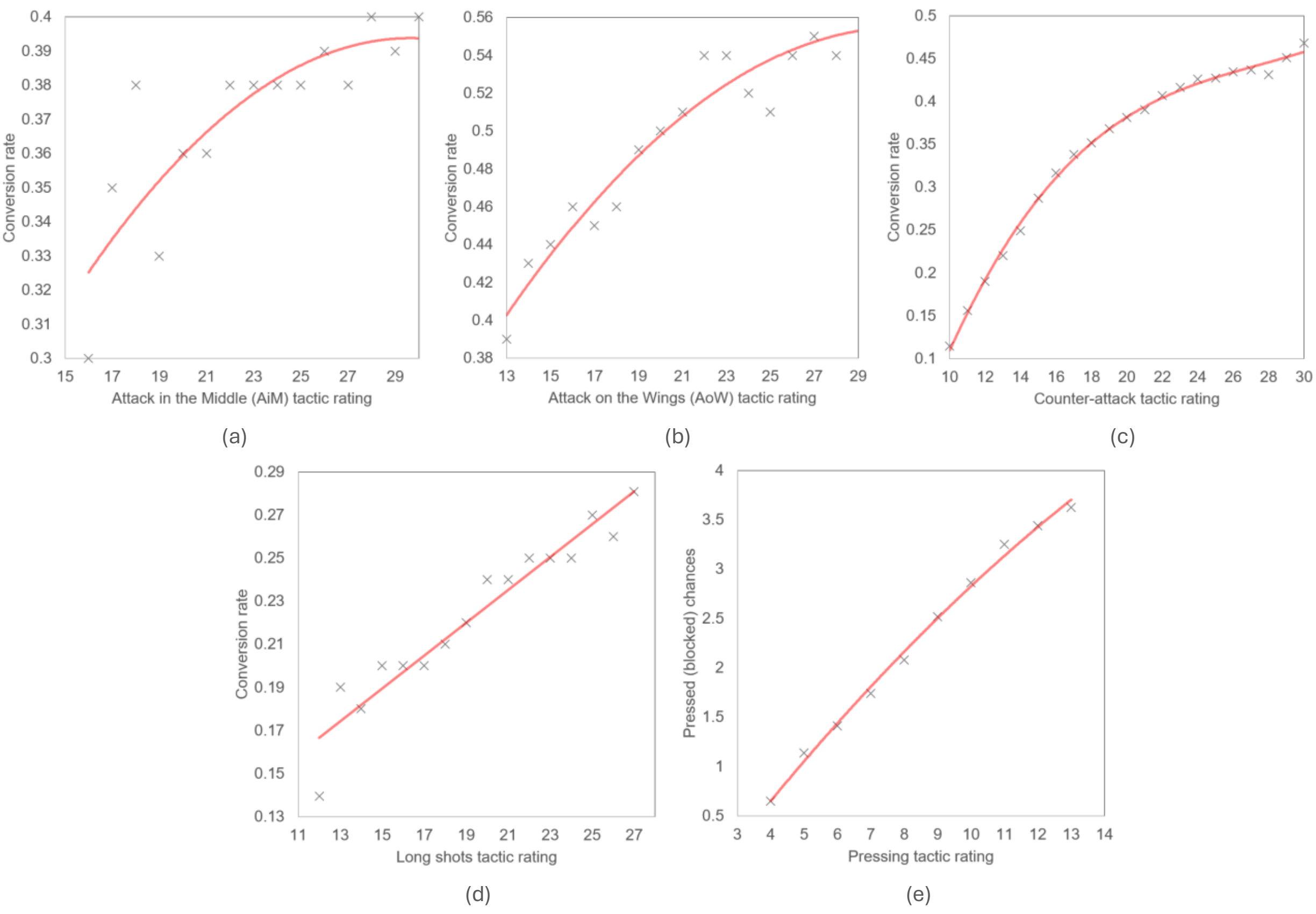

**Figure C.2.** The relationships between conversion rates and tactic ratings for the five specified tactics. Marker x denotes empirical data.

c. Table C.1 reports the frequencies implemented for team-based events, based on related in-game forum posts 17192231.1, and 17192231.3.

**Table C.1.** The assumed team-based events and their overall frequency, based on related in-game forum posts.

| Team-based event | Frequency |
|---:|---|
| CornerToHeader - goal | 8.19% |
| CornerToHeader - no goal | 5.21% |
| CornerToAnyone - goal | 13.94% |
| CornerToAnyone - no goal | 5.58% |
| TiredDefenderMistake - goal | 0.31% |
| TiredDefenderMistake - no goal | 0.22% |
| ExperiencedForward - goal | 2.69% |
| ExperiencedForward - no goal | 1.81% |
| InexperiencedDefender - goal | 1.21% |
| InexperiencedDefender - no goal | 3.45% |

d. Assumes that having a PDIM will stop 6.5% of the opponent's normal attacks, on average, based on in-game forum post 17342010.652, and that the conversion rate for PNFs is as described by Equation C.3, that maps the number of PNFs against the number of opponent CDs to estimate the conversion rate $CR_PNF$.

$$CR_PNF(PNF, CD) = \begin{cases} 0.096, & PNF = 1, CD = 0 \\ 0.069, & PNF = 1, CD = 1 \\ 0.033, & PNF = 1, CD = 2 \\ 0.020, & PNF = 1, CD = 3 \\ 0.117, & PNF = 2, CD = 0 \\ 0.096, & PNF = 2, CD = 1 \\ 0.052, & PNF = 2, CD = 2 \\ 0.031, & PNF = 2, CD = 3 \\ 0.066, & PNF = 3 \end{cases} \tag{C.3}$$

## **Appendix D:** Graphical structures produced by the algorithms

This section presents the graphical outputs generated by the structure learning algorithms; the PDAG for PC and the DAGs for all other algorithms. In these graphs, observed input nodes are highlighted in orange, while output nodes of interest are shown in green. These visualisations provide insights into complexity and correctness in relation to the knowledge-based graph depicted in Figure 17.

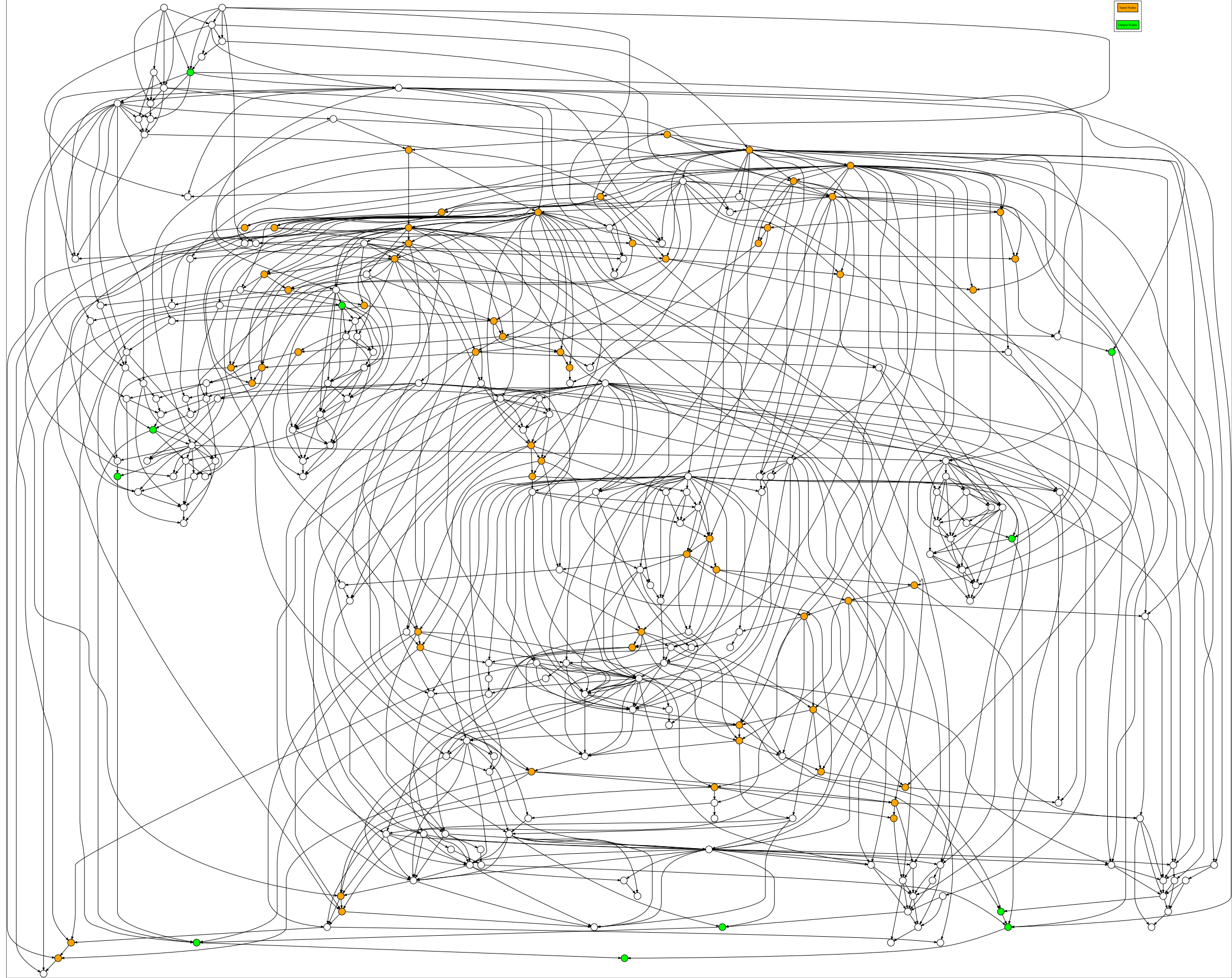

**Figure D.1.** The learnt graph of the FGES structure learning algorithm, where orange nodes represent input nodes, and green nodes denote output nodes of interest.

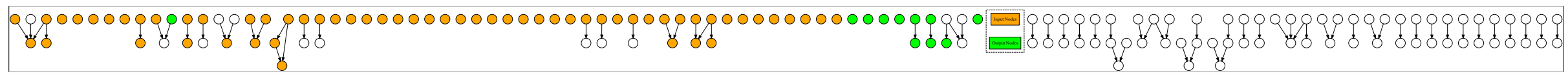

**Figure D.2.** The learnt graph of the GS structure learning algorithm, where orange nodes represent input nodes, and green nodes denote output nodes of interest.

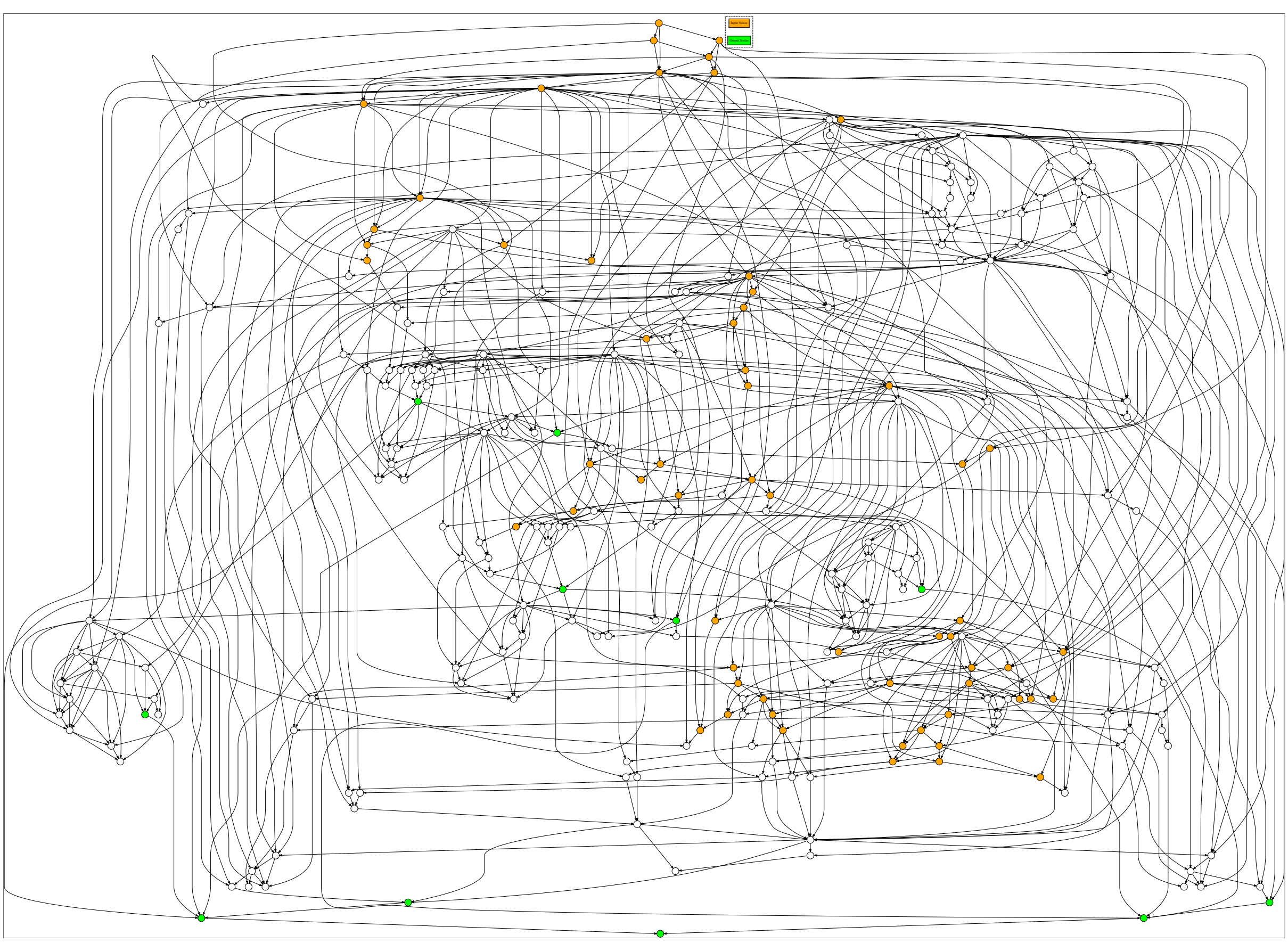

**Figure D.3.** The learnt graph of the HC structure learning algorithm, where orange nodes represent input nodes, and green nodes denote output nodes of interest.

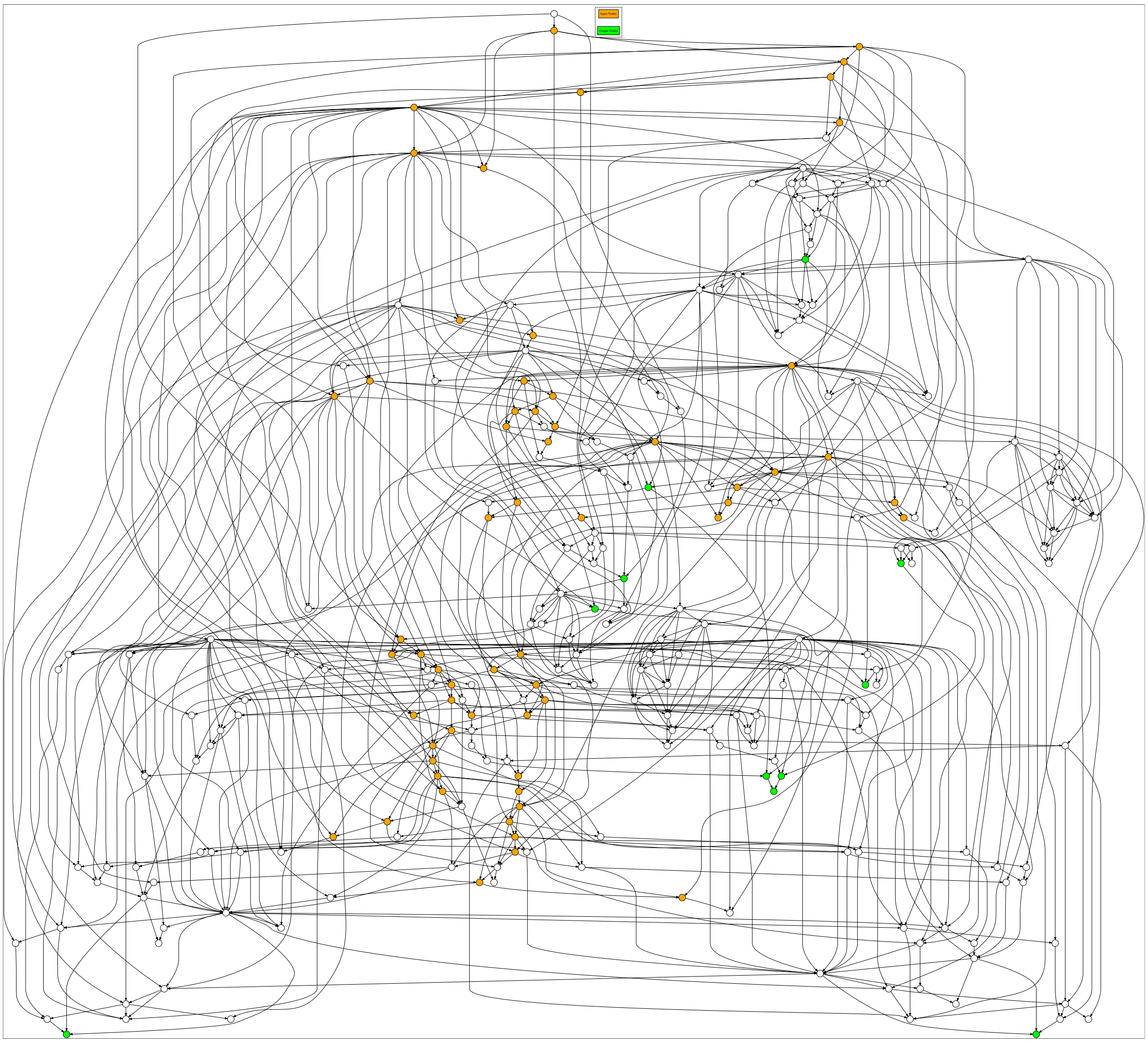

**Figure D.4.** The learnt graph of the HC-Stable structure learning algorithm, where orange nodes represent input nodes, and green nodes denote output nodes of interest. Excludes four edges entering *Goals_HT* and *Goals_AT* (two in each) to enable generation of image.

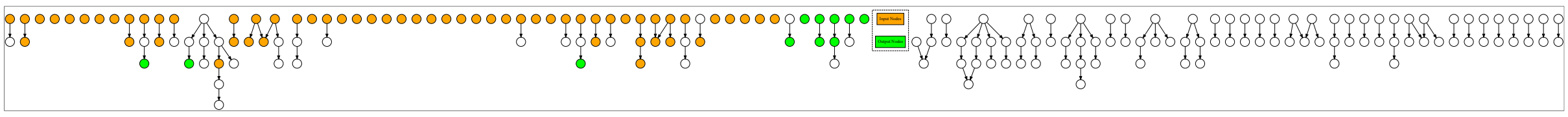

**Figure D.5.** The learnt graph of the MMHC structure learning algorithm, where orange nodes represent input nodes, and green nodes denote output nodes of interest.

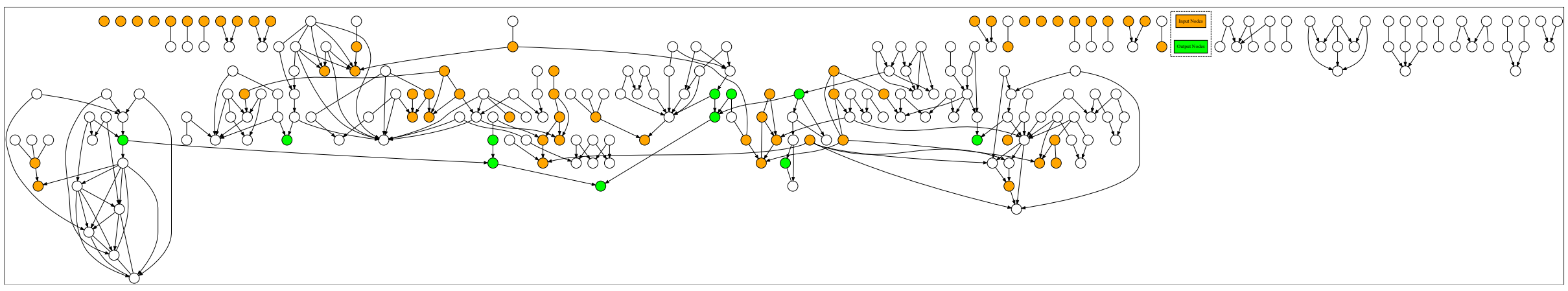

**Figure D.6.** The learnt graph of the PC-Stable structure learning algorithm, where orange nodes represent input nodes, and green nodes denote output nodes of interest.

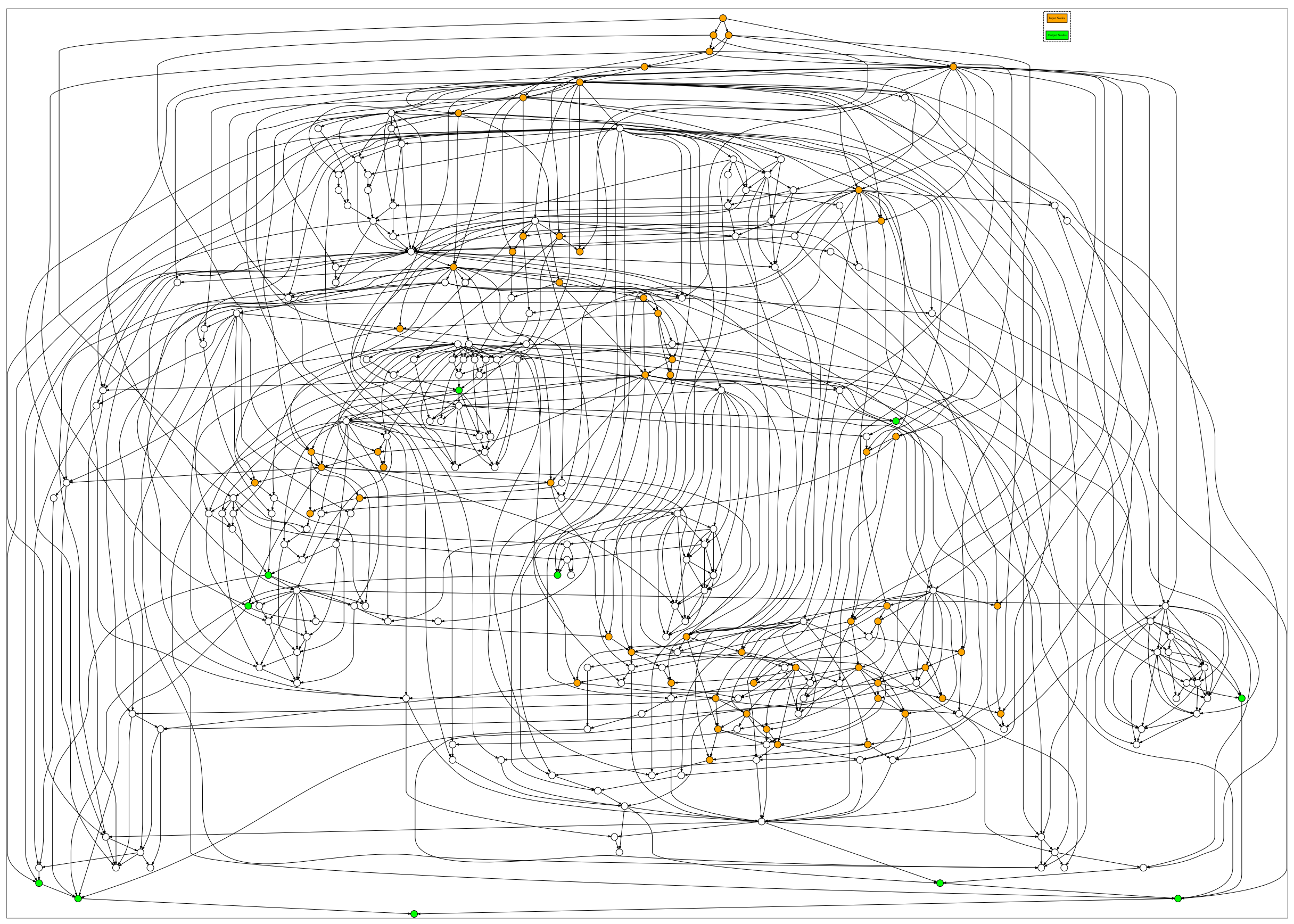

**Figure D.7.** The learnt graph of the TABU structure learning algorithm, where orange nodes represent input nodes, and green nodes denote output nodes of interest.

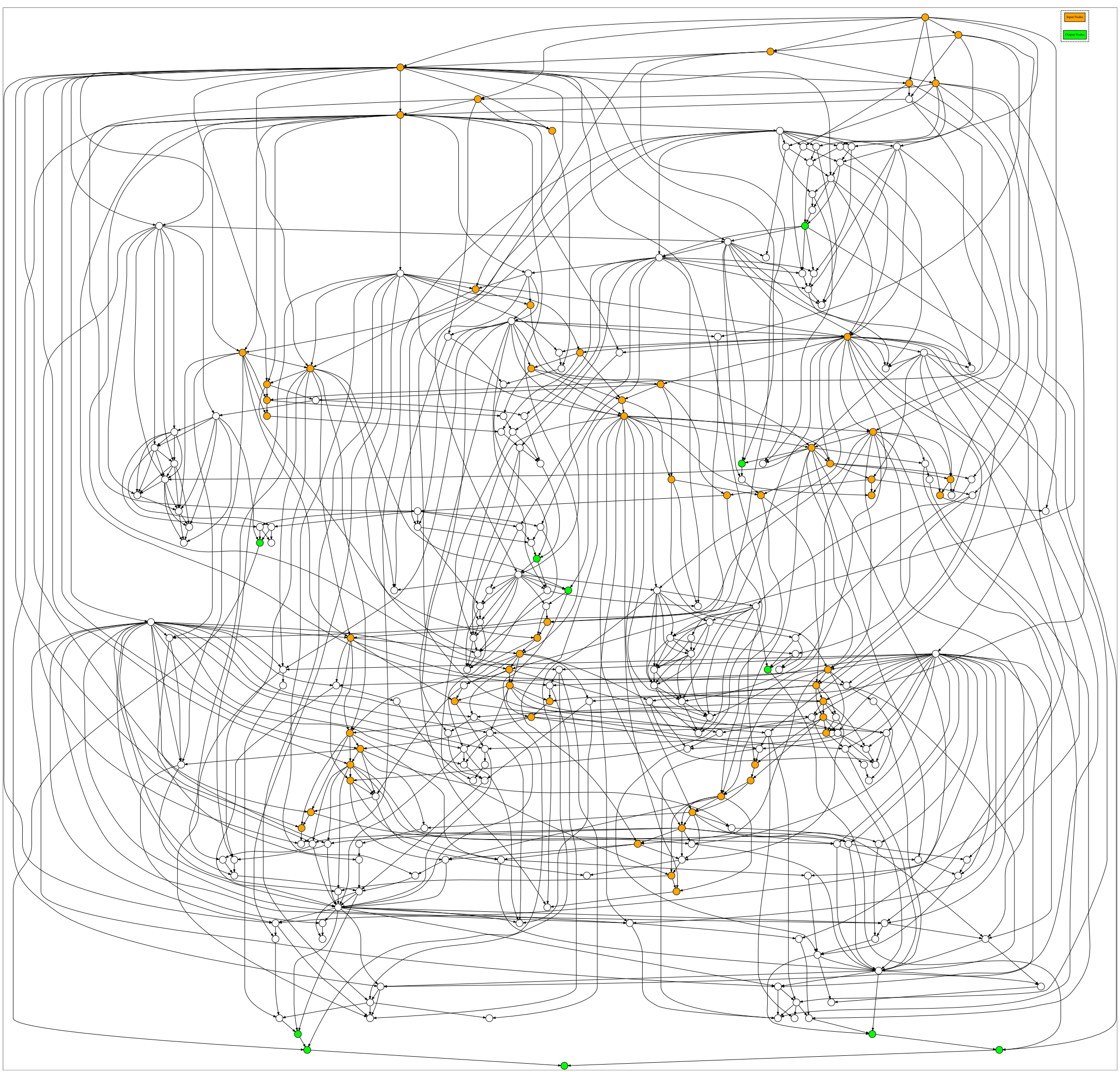

**Figure D.8.** The learnt graph of the TABU-Stable structure learning algorithm, where orange nodes represent input nodes, and green nodes denote output nodes of interest. Excludes four edges entering *Goals_HT* and *Goals_AT* (two in each) to enable generation of image.

## **Appendix E:** Supplementary results highlighting directions for future improvements

This appendix focuses on the best-performing BN model identified in this study, KB-probabilistic, which integrates expert knowledge with data, alongside the best-performing Hattrick model, Nickarana. It provides a deeper analysis of their results, disentangling performance across different team tactics discussed in subsection 4.3. This examination aims to identify potential modelling limitations for each playing strategy and highlight directions for future improvements.

Figure E.1 presents the HDA performance of the KB-probabilistic (left) and Nickarana (right) models, evaluated across all possible team tactics. The evaluation considers both (a) any tactic versus the default normal tactic, and (b) each tactic against itself to enable a more extreme assessment of its effectiveness by the two models, capturing 88% of all matches. The charts are ordered by best (lowest) RPS score and demonstrate a consistency in the ranking of each combination of tactics, despite the two models being based on entirely different modelling frameworks. Figure E.2 replicates this analysis for goals scored, and the results largely mirror those of Figure E.1 which are based on RPS.

However, both the HDA and score analyses showed two inconsistencies worth exploring. Firstly, the NM vs PS tactic achieves a lower RPS value for KB-probabilistic (Figure E.1) in HDA outcomes, indicating better performance. However, this improvement in RPS is accompanied by higher variance compared to the Nickarana model, which is not ideal. In terms of goal difference (Figure E.2), KB-probabilistic also outperforms Nickarana for matches involving the PS tactic, generating a lower goal difference. However, in this case, the variance is lowest compared to all other tactic combinations for both models.

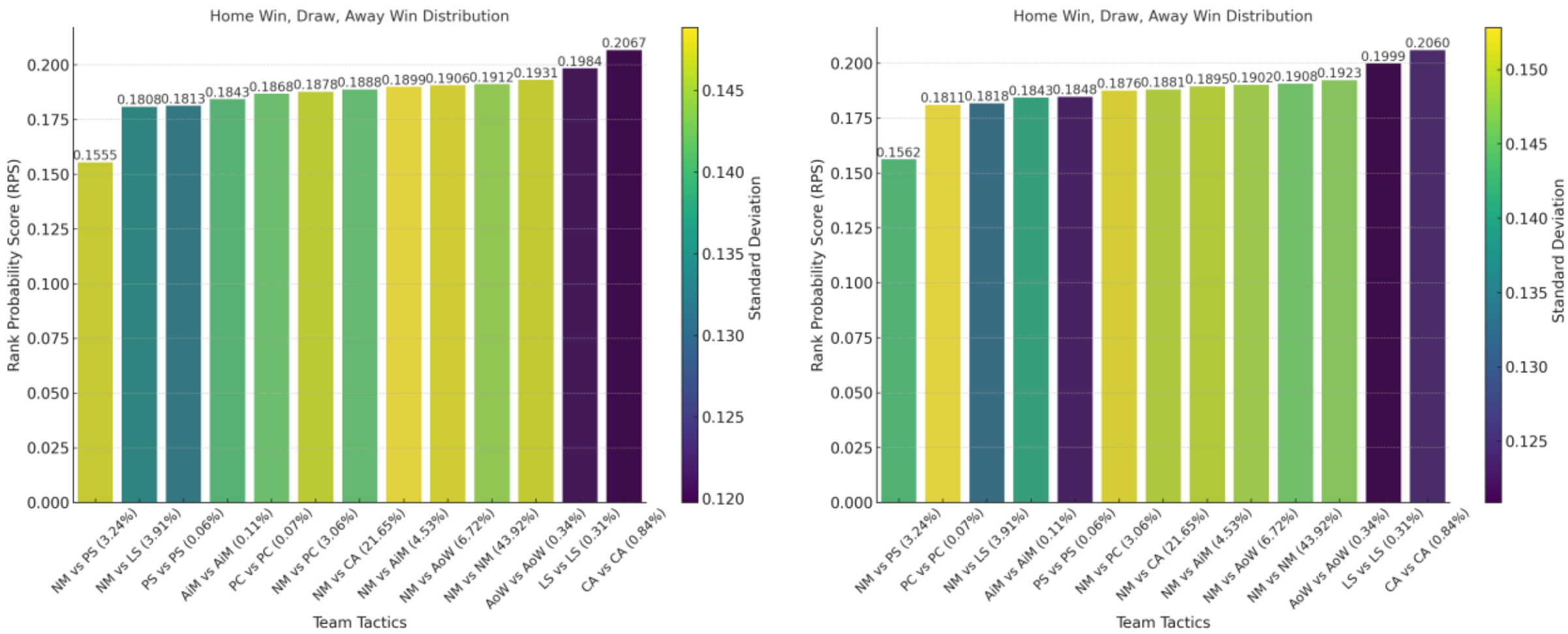

**Figure E.1.** The RPS performance of the KB-probabilistic (left) and Nickarana (right) models across different team tactics, covered in subsection 4.3, iterating over all tactics against the normal (NM) tactic, and each tactic against itself, ordered by best (lowest) RPS score. The percentages on $x$-axis indicate the occurrence rate of each tactic combination. Darker coloured bars represent lower standard deviation.

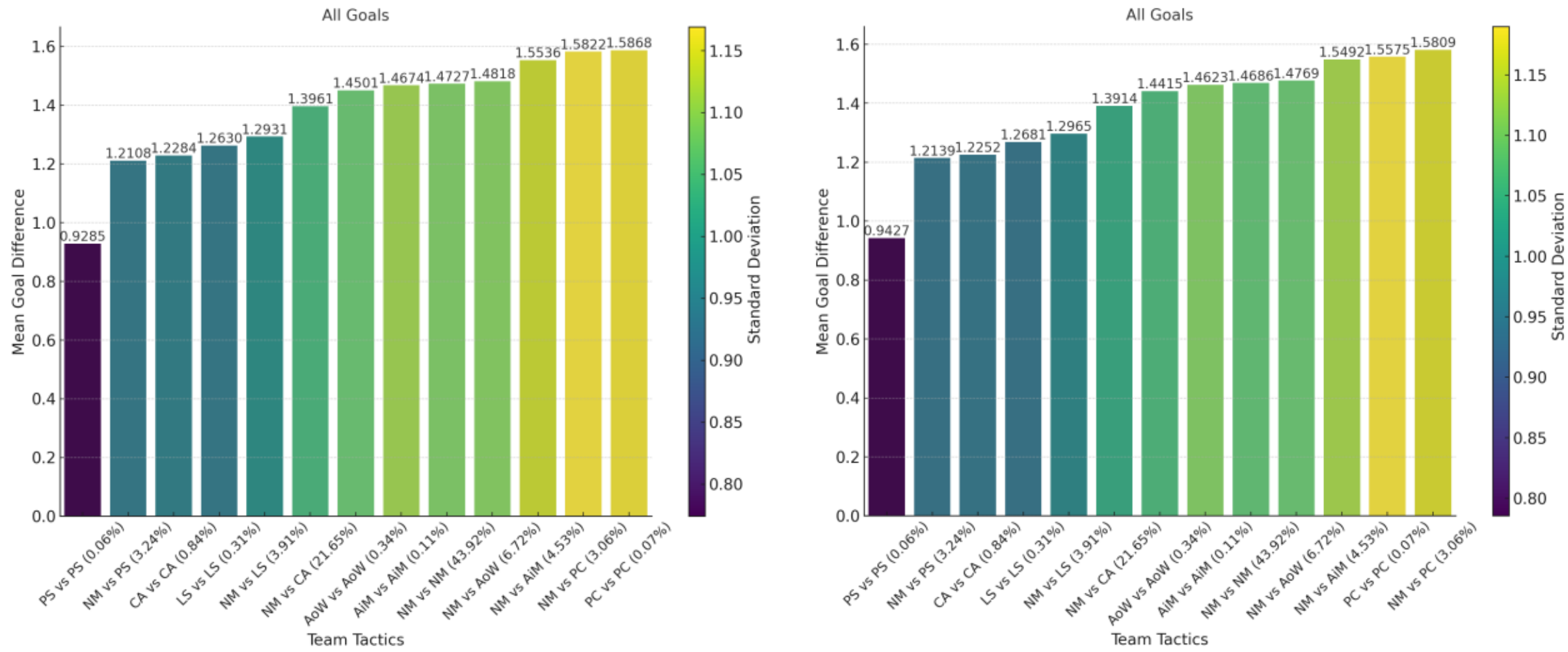

**Figure E.2.** The goals scored performance of the KB-probabilistic (left) and Nickarana (right) models across different team tactics covered in subsection 4.3, iterating over all tactics against the normal (NM) tactic, and each tactic against itself, ordered by best (lowest) goal difference. The percentages on $x$-axis indicate the occurrence rate of each tactic combination. Darker coloured bars represent lower standard deviation.

Secondly, a similar pattern is observed for the PC vs PC (extends to NM vs PC) tactic. The Nickarana model achieves a lower RPS value (Figure E.1), but, similarly to the above observation, this improvement comes at the cost of higher variance compared to KB-probabilistic. In terms of goal difference (Figure E.2), Nickarana also produces a lower goal difference for matches involving the PC tactic compared to KB-probabilistic. However, in this case, the variance is highest for both models, indicating a potential area for improvement.

Figure E.3 extends the analysis to actual goals scored, rather than goal difference, comparing the inferred distributions obtained from both models to the empirical data distributions. Overall, the results indicate that while the expected values of actual goals scored are similar, the observed distributional variance is higher than the estimated variance across almost all tactics. This suggests the presence of latent factors not includes in the models.

Specifically, both the KB-probabilistic and Nickarana models perform well in *Normal*, AiM (Attack in the Middle), AoW (Attack on the Wings), and PS (Pressing) tactics, accurately estimating both the expected value and variance; though they slightly underestimate the former in all four cases. For the PC tactic, however, both models deviate from the expected value. That is, KB-probabilistic underestimates goals scored, while Nickarana overestimates them. It is not surprising that PC remains poorly understood given its long-standing debate and uncertainty within the community. In contrast, the LS (Long-shots) tactic is assumed to be well understood by the community, and yet both models underestimate goals scored, highlighting that some relevant mechanisms continue to remain unknown.

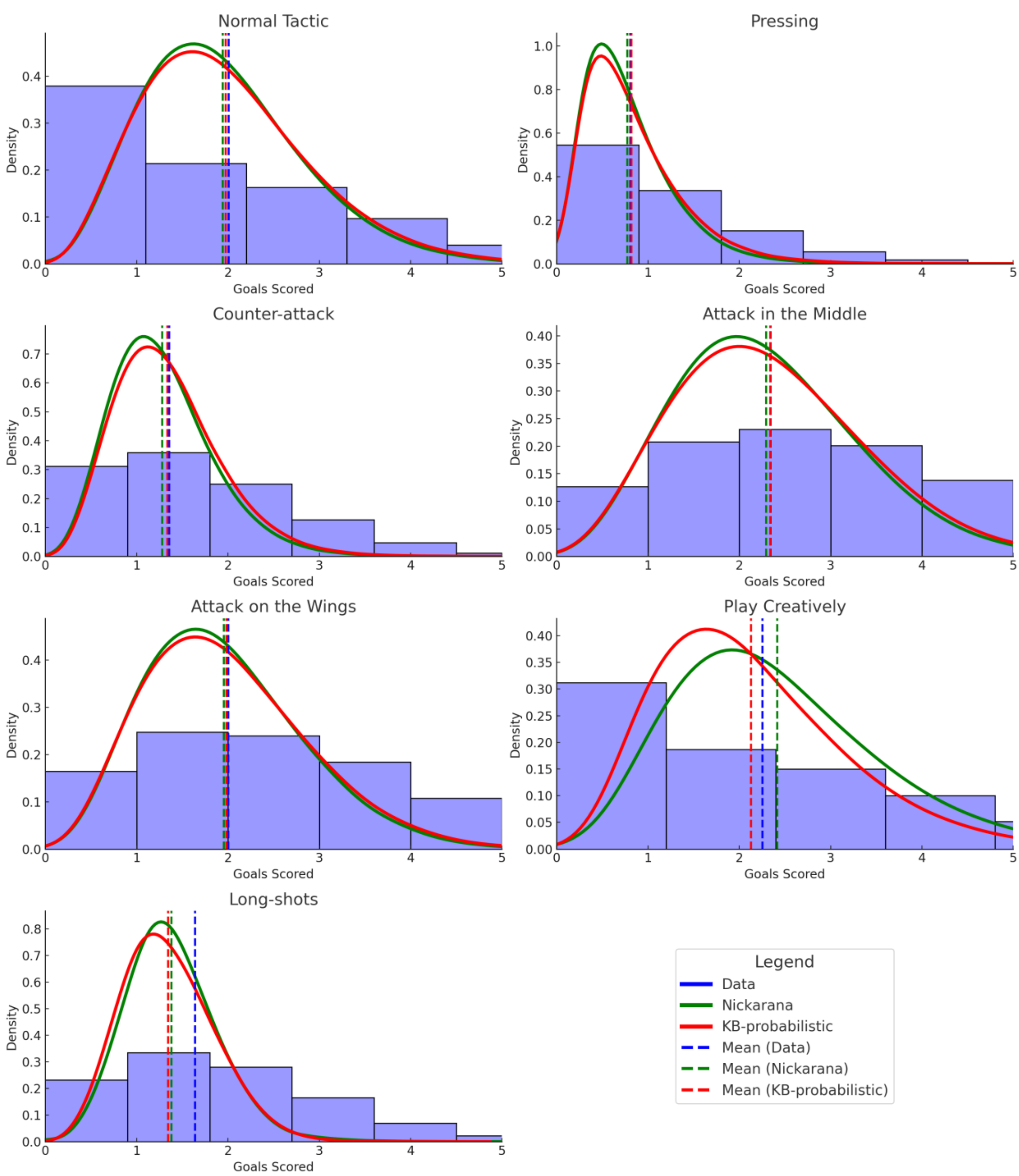

**Figure E.3.** Goals scored for each tactic as predicted by the KB-probabilistic and Nickarana models, compared against observed data. Model results are presented as KDE functions, while observed data using histograms due to its discrete nature.